\documentclass[runningheads]{llncs}

\usepackage{eccv}

\usepackage{eccvabbrv}

\usepackage{graphicx}
\usepackage{booktabs}
\usepackage{soul,multirow}
\usepackage{amssymb}    
\usepackage[normalem]{ulem} 

\definecolor{eccvblue}{rgb}{0.12,0.49,0.85}
\definecolor{eccvred}{rgb}{1.0,0.0,0.0}

\usepackage[pagebackref,breaklinks, colorlinks,citecolor=eccvblue]{hyperref}

\usepackage{orcidlink}

\usepackage{stackengine}

\usepackage{fancyhdr}
\makeatletter
\fancypagestyle{myfootnotes}{%
	\fancyhf{}%
	\fancyfoot[L]{%
		\parbox[b]{\textwidth}{\footnotesize
			\parindent1em\leftskip1em
			\kern-3\p@
			\hrule\@width 2truecm\relax
			\kern2.6\p@
			\noindent\llap{\hb@xt@1em{\hss\textsuperscript{*}}\ }Equal contribution.\par
			\noindent\llap{\hb@xt@1em{\hss\textsuperscript{\dag}}\ }Corresponding author: \email{lizhang@hfut.edu.cn}}%
	}%
}
\makeatother
\begin{document}
	
	\title{Towards Robust Driving Perception: A Flexible Scale-Driven Family for Self-Supervised Monocular Depth Estimation} 
	
	\titlerunning{FlexDepth}
	
	\author{
		Zhaowen Zhu\inst{1,3}\textsuperscript{*}\orcidlink{0009-0001-7194-0089}, 
		Li Zhang\inst{1}\textsuperscript{\dag}\orcidlink{0000-0002-9208-1949}, 
		Yujie Chen\inst{1}\textsuperscript{*}\orcidlink{0009-0005-6360-9993}, 
		Tian Zhang\inst{1,4}\orcidlink{0009-0004-2646-3430}, \\
		Yingjie Wang\inst{2}\orcidlink{0000-0001-6852-8491}, 
		Mingxia Zhan\inst{1}\orcidlink{0009-0000-4345-8490}
	}
	
	\authorrunning{Z.~Zhu et al.}
	
	\institute{Hefei University of Technology, Hefei, China \and
		Nanjing University of Posts and Telecommunications, Nanjing,  China\and
		Chengpin Home Tech, Nanjing, China\and
		Differential Robotics, Hangzhou, China
	}
	
	\maketitle
	\thispagestyle{myfootnotes}

	\begin{figure}[htbp]
		\centering
		\setlength{\tabcolsep}{1pt}
		\begin{minipage}[c]{0.46\textwidth}
			\centering
			\includegraphics[width=\textwidth, trim=0 0 0 5, clip]{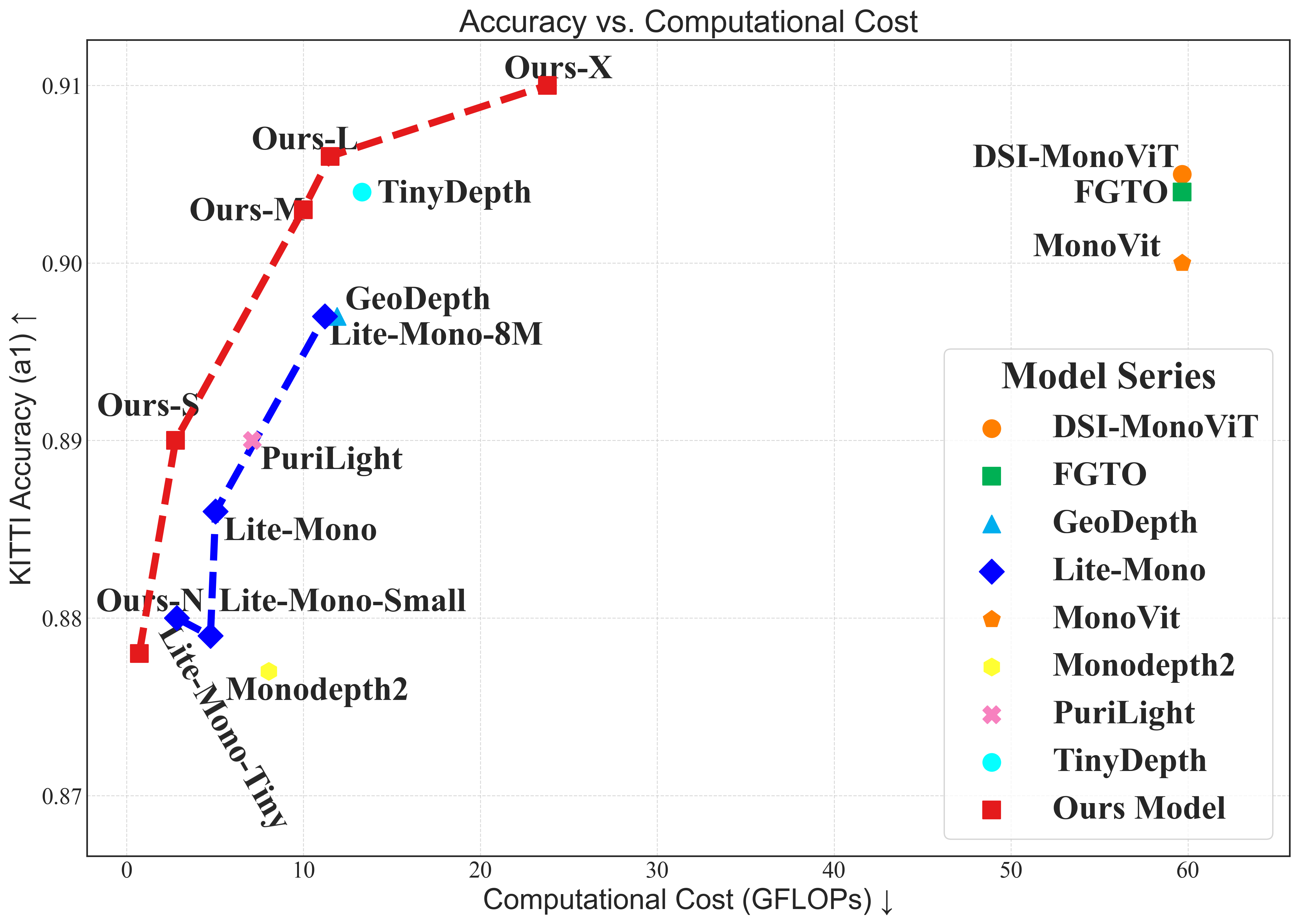}
			
		\end{minipage}
		\hfill
		\begin{minipage}[c]{0.52\textwidth}
			\centering
			\renewcommand{\arraystretch}{0.8}
			\begin{tabular}{rcc}
				\rotatebox{90}{\tiny \hspace{1em}Input} & 
				\includegraphics[width=0.45\linewidth, ]{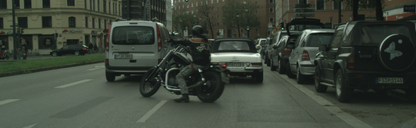} &
				\stackinset{r}{3pt}{t}{18pt}{\tiny \color{white}\textmd{Ground-truth}}{\includegraphics[width=0.45\linewidth, trim=20 20 20 20, clip]{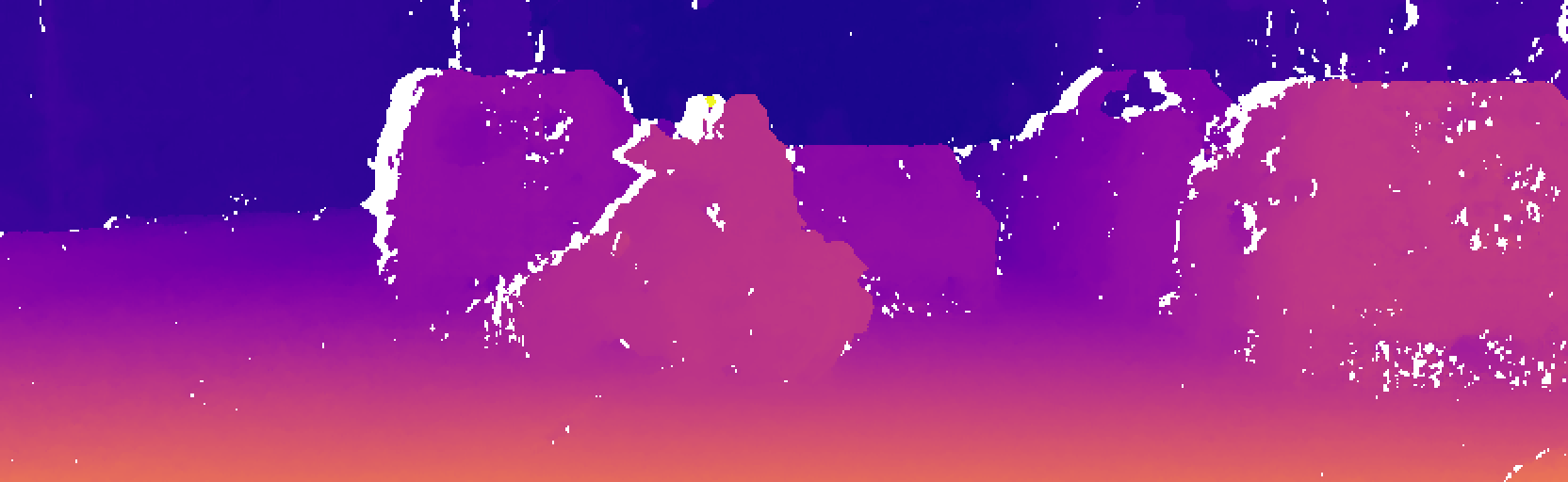}} \\
				
				\rotatebox{90}{\tiny \hspace{1em}~\cite{nguyen2024mining}} &
				\stackinset{r}{3pt}{t}{18pt}{\tiny \color{white}\textmd{Error map}}{\includegraphics[width=0.45\linewidth, trim=20 20 20 20, clip]{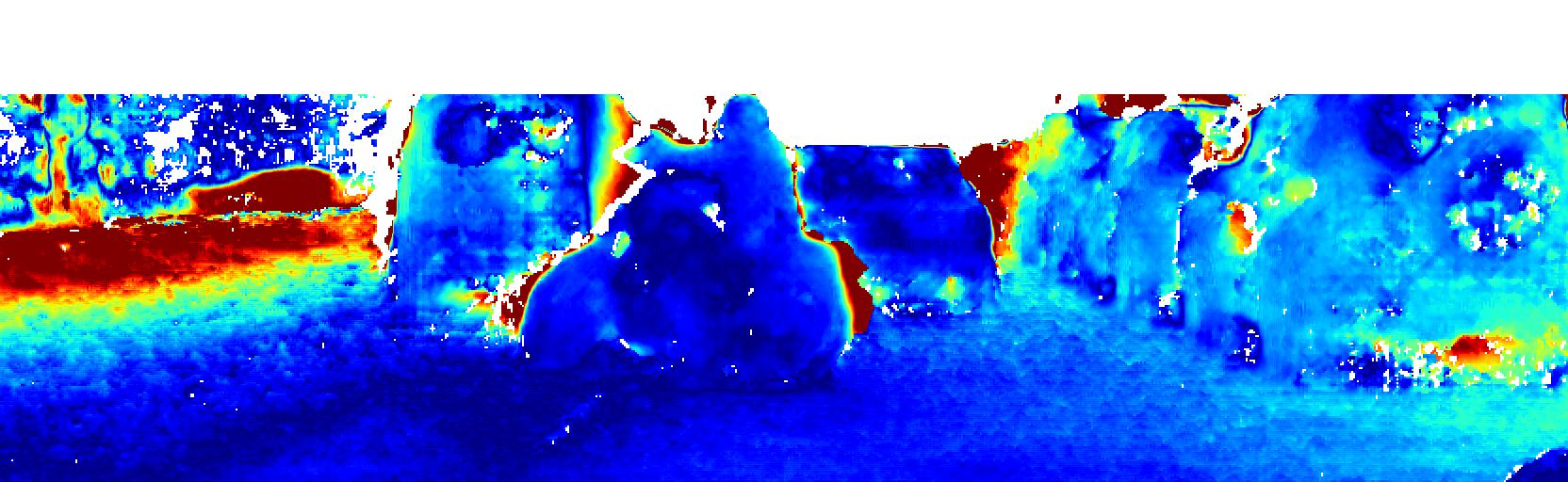}} &
				\stackinset{r}{3pt}{t}{18pt}{\tiny \color{white}\textmd{Prediction}}{%
					\stackinset{l}{22pt}{t}{20pt}{\tikz[overlay]{\draw[red, line width=1.2pt] (0,0) rectangle (25pt,18pt);}}{%
						\includegraphics[width=0.45\linewidth, trim=20 20 20 20, clip]{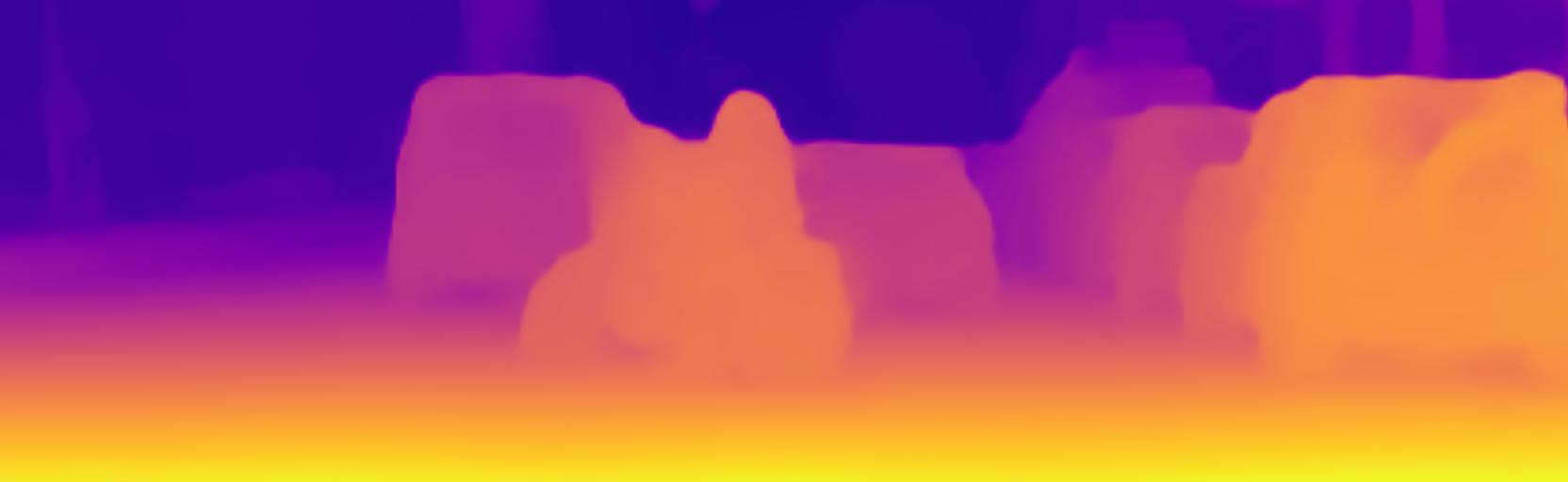}}%
				} \\
				
				\rotatebox{90}{\tiny \hspace{1em}Ours} &
				\stackinset{r}{3pt}{t}{18pt}{\tiny \color{white}\textmd{Error map}}{\includegraphics[width=0.45\linewidth, trim=20 20 20 20, clip]{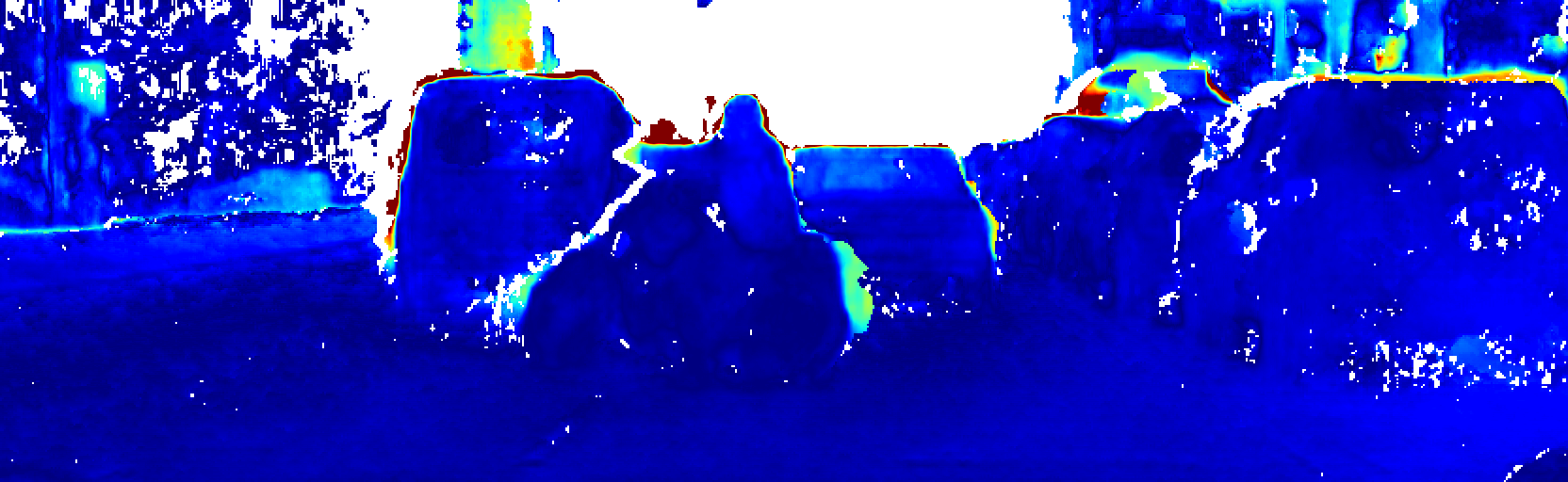}} &
				\stackinset{r}{3pt}{t}{18pt}{\tiny \color{white}\textmd{Prediction}}{%
					\stackinset{l}{22pt}{t}{20pt}{\tikz[overlay]{\draw[red, line width=1.2pt] (0,0) rectangle (25pt,18pt);}}{%
						\includegraphics[width=0.45\linewidth, trim=20 20 20 20, clip]{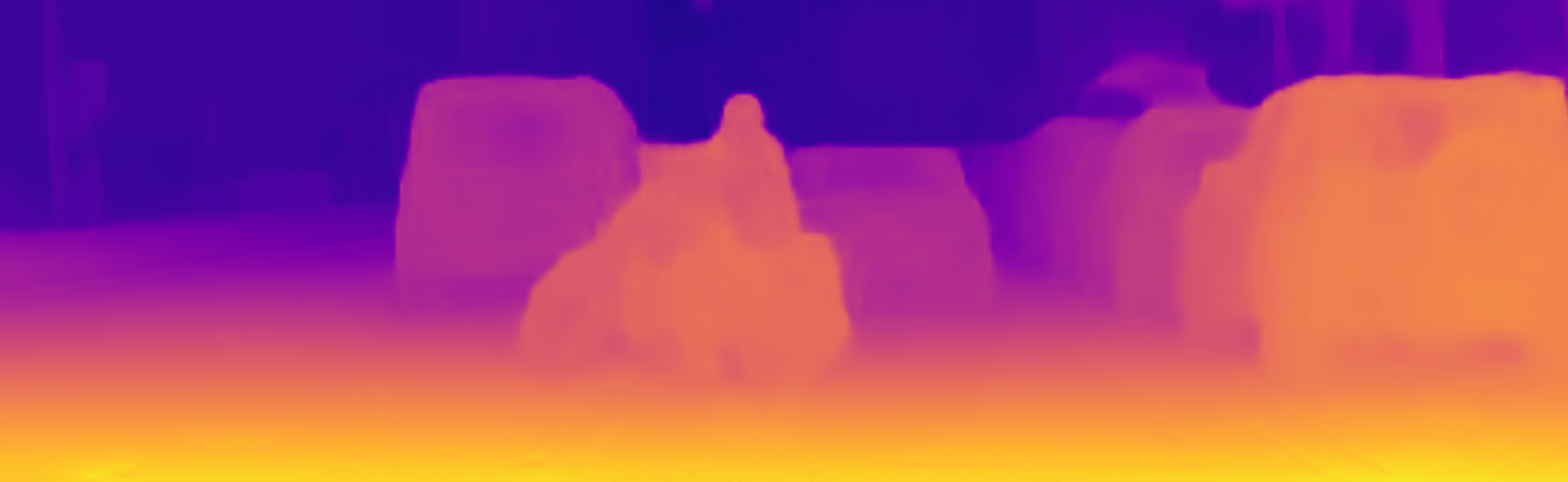}}%
				} 
			\end{tabular}
		\end{minipage}

		\caption{\textbf{Performance and visual comparison.} (left) Accuracy vs. cost on KITTI~\cite{6248074}. (right) Cityscapes~\cite{cordts2016cityscapes}: sharper boundaries and robust dynamic depth.}
		\label{fig:main_results}
	\end{figure}
	
	\begin{abstract}
		Self-Supervised Monocular Depth Estimation (MDE) has garnered attention in recent years due to its independence from ground truth. However, most existing models are limited to a single scale and exhibit considerable performance degradation in complex driving environments. Networks specifically designed to handle dynamic traffic participants tend to be overly complex, hindering their deployment on resource-constrained automotive edge devices. To address these limitations and move towards robust driving perception, we propose FlexDepth, a scale-driven and flexible family of self-supervised MDE models tailored for challenging road scenarios. FlexDepth employs a two-stage static-dynamic decoupled training strategy, enabling the independent assessment of confidence for both static backgrounds and dynamic road objects. Furthermore, it introduces a meticulously designed Scale-Driven Decoder (SDD) to dynamically select components based on scale size, facilitating efficient feature fusion and the output of high-precision depth maps. Extensive experiments on standard driving benchmarks demonstrate that without any auxiliary information, our model achieves state-of-the-art performance across arbitrary scales with minimal computational overhead. Our smallest model, Flex-Nano, requires only 0.7 GFLOPs and achieves 37.6 FPS on mobile platforms, ensuring reliable real-time perception while maintaining excellent zero-shot generalization. Our source code is available: \url{https://github.com/startnew/flexdepth}
	\end{abstract}
	
	\section{Introduction}
\label{sec:intro}

Monocular depth estimation aims to recover 3D scene geometry from a single image and plays a fundamental role in autonomous driving, robotic navigation, and Unmanned Aerial Vehicle (UAV) perception. 
Due to the scarcity of dense ground-truth depth in real-world scenarios, self-supervised approaches~\cite{garg2016unsupervised,godard2019digging,guizilini20203d,watson2021temporal,zhang2023lite} have gained significant attention. 
These methods construct supervision from stereo pairs~\cite{garg2016unsupervised,godard2017unsupervised,gonzalezbello2020forget} or monocular video sequences~\cite{godard2019digging,watson2021temporal,zhao2022monovit}, enabling scalable learning without explicit depth labels. 
Early models often rely on single-scale designs~\cite{johnston2020self,aleotti2018generative,wang2018learning} and degrade notably in dynamic scenes~\cite{zhou2024recurrent}. 
Later works improve global reasoning by adopting Vision Transformers~\cite{zhao2022monovit,ranftl2021vision} or diffusion priors~\cite{rombach2022high,ke2025marigold,guo2025multi}, but at the cost of substantially increased model complexity and inference latency.

To reduce computational overhead, lightweight architectures have been explored~\cite{zhang2023lite,lyu2021hr,hui2022rm,zhao2022monovit,zhou2024recurrent}. Zhang \etal.~\cite{zhang2023lite} combined CNN and Transformer modules, while Chen \etal.~\cite{chen2026purilight} introduced a lightweight shuffle and purification framework. However, these designs often compromise structural fidelity, particularly around thin structures and object boundaries, and still struggle in dynamic environments. 
Since most self-supervised methods rely on photometric reprojection, moving objects violate static-scene assumptions and degrade supervision quality. 
Existing attempts include motion disentanglement~\cite{watson2021temporal,woo2024prodepth,feng2022disentangling,bangunharcana2023dualrefine}, semantic or flow assistance~\cite{lee2021learning,hui2022rm,sun2023dynamo,li2021unsupervised,nguyen2024mining}, and cross-dataset or pseudo-label training strategies~\cite{yu2025exploiting,luginov2024nimbled,zhang2025hybrid}. 
Although Nimbled~\cite{luginov2024nimbled} enhances supervision via pseudo-label refinement and large-scale video pretraining, these approaches typically introduce additional priors, auxiliary tasks, or complex pipelines, limiting flexibility and deployment efficiency.

Parallelly, supervised foundation models like Metric3D~\cite{yin2023metric3d} and Depth Anything V2~\cite{yang2024depth} have set new benchmarks in zero-shot generalization. However, their reliance on massive curated datasets and heavy backbones limits their deployment in resource-constrained environments. In contrast, self-supervised frameworks offer superior flexibility and environmental adaptability.  Nevertheless, existing lightweight self-supervised designs struggle to maintain structural fidelity, particularly near object boundaries, and exhibit poor robustness in dynamic environments such as driving scenes where moving objects violate the static-camera assumption. Current solutions for handling dynamic scenes typically introduce auxiliary tasks or complex pipelines, which limit their deployment capabilities on edge devices. Thus, improving robustness and scalability within lightweight self-supervised frameworks remains an important complementary direction.

Consequently, a key question arises: \emph{can self-supervised monocular depth estimation narrow the robustness gap with large supervised models while preserving lightweight design and data efficiency?} 
Addressing this requires jointly reconsidering architectural scaling and dynamic-scene handling within a unified optimization framework rather than relying on larger backbones or external supervision.

To this end, we propose FlexDepth, a family of scale-driven self-supervised monocular depth models that balance architectural efficiency with robustness in driving scenarios. Unlike traditional lightweight decoders that rely on simple upsampling methods, we propose a Scale-Driven Decoder (SDD). The SDD adaptively selects its constituent components to enhance feature fusion and yield high-precision depth maps. Furthermore, this paper introduces a two-stage static-dynamic decoupled training strategy to disentangle dynamic driving environments from static backgrounds. By performing separate confidence assessments, our approach achieves superior robustness in highly dynamic environments. Fig. \ref{fig:main_results} illustrates the core contributions of our work. The FlexDepth model family achieves state-of-the-art performance on the KITTI~\cite{6248074} dataset with low computational overhead and reduced inference latency.

Our contributions are summarized as follows:

\begin{itemize}
	\item We propose FlexDepth, a scale-driven family of lightweight self-supervised depth models that systematically study accuracy–efficiency trade-offs while maintaining robustness in dynamic driving environments.
	
	\item We design a two-stage static-dynamic decoupled training strategy that enables independent confidence assessments for static backgrounds and dynamic objects, thereby providing more effective guidance for the training of the depth prediction network.
	
	\item A novel Scale-Driven Decoder (SDD) is introduced to facilitate efficient inference and precise depth mapping, supporting scalable deployment on edge devices according to specific scene requirements.
\end{itemize}

	\section{Related Work}

\subsection{Self-Supervised Monocular Depth Estimation}

Owing to the limited availability of high-quality ground-truth depth data, self-supervised monocular depth estimation has attracted considerable research interest. Garg \etal.~\cite{garg2016unsupervised} first formulated the task as a novel view synthesis problem using stereo imagery. Subsequently, a series of studies have been proposed, such as designing robust loss functions~\cite{zhang2018joint,zhan2018unsupervised,shu2020feature}, incorporating auxiliary information during training~\cite{klodt2018supervising,watson2019self}, and introducing additional geometric constraints~\cite{yang2018unsupervised,li2021structdepth}. Godard \etal.~\cite{godard2019digging} further advanced the field with Monodepth2, which introduced a per-pixel minimum reprojection loss to handle occlusions and a multi-scale sampling strategy to recover finer details. Recently, Zhang et al. ~\cite{zhang2025hybrid} proposed Hybrid-depth, a framework that leverages large-scale multimodal backbones (CLIP~\cite{radford2021learning} and DINO~\cite{caron2021emerging,oquab2023dinov2}) .

\subsection{Efficient Architecture Design}

Building upon the advancements in self-supervised monocular depth estimation, a growing research direction has focused on developing lightweight models~\cite{zhang2023lite,wang2025mg,cheng2024tinydepth,zhou2024recurrent} to overcome the computational bottlenecks associated with deploying these systems on resource-constrained edge devices. Lyu \etal.~\cite{lyu2021hr} adopted MobileNetV3 and a Feature Fusion Squeeze-Excitation module, while Zhou \etal.~\cite{zhou2024recurrent} introduced R-MSFM with a parameter-sharing decoder to cut training parameters. Zhang \etal.~\cite{zhang2023lite} combined CNN and Transformer using dilated convolution and cross-channel covariance attention. Chen \etal.~\cite{chen2026purilight} implemented global feature purification by operating in the frequency domain. However, these models often sacrifice structural accuracy or incur high computational costs, failing to balance performance and lightweight design. Moreover, they struggle in dynamic environments, thereby limiting their robustness in autonomous driving scenarios.

\subsection{Adaptation to Dynamic Environment}

Constrained by reprojection loss, mitigating the impact of dynamic objects on model training has become a challenge in self-supervised approaches. To address this issue, Monodepth2~\cite{godard2019digging} introduced an auto-masking mechanism, which, however, is only effective for objects moving at the same velocity as the observer. Other methods integrate depth estimation with semantic segmentation~\cite{feng2022disentangling,moon2024ground} or optical flow ~\cite{lee2021learning,hui2022rm,sun2023dynamo,li2021unsupervised,nguyen2024mining}, but they struggle in unfamiliar scenes and cannot eliminate side effects like shadows or motion blur. Recently, Zhang \etal.~\cite{zhang2025depth} proposed the DSI framework to address these issues, yet its reliance on empirical hyperparameters restricts generalization. Furthermore, most of these approaches rely heavily on computationally expensive backbones and external priors, which hinders their efficient deployment on resource-constrained edge devices.

\begin{figure*}[t]
	\centering
	
	\includegraphics[width=\textwidth]{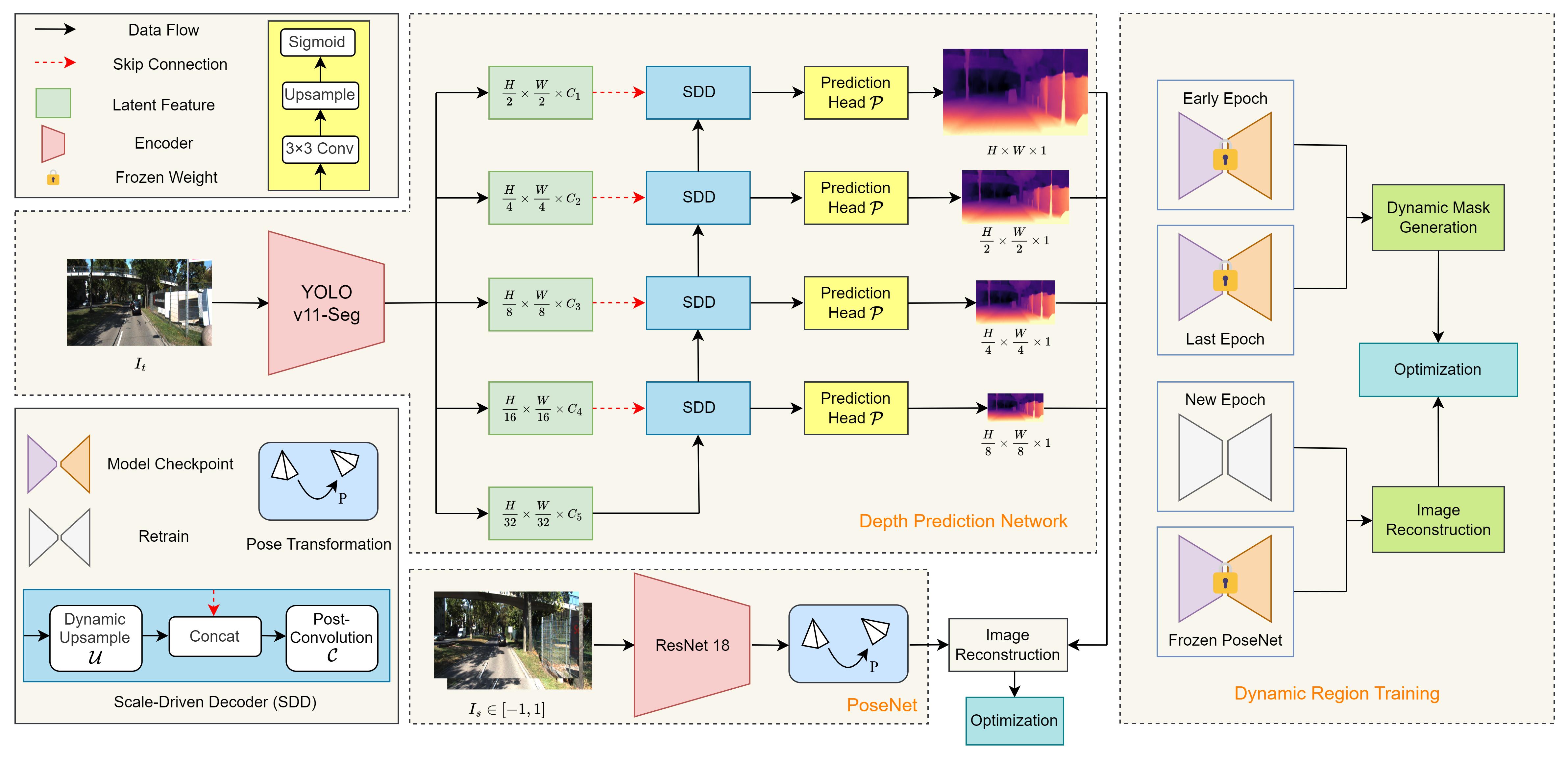}
	\caption{\textbf{FlexDepth system overview.} FlexDepth implements a two-stage training pipeline. In the first stage, both the Depth Prediction Network and the PoseNet are optimized simultaneously. The second stage focuses on dynamic scene refinement, where motion-induced epoch-wise inconsistency is leveraged to generate dynamic object masks. Subsequently, the Depth Prediction Network is optimized using the frozen PoseNet.  }
	\label{fig:system_overview}
	
\end{figure*}

	\section{Method}
\label{sec:framework}

This paper introduces a scale-driven family of flexible self-supervised monocular depth estimation models, termed FlexDepth, designed for dynamic driving environments. Fig. \ref{fig:system_overview} shows the proposed architecture. The FlexDepth model family comprises five models across varying scales—Nano, Small, Medium, Large, and X-Large—tailored for flexible edge-side deployment based on specific resource constraints. Central to this architecture is our Scale-Driven Decoder (SDD), which adaptively selects specific modules and training pipelines corresponding to each model scale. Furthermore, the training process is structured into two distinct stages: the first stage optimizes the pose and depth networks to establish a lightweight yet efficient framework. In the second stage, a simple yet effective static-dynamic decoupled mask is introduced to facilitate the retraining of the depth network. This decoupling enhances the model's representational power in dynamic driving scenes, ultimately yielding superior robustness and environmental adaptability without compromising efficiency.


\subsection{Motivation and Analysis}
\label{subsec:motivation}

In the field of lightweight self-supervised monocular depth estimation, research has primarily focused on encoder design~\cite{godard2019digging,zhang2023lite,chen2026purilight}, while decoders are often marginalized with simplified feature fusion and upsampling methods. Although encoders can extract rich representations, the inductive bias inherent in U-Net-like architectures acts as a bottleneck, limiting the transformation of low-resolution abstract features into precise depth maps. We argue that the decoder is equally crucial for overall performance gain. To this end, we revisit decoder construction and propose the Scale-Driven Decoder (SDD), which dynamically selects model components and training strategies based on varying model scales. Furthermore, from the perspectives of upsampling and feature integration, we demonstrate that models of different scales exhibit distinct dependencies. This highlights the rationality and superiority of SDD compared to single-scale, fixed-component architectures. Our approach ultimately achieves real-time inference of high-precision depth maps.

The processing of dynamic scenes has emerged as a prominent research area in recent years. However, most existing models~\cite{sun2023dynamo,woo2024prodepth,moon2024ground,nguyen2024mining} for handling dynamic environments exhibit high computational complexity or require extensive prior knowledge, posing significant challenges for deployment in resource-constrained settings. By adopting a two-stage training approach complemented by a simple yet efficient static-dynamic decoupled mask, our method evaluates confidence levels for static and dynamic regions separately, enabling precise distinction between them.

\subsection{Depth Prediction Network}
\label{subsec:encoder}
\textbf{Depth Encoder.} We adopt a standard multi-scale feature extraction architecture based on YOLOv11~\cite{Jocher_Ultralytics_YOLO_2023} as the depth encoder. Given an input image $I \in \mathbb{R}^{H \times W \times 3}$, the encoder progressively generates five feature maps through successive downsampling operations, with spatial resolutions ranging from $\frac{1}{2}$ to $\frac{1}{32}$ of the original input size.

\textbf{Depth Decoder.} Conventional approaches~\cite{godard2019digging,zhang2023lite,chen2026purilight} typically follow a pipeline of ``pre-convolution $\rightarrow$ upsampling $\rightarrow$ concatenation $\rightarrow$ post-convolution''. Most lightweight tasks often prioritize this pipeline, which applies convolution kernels prematurely before scale restoration to reduce computational overhead. However, this forces the compression of already sparse global semantic information, leading to the irreversible loss of critical depth cues before spatial expansion occurs. This typically manifests as blurred object boundaries in depth maps and a failure to predict small-scale objects. We rethink this fixed pipeline in lightweight networks, maintaining that the pursuit of efficiency should not come at the cost of global semantic information in deep features. Consequently, we remove the pre-convolution stage entirely, shifting our design focus toward upsampling methods $\mathcal{U}$, post-convolution $\mathcal{C}$, and the prediction head $\mathcal{P}$. The depth map $D$ generation process in our revised pipeline can be formulated as follows:  
\begin{equation}
	D_i = \mathcal{P} \{ \mathcal{C}(\text{Concat}[\mathcal{U}(\mathbf{F}_{i+1}), \mathbf{F}_{enc,i}]) \},
\end{equation}
where $\mathbf{F}_{i+1}$ is the $(i+1)$-th layer decoder feature, and $\mathbf{F}_{enc,i}$ is the $i$-th layer encoder. They are concatenated via a skip connection. The implementation of the FlexDepth model across different scales varies specifically in terms of $\mathcal{C}$, $\mathcal{U}$, and $\mathcal{P}$. We begin by introducing these three key components.

\textbf{Post-Convolution $\mathcal{C}$.} Since the pre-convolution module has been removed, the design of the post-convolution module becomes particularly crucial. Drawing inspiration from~\cite{Jocher_Ultralytics_YOLO_2023}, we designed a combination of two types of Bottleneck units (B): the High-Efficiency Bottleneck (HEB) and the High-Performance Bottleneck (HPB). The output of each HEB layer is concatenated into the final output. This dense connection shortens the path for gradient backpropagation, making the model easier to train. HEB is particularly suitable for small scale models that prioritize gradient flow richness and stability. Given the input feature $\mathbf{X}$ and the uniform split operator $\mathcal{S}$, the output $\mathbf{X}_{out}$ can be represented as:

\begin{equation}
	[f_a, f_b] = \mathcal{S}(Conv(X)),
\end{equation}
\begin{equation}
	F_1 = B_1(f_a),F_2 = B_2(F_1),\cdot\cdot\cdot,F_n = B_n(F_{n-1}),
\end{equation}
\begin{equation}
	X_{out} = Conv(Concat[f_a, f_b,F_1,\cdot\cdot\cdot,F_n]).
\end{equation}
On the other hand, the core logic of HPB is similar to that of HEB. The primary distinction lies in the replacement of the standard Bottleneck unit $B$ with a bottleneck block $B^k$ featuring a Cross Stage Partial (CSP) structure. Furthermore, the feature concatenation logic places greater emphasis on modular CSP fusion. This approach offers the advantage of capturing finer-grained local features, thereby enhancing model performance without significantly increasing the computational overhead. The process of $B^k$ is illustrated as follows:
\begin{equation}
	B^k(x) = Conv(Concat[ \prod_{j=1}^n B^{'}_j(Conv(x)),Conv(x)]),
\end{equation}
where $\prod$ denotes the sequential composition of bottleneck layers. To distinguish from the outer block $B_i$, $B'_j$ represents the minimal convolutional unit within the internal CSP structure.

\textbf{Upsampling Operation $\mathcal{U}$.} Inspired by~\cite{liu2023learning}, we develop a dynamic upsampling approach tailored for models across various scales. Traditional lightweight upsampling methods based on simple bilinear interpolation are inherently content-agnostic; they apply fixed upsampling weights to every region of an image, regardless of whether it is a smooth background or a complex edge, which frequently results in blurred boundaries. Furthermore, as a form of simple weighted averaging, bilinear interpolation exerts a low-pass filtering effect that suppresses high-frequency information, leading to halo artifacts or blurring at object contours. To address these limitations, we implement a two-stage process for each scale: offset generation and dynamic resampling. By leveraging a convolutional layer to predict upsampling offsets $\Delta P$ dynamically, the model can adaptively determine the optimal sampling locations from the original feature map based on semantic content, such as object silhouettes. By integrating a sampling grid with these predicted offsets, we achieve non-uniform upsampling, thereby generating sharper and more accurate structural boundaries. The general procedure for the dynamic upsampling operation $\mathcal{U}$ is formulated as:
\begin{equation}
	\mathcal{U}(X)  = GS(X, PS ( \frac{2 \cdot (\mathbf{C} + \Delta \mathbf{P})}{\text{Norm}} - 1, s )),
\end{equation}
where $\mathbf{C}$ denotes the uniform sampling grid coordinates, $\text{Norm}$ represents the normalization factor, and $s$ is the upsampling scale factor. $PS$ refers to the Pixel Shuffle operator, which transforms the predicted offset information into the spatial coordinate dimension. $GS$ stands for the Grid Sample function, which performs bilinear interpolation based on the calculated coordinates.

It is worth noting that for larger models, we experimentally employ an inverted upsampling prediction head $\mathcal{P}$, which reverses the traditional sequence of prediction followed by upsampling. The rationale is that high-channel features maintain structural and semantic consistency after upsampling, thereby enabling subsequent prediction convolutions to generate high-quality outputs more accurately. Conversely, smaller models adhere to the traditional paradigm to preserve real-time inference speeds on resource-constrained devices.

In summary, FlexDepth (Nano/Small) utilize a Post-Convolution $\mathcal{C}$ based on HEB, the dynamic upsampling operator $\mathcal{U}$, and a traditional prediction head $\mathcal{P}$. Conversely, the Medium, Large, and X-Large versions adopt a Post-Convolution $\mathcal{C}$ based on HPB, the dynamic upsampling operator $\mathcal{U}$, and an inverted prediction head $\mathcal{P}$.

\subsection{Two-Stage Static-Dynamic Decoupled Training}
\label{subsec:dsi}

Recent research in dynamic environments has demonstrated that dynamic regions exhibit significantly higher predictive instability during the training process. Despite this observation, most existing methodologies rely heavily on complex network architectures and extensive prior knowledge, which inherently limits their potential for lightweight deployment. Inspired by the work of Zhang et al. \cite{zhang2025depth}, we propose a static-dynamic decoupling approach. A fundamental limitation of \cite{zhang2025depth} lies in its reliance on a fixed threshold for region differentiation. In real-world applications, the proportion of dynamic objects varies significantly across different scenes and individual frames. Consequently, methodologies that depend on dataset-wide statistics lack the necessary robustness to generalize across diverse and unpredictable environments. We employ a two-stage training strategy to achieve robustness in dynamic scenes without introducing additional architectural complexity or requiring prior environmental knowledge. At its core, our method focuses on the retraining of the depth prediction network to better handle temporal inconsistencies. Specifically, the training instability of dynamic regions is characterized by the discrepancy between the model's outputs during the early and late stages of training. While the predicted depth of static backgrounds remains relatively consistent throughout the training process, dynamic regions exhibit significant fluctuations. To address this, we designed the Static-Dynamic Decoupled Mask to effectively decouple the static and dynamic components of the environment, thereby enhancing the model's overall stability. 

In the initial stage, we simultaneously train the depth prediction network and the pose estimation network. However, due to the inherent predictive instability of moving objects, their depth outputs exhibit significant variance between early and late training epochs, whereas static backgrounds remain relatively consistent. To leverage this instability for decoupling, we compute disparity maps using weights from the initial ($disp_i$) and final ($disp_l$) epochs of the first stage. The core of the second-stage training involves constructing a static-dynamic decoupled mask $M$ to recalibrate the confidence assessments for dynamic objects and the static background:
\begin{equation}
	M = [ |\text{disp}_i^\beta - \text{disp}_l^\beta| < g(\phi(f_{s_i}, f_{s_l})) ],
\end{equation}
where $\beta$ is set to 0.7, $ \lbrack \cdot \rbrack$ means the Iverson bracket. $g(\phi(f_{s_i}, f_{s_l}))$ serves as an adaptive thresholding function. Specifically, $\phi$ measures the structural evolution between early-stage ($f_{s_i}$) and late-stage ($f_{s_l}$) feature representations, while $g$ (See Appendix) maps these variances into a pixel-wise threshold space. This mechanism allows the network to dynamically adjust its sensitivity: in regions with high semantic complexity or rapid motion, the threshold adapts to ensure that only stable, static pixels contribute to the subsequent confidence evaluation. Our method essentially conducts a refined retraining of the depth network, achieving superior robustness in highly heterogeneous dynamic environments without any additional inference overhead.

\subsection{Pose Network and Self-Supervised Training}
\label{subsec:selfsupervised}

Following the self-supervised paradigm~\cite{godard2019digging,zhou2017unsupervised}, we jointly optimize the depth and pose networks through photometric consistency without ground truth supervision. Concretely, our PoseNet adopts a ResNet-18 encoder with a four-layer decoder to predict the relative 6-DoF pose transformation $P_{t \to s}$ between adjacent frames.

To train the networks, we warp the source image $I_s$ into the target view to generate the reconstructed image $\hat{I}_t$ according to the predicted depth $\hat{D}_t$ and pose $P_{t \to s}$:
\begin{equation}
	\hat{I}_t = I_s \langle \text{proj}(\hat{D}_t, P_{t \to s}, K) \rangle,
\end{equation}
where $\langle \cdot \rangle$ denotes the differentiable bilinear sampling operator following ~\cite{godard2019digging}. The photometric error is formulated as:
\begin{equation}
	L_p = \alpha \frac{1 - \text{SSIM}(\hat{I}_t, I_t)}{2} + (1-\alpha) \| \hat{I}_t - I_t \|_1,
\end{equation}
where $\alpha = 0.85$ by default. To handle occlusions and moving objects, we employ the minimum reprojection loss with auto-masking~\cite{godard2019digging}:
\begin{equation}
	L_r = \mu \cdot \min_{s} L_p(\hat{I}_t, I_t),
\end{equation}
where $\mu$ excludes pixels with inconsistent photometric errors. We also adopt the edge-aware smoothness loss ~\cite{godard2017unsupervised} to regularize the predicted depth:
\begin{equation}
	L_{\text{smooth}} = |\partial_x d_t^*| e^{-|\partial_x I_t|} + |\partial_y d_t^*| e^{-|\partial_y I_t|},
\end{equation}
where $d_t^*$ denotes the mean-normalized inverse depth. The total loss for the first training stage is:
\begin{equation}
	L_{raw} = \frac{1}{4} \sum_{s \in \{1, \frac{1}{2}, \frac{1}{4}, \frac{1}{8}\}} (L_r + \lambda L_{\text{smooth}}),
\end{equation}
where $\lambda = 0.001$ by default.

In the second stage, with the static-dynamic decoupled mask $M$ obtained from Sec.~\ref{subsec:dsi}, we incorporate the masked normal distribution loss~\cite{zhang2025depth}:
\begin{equation}
	L_m = M \cdot \left[ \log(\sigma + 1) + \frac{\lambda L_r^2}{2\sigma^2} \right],
\end{equation}
where $\lambda = 0.5$ and $\sigma$ is predicted by a variance network. We further apply geometric depth smoothing~\cite{zhang2025depth}:
\begin{equation}
	L_{\text{GDS}} = |\partial_x \hat{d}_t| e^{-|\partial_x I_t|} + [\eta(1 - M) + M] |\partial_y \hat{d}_t| e^{-|\partial_y I_t|},
\end{equation}
with $\eta = 100$. The final loss is formulated as $L_{\text{total}} = \frac{1}{HW} \sum_t (L_m + \gamma L_{\text{GDS}})$.

\begin{table*}[h]
	\centering
	
	\caption{Quantitative Results on the KITTI~\cite{6248074}  (Eigen split ~\cite{eigen2015predicting}) and Cityscapes~\cite{cordts2016cityscapes} datasets. The input resolutions are $640\times192$ for KITTI and $416\times128$ for Cityscapes. \textbf{M}: monocular videos, \textbf{M+Se}: monocular videos + semantic segmentation,  \textbf{T.F.}: test input frames, \textbf{Bold} denotes the best result. }
    \label{tab:results_kitti_cityscapes}

	\centering
    
    \resizebox{\textwidth}{!}{
	\setlength{\tabcolsep}{2pt}
	{ 
		\begin{tabular}{ccccccccccccc}
			\toprule
			\multirow{2}{*}{\textbf{D}} &
			\multicolumn{1}{c}{\multirow{2}{*}{\textbf{Method}}} &
            \multicolumn{1}{c}{\multirow{2}{*}{\textbf{Year}}} &
			\multicolumn{1}{c}{\multirow{2}{*}{\textbf{T.F.}}} &
			\multicolumn{1}{c}{\multirow{2}{*}{\textbf{Data}}} &
			\multicolumn{4}{c}{\textbf{Depth Error$\downarrow$}} &
			\multicolumn{3}{c}{\textbf{Depth Accuracy$\uparrow$}} &
			\multicolumn{1}{c}{\multirow{2}{*}{\textbf{FLOPs.(G)$\downarrow$}}} 
			 \\
			&
			\multicolumn{1}{c}{} &
			\multicolumn{1}{c}{} &
            \multicolumn{1}{c}{} &
			\multicolumn{1}{c}{} &
			\multicolumn{1}{c}{Abs Rel} &
			\multicolumn{1}{c}{Sq Rel} &
			\multicolumn{1}{c}{RMSE} &
			\multicolumn{1}{c}{RMSE log} &
			\multicolumn{1}{c}{$\delta < 1.25$} &
			\multicolumn{1}{c}{$\delta < 1.25^2$} &
			\multicolumn{1}{c}{$\delta < 1.25^3$} &
			
			\\
			\midrule 
			\multirow{16}{*}{\rotatebox[origin=c]{90}{KITTI}} &
			Monodepth2\cite{godard2019digging} &
			2019 &
			1 &
			M &
			0.115 &
			0.903 &
			4.863 &
			0.193 &
			0.877 &
			0.959 &
			0.981 &
			8.0 \\
			&
			Lite-Mono-Tiny\cite{zhang2023lite} &
			2023 &
			1 &
			M &
			0.110 &
			0.837 &
			4.710 &
			0.187 &
			0.880 &
			0.960 &
			0.982 &
			2.8 \\
			&
			Lite-Mono-Small\cite{zhang2023lite} &
			2023 &
			1 &
			M &
			0.110 &
			0.802 &
			4.671 &
			0.186 &
			0.879 &
			0.961 &
			0.982 &
			4.7  \\
			&
			Lite-Mono \cite{zhang2023lite}&
			2023 &
			1 &
			M &
			0.107 &
			0.765 &
			4.561 &
			0.183 &
			0.886 &
			0.963 &
			0.983 &
			5.0  \\
            &
            PuriLight \cite{chen2026purilight} &
            2026 &
            1 &
            M &
            0.106 &
            0.747 &
            4.536 &
            0.182 &
            \textbf{0.890} &
            \textbf{0.964} &
            \textbf{0.984} &
            7.1 \\
			&
			\textbf{Flex-Nano (Ours)} &
			2026 &
			1 &
			M &
			0.110 &
			0.794 &
			4.678 &
			0.184 &
			0.878 &
			0.961 &
			0.983 &
			\textbf{0.7} \\
			&
			\textbf{Flex-Small (Ours)} &
				2026 &
			1 &
			M &
			\textbf{0.104} &
			\textbf{0.713} &
			\textbf{4.458} &
			\textbf{0.179} &
			\textbf{0.890} &
			\textbf{0.964} &
			0.983 &
			2.8  \\
			\cline{2-13}
			
			&
			Lite-Mono-8M\cite{zhang2023lite} &
				2023 &
			1 &
			M &
			0.101 &
			0.729 &
			4.454 &
			0.178 &
			0.897 &
			0.965 &
			0.983 &
			11.2 \\
                &
                
			GeoDepth\cite{wu2025geodepth} &
				2025 &
			1 &
			M &
			0.100 &
			0.694 &
			4.381 &
			0.176 &
			0.897 &
			0.966 &
			0.984 &
			11.9 \\
            
			&
			TinyDepth\cite{cheng2024tinydepth} &
				2024 &
			1 &
			M &
			0.096 &
			0.665 &
			4.249 &
			\textbf{0.171} &
			0.904 &
			\textbf{0.968} &
			\textbf{0.985} &
			13.3  \\
			
			&
			\textbf{Flex-Medium (Ours)} &
				2026 &
			1 &
			M &
			0.096 &
			\textbf{0.639} &
			4.253 &
			0.172 &
			0.903 &
			\textbf{0.968} &
			\textbf{0.985} &
			\textbf{10.0}  \\
			&
			\textbf{Flex-Large (Ours)} &
				2026 &
			1 &
			M &
			\textbf{0.095} &
			0.642 &
			\textbf{4.199} &
			\textbf{0.171} &
			\textbf{0.906} &
			\textbf{0.968}  &
			0.984 &
			11.5 \\
			\cline{2-13}
			
			&
			MonoVit\cite{zhao2022monovit} &
				2022 &
			1 &
			M &
			0.099 &
			0.708 &
			4.372 &
			0.175 &
			0.900 &
			0.967 &
			0.984 &
			59.7  \\
			&
			
			FGTO \cite{moon2024ground} &
				2024 &
			1 &
			M+Se &
			0.096 &
			0.696 &
			4.327 &
			0.174 &
			0.904 &
			0.968 &
			{\textbf {0.985}} &
			59.7  \\
			&
			
			DSI-MonoViT\cite{zhang2025depth} &
				2025 &
			1 &
			M &
			0.096 &
			0.711 &
			4.321 &
			0.172 &
			0.905 &
			0.968 &
			0.984 &
			59.7  \\
			 &
			Self-Distillation\cite{yu2025exploiting}&
            	2025 &
			1 &
			M &
			0.095 &
			0.619&
			4.156 &
			0.169 &
			0.905 &
			\textbf{0.969} &
			\textbf{0.985} &
			64.5 \\
			&
			
			\textbf{Flex-X-Large (Ours)} &
				2026 &
			1 &
			M &
			\textbf{0.093} &
			\textbf{0.605} &
			\textbf{4.114} &
			\textbf{0.167} &
			\textbf{0.910} &
			\textbf{0.969} &
			\textbf{0.985} &
			\textbf{24.6} \\
			\midrule 
			\multirow{10}{*}{\rotatebox[origin=c]{90}{Cityscapes}}
                &
                
			ManyDepth\cite{watson2021temporal} &
				2021 &
			2(-1,0) &
			M &
			0.114 &
			1.193 &
			6.223 &
			0.170 &
			0.875 &
			0.967 &
			0.989 &
			12.1  \\
			&
			DynamicDepth\cite{feng2022disentangling} &
			2022 &
			2(-1,0) &
			M+Se &
			0.103 &
			1.000 &
			5.867 &
			0.157 &
			0.895 &
			0.974 &
			0.991 &
			18.5+  \\
            &
            PuriLight \cite{chen2026purilight} &
            2026 &
            1 &
            M &
            0.099 &
            1.014 &
            5.892 &
            0.153 &
            0.901 &
            0.976 &
            0.992 &
            5.7 \\
			&
            
			\textbf{Flex-Nano (Ours)} &
			2026 &
			1 &
			M &
			0.107 &
			1.261 &
			6.133 &
			0.164 &
			0.893 &
			0.971 &
			0.989 &
			\textbf{0.6} \\
			&
			\textbf{Flex-Small (Ours)} &
			2026 &
			1 &
			M &
			0.100 &
			1.078 &
			5.813 &
			0.153 &
			0.904 &
			0.975 &
			0.991 &
			2.2  \\
			&
			\textbf{Flex-Medium (Ours)} &
			2026 &
			1 &
			M &
			0.089 &
			\textbf{0.885} &
			5.358 &
			0.143 &
			0.917 &
			0.979 &
			\textbf{0.993} &
			8.0  \\
			&
			\textbf{Flex-Large (Ours)} &
			2026 &
			1 &
			M & \textbf{0.087} &	0.911&	\textbf{5.310}	& \textbf{0.139}	& \textbf{0.924}	& \textbf{0.981} & \textbf{0.993} &9.2 \\
			\cline{2-13}
			&
			ProDepth\cite{woo2024prodepth} &
			2024 &
			2(-1,0) &
			M &
			0.095 &
			0.876 &
			5.531 &
			0.146 &
			0.908 &
			0.978 &
			\textbf{0.993} &
			35.5  \\
				&
			
			MonoViT\cite{zhao2022monovit} &
			2022 &
			1 &
			M &
			0.114 &
			1.238 &
			6.589 &
			0.174 &
			0.860 &
			0.965 &
			0.990 &
			47.7 \\
				&
			
			FGTO\cite{moon2024ground} &
			2024 &
			1 &
			M+Se &
			0.096 &
			 0.930 &
			 5.806 &
			0.152 &
			0.905 &
			0.976 &
			0.992 &
			47.7 \\
			 &
			Self-Distillation\cite{yu2025exploiting}&
            2025 &
			1 &
			M &
			0.118 &
			1.091 &
			6.076 &
			0.173 &
			0.854 &
			0.960 &
			0.988 &
			47.7+ \\
			
			&
			DSI-MonoViT\cite{zhang2025depth} &
			2025 &
			1 &
			M &
			0.088 &
			\textbf{0.872} &
			5.410 &
			0.141 &
			0.918 &
			0.981 &
			\textbf{0.993} &
			47.7  \\

			&
			\textbf{Flex-X-Large (Ours)} &
			2026 &
			1 &
			M &\textbf{0.086}&	0.877 &	\textbf{5.268}&	\textbf{0.137}&	\textbf{0.926}&	\textbf{0.982}&	\textbf{0.993}
			&\textbf{19.6}
		\\
			\bottomrule
		\end{tabular}
		
	}
    }

\end{table*}

\section{Experiments}
\label{sec:exper}

\subsection{Datasets}
\label{subsec:datasets}

Following standard evaluation protocols, we validate our proposed method across three benchmark datasets. Specifically, the KITTI~\cite{6248074} and Cityscapes~\cite{cordts2016cityscapes} datasets are utilized as domain-specific driving scenarios to evaluate the performance of our model within dynamic urban environments. To assess the generalization capabilities of FlexDepth, we conduct zero-shot evaluation on the Make3D dataset~\cite{saxena2008make3d}. It is important to note that all training is performed using unlabeled video frames exclusively. Our approach does not rely on any auxiliary data, such as semantic labels, optical flow, or textual descriptions, ensuring an entirely self-supervised learning pipeline.

\textbf{KITTI~\cite{6248074} Dataset}. We adhere to the Eigen split~\cite{eigen2015predicting} following established practices~\cite{zhou2017unsupervised,godard2019digging}, with 39,180 training images, 4,424 validation images, and 697 test images at a resolution of 640$\times$192.

\textbf{Cityscapes~\cite{cordts2016cityscapes} Dataset}. This dataset comprises a variety of dynamic scenarios and functions as the principal benchmark for evaluating dynamic models. We use 58,335 training images and 1,525 testing images provided by~\cite{feng2022disentangling}  at a resolution of 416$\times$128.  

\textbf{Make3D~\cite{saxena2008make3d} Dataset}. It comprises 134 test images of outdoor scenes, commonly used to evaluate MDE generalization. To assess our method's generalization capability, we directly apply the KITTI-trained model (640$\times$192)  to this dataset and report detailed evaluation metrics.
\subsection{Specific Details in Training}
\label{subsec:implementation} 

We train all models using the NAdam optimizer~\cite{dozat2016incorporating}. Both datasets undergo a two-stage training process. In the first stage, the depth prediction network and PoseNet are trained. The second stage focuses on training the dynamic region processing capability. For the KITTI~\cite{6248074} dataset, the two stages are trained for 30 and 20 epochs respectively, while for the Cityscapes dataset, they are trained for 30 and 10 epochs respectively. We employ distinct learning rate schedulers for each phase: Stage 1 utilizes a step scheduler with a step size of 15 and a decay factor of 0.1, whereas Stage 2 adopts an ExponentialLR scheduler with a decay factor of 0.8. For the Nano/Small models with a batch size of 12, the initial learning rates for the encoder and decoder are set to 5e-5 and 1e-4, respectively. For the Medium/Large/X-Large models with a batch size of 6, both initial learning rates are set to 5e-5. All experiments are conducted using NVIDIA GPUs (RTX 4090 and RTX 2080 Ti). On the RTX 4090, our largest model (X-Large) requires approximately 10 hours to complete 50 epochs on the KITTI~\cite{6248074} dataset.
\begin{figure*}[t]
	\centering
	
	\includegraphics[width=\textwidth]{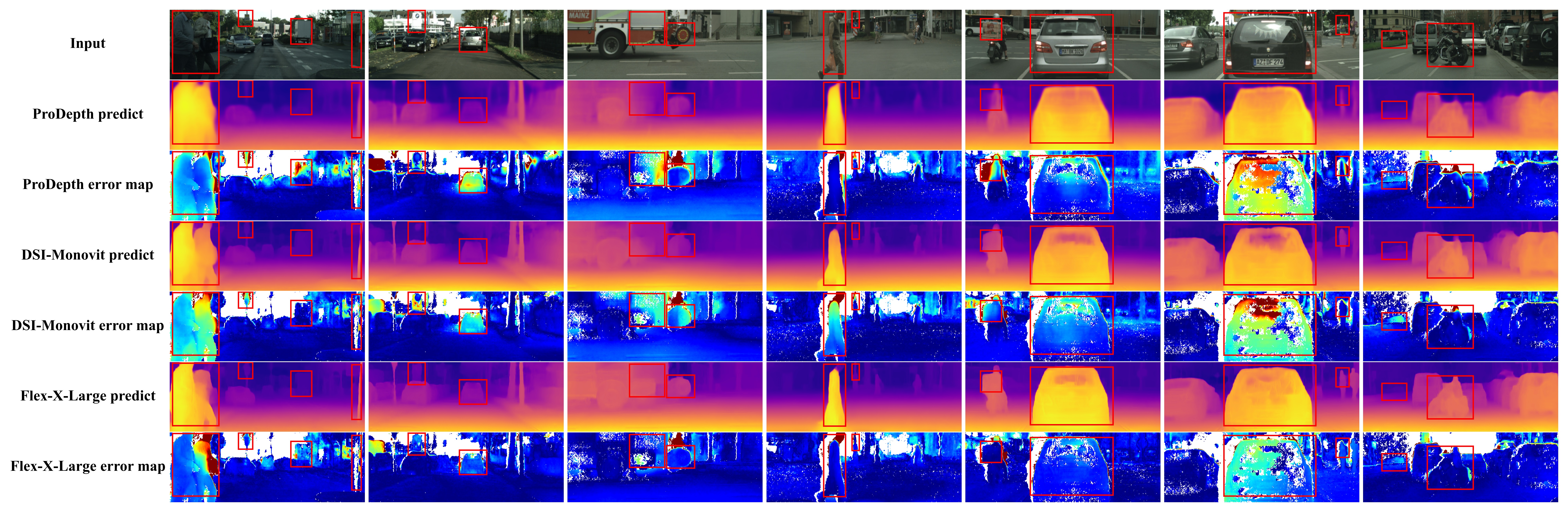}
	\caption{\textbf{Qualitative evaluation on Cityscapes~\cite{cordts2016cityscapes}.} Our method is compared to dynamic scene handlers DSI~\cite{zhang2025depth} and ProDepth~\cite{woo2024prodepth}. In the error maps, blue indicates smaller errors and brighter red indicates larger ones. Our FlexDepth framework consistently yields more accurate depth with significantly reduced errors.}
	\label{fig:cityscapes_comparison}
\end{figure*}
\subsection{Experiment Results}
\label{sec:comparison}
\begin{figure*}[t]
	\centering
	\includegraphics[width=\textwidth]{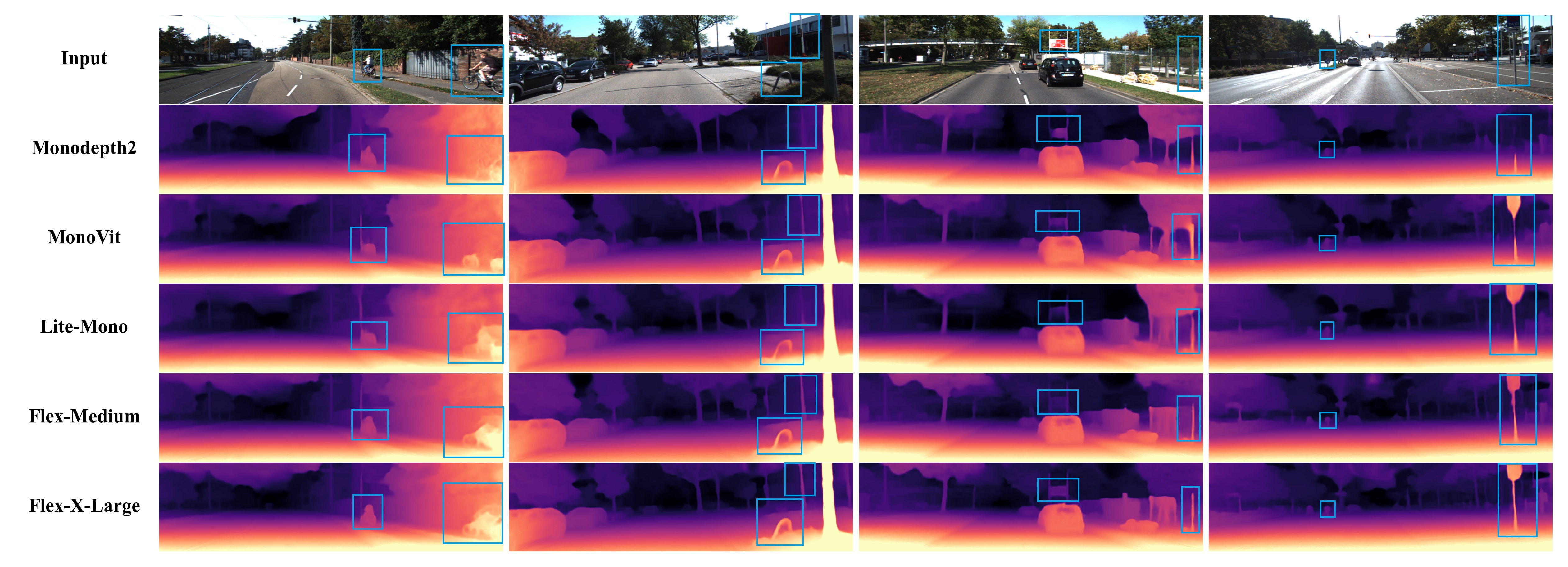}
	\caption{\textbf{Qualitative comparison on KITTI~\cite{6248074}.} Our method is compared with Monodepth2 \cite{godard2019digging}, MonoVit \cite{zhao2022monovit}, and Lite-Mono \cite{zhang2023lite}. The proposed FlexDepth demonstrates superior performance in preserving fine details and structural integrity, particularly in challenging scenarios with complex object boundaries and textureless regions.}
	\label{fig:kitti_comparison}
\end{figure*}

Table \ref{tab:results_kitti_cityscapes} presents the quantitative results on the KITTI~\cite{6248074} and Cityscapes~\cite{cordts2016cityscapes} datasets. The evaluation metrics adopted are the seven commonly used indicators proposed in \cite{eigen2014depth}, namely Abs Rel, Sq Rel, RMSE, RMSE log, $\delta < 1.25$, $\delta < 1.25^2$, and $\delta < 1.25^3$. We compare representative multi-scale models ranging from lightweight to heavy configurations. For the KITTI~\cite{6248074} dataset, models are categorized into three scales (0–9.9, 10.0–20.0, and above 20.0 GFLOPs). Since methods on the Cityscapes dataset are generally more complex, we define two scales (0–19.0 and above 19.0 GFLOPs). As shown in the Table \ref{tab:results_kitti_cityscapes}, our approach achieves the lowest computational cost and state-of-the-art performance across all scales. Our smallest model, Flex-Nano, with only 0.7G FLOPs, delivers superior results compared to Lite-Mono~\cite{zhang2023lite}'s Tiny (2.8G) and Small (4.7G) versions. In the large-model category, our Flex-X-Large sets new state-of-the-art performance on both datasets. 

The qualitative results on the two datasets are presented in Fig. \ref{fig:cityscapes_comparison}  and Fig. \ref{fig:kitti_comparison}, respectively. It can be observed that Flex-Medium and Flex-X-Large exhibit extensive receptive fields along with finer object contours and detailed textures. Particularly in highly dynamic scenarios, our model demonstrates strong adaptability, highlighting its superiority. As summarized in Table~\ref{tab_dyn}, we conduct a decoupled evaluation of dynamic regions and static backgrounds. The results demonstrate that, compared to existing methods relying on heavy backbones or complex network architectures, our Flex-X-Large variant achieves a substantial performance boost in dynamic areas while maintaining the lowest computational overhead. Notably, our approach achieves these results without any reliance on auxiliary information, such as pixel-wise motion fields or pseudo-depth maps. This demonstrates the inherent superiority of our architecture in handling complex, non-rigid scenes through a purely self-supervised pipeline. We provide a comparison with the foundation model in the Appendix.

\textbf{Zero-Shot Cross-Dataset Generalization.} To evaluate the zero-shot generalization capability, we conducted a cross-dataset evaluation on the Make3D~\cite{saxena2008make3d} benchmark using a model trained exclusively on the KITTI~\cite{6248074} dataset. As shown in Table \ref{tab:make3d}, FlexDepth demonstrates superior cross-dataset generalization. This exceptional transfer learning performance underscores the robustness and domain-invariant characteristics of the representations learned by our model, validating its potential applicability across diverse real-world scenarios.

\begin{table*}[htbp]
	\caption{Ablation experiments on model architectures. Legend: $\mathcal{C}$: \checkmark=HPB, $\circ$=HEB. $\mathcal{U}$: \checkmark=dynamic upsampling, $\circ$=bilinear. $\mathcal{P}$: \checkmark=inverted head, $\circ$=conventional head. Dynamic: \checkmark=static-dynamic decoupled mask $M$, $\circ$=fixed threshold, $\times$=w/o stage-2. F.=FLOPs, P.=Params.}
	\label{tab:ablation_decoder}
	\centering
    \resizebox{\textwidth}{!}{
	
	\setlength{\tabcolsep}{2pt}
	\begin{tabular}{cccccccccccccccccc}
		\toprule
		\multirow{2}{*}{ID} &
		
		\multicolumn{3}{c}{Decoder} &
		\multirow{2}{*}{Dynamic} &
		\multirow{2}{*}{F.(G)} &
		\multirow{2}{*}{P.(M)} &
		\multicolumn{4}{c}{Depth Error $\downarrow$} &
		\multicolumn{3}{l}{Depth Accuracy $\uparrow$} \\ 
		& $\mathcal{C}$  &
		$\mathcal{U}$ &
		$\mathcal{P}$  &
		 &
		&
		&
		\multicolumn{1}{c}{Abs Rel} &
		\multicolumn{1}{c}{Sq Rel} &
		\multicolumn{1}{c}{RMSE} &
		\multicolumn{1}{c}{RMSE log} &
		$\delta < 1.25 $ &
        $\delta < 1.25^2$ &
        $\delta < 1.25^3$
        \\ \midrule
		
		1 & $\circ$ & \checkmark &\checkmark & \checkmark & \textbf{24.0} & \textbf{32.0} & \textbf{0.093} & 0.614 & 4.188 & 0.169 & 0.907 &   \textbf{0.969} & \textbf{0.985} \\
		2 & \checkmark & $\circ$ & \checkmark & \checkmark & 24.3 & 32.3 & 0.094  &   0.627  &   4.204  &   0.169  & 0.906  &   0.968  &   \textbf{0.985}
		\\
		3 & \checkmark & \checkmark &$\circ$ & \checkmark & 24.3 & 32.3 & 0.094 &	0.631 &	4.205 &	0.170 &	0.906 & 0.968 & \textbf{0.985}
		\\
		4  & \checkmark & \checkmark & \checkmark & $\times$ & 24.6 & 32.3 & 0.100 &	0.729 &	4.431 &	0.175 &	0.900 & 0.967 & 0.984 
		\\
		5 & \checkmark & \checkmark &\checkmark & $\circ$ & 24.6 & 32.3 & \textbf{0.093} & 0.629 & 4.195 & 0.170 & 0.909 & 0.968 & \textbf{0.985}
		\\
		6 & \checkmark & \checkmark & \checkmark & \checkmark & 24.6 & 32.3 & \textbf{0.093} & \textbf{0.605} & \textbf{4.114} & \textbf{0.167} & \textbf{0.910} & \textbf{0.969} & \textbf{0.985}
		\\
		
		\bottomrule
        
	\end{tabular}
    }

\end{table*}

\subsection{Ablation Experiment}
\label{subsec:ablation_studies}

The ablation results for all components are summarized in Table~\ref{tab:ablation_decoder}. For consistency, all experiments are conducted using the Flex-X-Large variant under identical training protocols.

\textbf{Impact of Post-Convolution Modules} (ID 1 vs. 6): As observed, employing the HEB-based post-convolution module slightly reduces computational overhead but results in a marginal performance degradation. This suggests that larger models are better suited for the HPB structure.

\textbf{Dynamic Upsampling} (ID 2 vs. 6): Traditional approaches relying on bilinear upsampling result in a moderate performance degradation, despite a marginal reduction in computational cost. This highlights the advantages of our proposed dynamic upsampling method: by adaptively determining optimal sampling locations from the original feature maps based on semantic content, our approach achieves more precise depth predictions.

\textbf{Inverted Prediction Head Order} (ID 3 vs. 6): Reversing the sequence of the prediction heads proves more beneficial for large-scale models. While this adjustment marginally increases computational cost by 0.2 GFLOPs, it yields a significant performance gain. This reinforces our rationale that high-channel features maintain superior structural and semantic consistency after upsampling, thereby enabling subsequent prediction convolutions to generate high-quality outputs with greater accuracy.

\textbf{Two-Stage Training Efficiency} (ID 4, 5, vs. 6): Excluding the second-stage processing for dynamic environments leads to a severe performance drop. This underscores the superiority of our Two-Stage Static-Dynamic Decoupled Training strategy. Compared to a fixed-threshold approach, our proposed static-dynamic decoupled mask $M$ provides higher precision and demonstrates better adaptability across a broader range of environments.








\begin{table}[t]
    \centering
    \begin{minipage}[t]{0.48\linewidth}
        \centering
        \caption{Quantitative Results on the Make3D dataset~\cite{saxena2008make3d}.}
        \label{tab:make3d}
         \resizebox{\linewidth}{!}{
        \setlength{\tabcolsep}{1pt}
        \scriptsize
        \begin{tabular}{ccccc}
            \toprule
            Method & Abs Rel & Sq Rel & RMSE & RMSE log \\
            \midrule
            Monodepth2~\cite{godard2019digging} & 0.322 & 3.589 & 7.417 & 0.163 \\
            Lite-Mono-Tiny~\cite{zhang2023lite} & 0.318 & 3.217 & 7.234 & 0.164 \\
            Lite-Mono-Small~\cite{zhang2023lite} & 0.310 & 2.878 & 6.842 & 0.160 \\
            Lite-Mono~\cite{zhang2023lite} & 0.305 & 3.060 & 6.981 & 0.158 \\
            Lite-Mono-8M~\cite{zhang2023lite} & 0.309 & 3.145 & 7.016 & 0.158 \\
            GeoDepth~\cite{wu2025geodepth} & 0.296 & \textbf{2.750} & 6.735 & 0.153 \\
            \midrule
            \textbf{Flex-Small (Ours)} & 0.315 & 3.127 & 6.865 & 0.159 \\
            \textbf{Flex-Medium (Ours)} & 0.292 & 2.864 & 6.686 & 0.152 \\
            \textbf{Flex-X-Large(Ours)} & \textbf{0.279} & 2.799 & \textbf{6.599} & \textbf{0.144} \\
            \bottomrule
        \end{tabular}
        }
    \end{minipage}
    \hfill
    \begin{minipage}[t]{0.48\linewidth}
        \centering
        \caption{Speed  at 640$\times$192 resolution (SD 8Elite: Snapdragon 8 Elite ).}
        \label{tab:speed_tab_simple}
        \resizebox{\linewidth}{!}{
        \setlength{\tabcolsep}{1pt}
        \scriptsize
        \begin{tabular}{cccccc}
			\toprule
			\multicolumn{1}{c}{\multirow{2}{*}{Method}} &
			\multicolumn{2}{c}{Full Model $\downarrow$} &
			\multicolumn{3}{c}{Speed (FPS) $\uparrow$} \\
			\multicolumn{1}{c}{} &
			Params(M) &
			FLOPs(G) &
			4090 &
			V100 &
			
			SD 8Elite \\ \midrule
				
			Lite-Mono-Tiny~\cite{zhang2023lite}   & 2.143 & 2.842  & 1048 &  312  & 15.9 \\
			Lite-Mono-Small~\cite{zhang2023lite}   & 2.472 & 4.746 & 816 &  298  & 10.7 \\
			Lite-Mono~\cite{zhang2023lite}  & 3.069 & 5.032 & 781 & 295 &  10.3 \\
			Lite-Mono-8M~\cite{zhang2023lite}   & 8.766 & 11.212  & 592 &  223  & 5.6 \\
			MonoVit~\cite{zhao2022monovit}  & 33.631 & 59.679  & 265 & 126  & 1.9 \\
            \midrule
			\textbf{Flex-Nano(Ours)} & \textbf{1.522} & \textbf{0.718}  & \textbf{1583} &  \textbf{774}  & \textbf{37.6} \\
			\textbf{Flex-Small(Ours)} & 6.059 & 2.776  & 1215   & 606  & 18.6 \\
			\textbf{Flex-Medium(Ours)} & 12.722 & 9.994 & 510 & 231  & 5.8 \\
			\textbf{Flex-Large(Ours)} & 15.203 & 11.519  & 467 &  216  & 5.2 \\
			\textbf{Flex-X-Large(Ours)} & 32.269 & 24.563  & 321 & 130  & 3.0 \\ \bottomrule
		\end{tabular}
        }
    \end{minipage}
\end{table}

\subsection{Complexity and Speed Evaluation}
\label{subsec:speed_comparison}
Table \ref{tab:speed_tab_simple} presents computational cost and inference speed, where GPU benchmarks (RTX 4090, V100) use batch size 16 and mobile platforms (Snapdragon 8 Elite) use batch size 1 in performance mode. Our small model achieves improvements while maintaining strong efficiency. Medium and large models involve slightly higher costs but significantly enhance performance. We provide a lightweight model for resource-constrained scenarios and a more accurate alternative for abundant computational resources, maintaining real-time inference without substantially increasing demands.


	\begin{table}[t]
	\centering
	\caption{Dynamic-scene depth estimation comparison on Cityscapes~\cite{cordts2016cityscapes}. Our method uses no auxiliary cues (pixel motion, pseudo depth) and is most efficient.}
	\label{tab_dyn}
	\resizebox{\columnwidth}{!}{
		\begin{tabular}{c ccc cccc c}
			\toprule
			\multirow{2}{*}{Method} & \multicolumn{3}{c}{Extra Info} & \multicolumn{2}{c}{Dynamic region} & \multicolumn{2}{c}{Static region} & \multirow{2}{*}{FLOPs(G)$\downarrow$} \\
			\cmidrule(lr){2-4} \cmidrule(lr){5-6} \cmidrule(lr){7-8}
			& \scriptsize{Pix.Mot.} & \scriptsize{Pseu.Dep.} & \scriptsize{Frames} & Abs Rel$\downarrow$ & $\delta < 1.25 \uparrow$ & Abs Rel$\downarrow$ & $\delta < 1.25 \uparrow$ & \\
			\midrule
			ProDepth~\cite{woo2024prodepth} & - & - & 2 & 0.115 & 0.884 & 0.092 & 0.911 & 35.5 \\
			Mining-BrNet~\cite{nguyen2024mining} & \checkmark & \checkmark & 1 & 0.119 & 0.872 & \textbf{0.080} & 0.922 & 27.0 \\
			DSI-MonoVit~\cite{zhang2025depth} & - & - & 1 & 0.097 & 0.918 & 0.086 & 0.919 & 47.7 \\
			\midrule
			\textbf{Flex-X-Large (Ours)} & - & - & 1 & \textbf{0.091} & \textbf{0.935} & 0.085 & \textbf{0.925} & \textbf{19.6} \\
			\bottomrule
		\end{tabular}
	}
\end{table}
	\subsection{Comparison with Foundation Depth Models}
\label{sec:foundation_compare}
Recent monocular depth foundation models~\cite{yang2024depth,yang2024depth_2} have demonstrated impressive zero-shot generalization, raising the
question of whether self-supervised methods trained on in-domain data
still provide practical advantages. 
We address this by comparing FlexDepth with relative depth model Depth Anything V2
(DA2)~\cite{yang2024depth_2} under a unified evaluation protocol.

\subsubsection{Fair Comparison Protocol.}
\label{sec:fair_protocol}
To ensure a fair comparison, we unify the evaluation protocols of foundation models and self-supervised methods on KITTI~\cite{6248074}. 

\textbf{Depth Alignment.} Self-supervised methods like~\cite{godard2019digging,zhang2023lite} typically use per-image median scaling to align the predicted disparity to metric depth:
$
	\hat{D}_i = D_i \cdot \frac{\text{median}(D^{\text{gt}})}{\text{median}(D)},
$
where $D_i = 1/\text{disp}_i$, $D^\text{gt}$ is the ground-truth depth, and $\hat{D}_i$ is the aligned depth.
Foundation models like ~\cite{yang2024depth,yang2024depth_2}  use least-squares.  Specifically, for each image  
$
	\frac{1}{\hat{D}_i} = s \cdot \text{disp}_i + t, \;
	[s, t] = \arg\min_{s,t} \sum_j \big\| s \cdot \text{disp}_j + t - \tfrac{1}{D^{\text{gt}}_j} \big\|^2.
$ Here $s,t$ are the scale and shift over valid pixels $j$.

\textbf{Ground Truth.} Foundation models are typically evaluated on the improved KITTI~\cite{6248074}
ground truth~\cite{uhrig2017sparsity} with dense multi-frame stereo
completion, whereas self-supervised methods usually use the sparse LiDAR-based.
We therefore additionally evaluate FlexDepth under the
foundation-model protocol (LS alignment + improved GT).
Full details are provided in the Appendix.
\begin{table}[t]
	\centering
	\caption{\textbf{Comparison with DA2~\cite{yang2024depth_2} on KITTI~\cite{6248074} Eigen~\cite{eigen2015predicting} split
			(LS alignment + improved GT).} DA2 results are reproduced using official weights. $^\dagger$: Re-evaluation of our model under this protocol for fair comparison, without introducing any new training or experiments.
		Full results in Appendix.}
	\label{tab:foundation_main}
	\small
	
	\begin{tabular}{ccccccc}
		\toprule
		Method & Type & Params & GFLOPs & Resolution & Abs Rel$\downarrow$ & $\delta<1.25$$\uparrow$ \\
		\midrule
		DA2 (ViT-L) &Zero-Shot& 335M & 1947 & $1722{\times}518$ & 0.070 & \textbf{0.956} \\
		DA2 (ViT-S) &Zero-Shot& 25M  & 137  & $1722{\times}518$ & 0.077 & 0.944 \\
		DA2 (ViT-L) &Zero-Shot& 335M & 276 & 644$\times$196 & 0.092 & 0.915 \\
		DA2 (ViT-S) &Zero-Shot& 25M  & 19   & $644{\times}196$  & 0.110 & 0.881 \\
		\midrule
		Flex-X-Large$^\dagger$ &Self-supervised        & 32M  & 25   & $640{\times}192$  & \textbf{0.063} & 0.952 \\
		\bottomrule
	\end{tabular}
\end{table}

\subsubsection{Discussion.}
Table~\ref{tab:foundation_main} reports results under the matched protocol (LS alignment + improved GT). DA2~\cite{yang2024depth_2} demonstrates strong zero-shot generalization and excels at fine details ( $\delta<1.25$ 0.956 vs. 0.952), but is sensitive to distant objects (see Appendix). In contrast, FlexDepth, self-supervised trained on KITTI~\cite{6248074}, achieves superior accuracy (Abs Rel 0.063 vs.\ 0.070) with far fewer parameters and FLOPs. This suggests that in-domain self-supervision still offers practical value particularly for resource-constrained deployment complementing the strengths of large foundation models.

	\section{Conclusion and Discussion}
\label{sec:.conclusion}
This paper presents FlexDepth, a scale-driven family of flexible self-supervised monocular depth estimation models specifically engineered for robust driving perception.  FlexDepth introduces a Scale-Driven Decoder (SDD), which adaptively selects its constitutive components to enhance feature fusion and generate high-precision depth maps. Furthermore, this paper proposes a two-stage static-dynamic decoupling training strategy designed to isolate dynamic driving entities from the static background. By performing independent confidence assessments, our approach achieves superior robustness in highly dynamic environments. Extensive evaluations demonstrate that FlexDepth achieves state-of-the-art performance across multiple autonomous driving benchmarks. Its minimal computational footprint allows for seamless deployment on edge devices, while maintaining excellent zero-shot generalization.

\noindent \textbf{Limitations.} \ Since FlexDepth is designed for robust driving perception, its applicability to man-made structures indoors with irregular textures and close-range occlusions is limited.

	\par\vfill\par
	\section*{Acknowledgements}
This work is supported by the National Natural Science Foundation of China under Grant 62332016.
	\bibliographystyle{splncs04}
	\bibliography{main}
	\par\vfill\par

	\section{Supplementary Material}
	\label{sec:x_intro}
	This document provides additional implementation details, ablation studies, and extended qualitative results for the proposed FlexDepth framework for self-supervised monocular depth estimation.
	
	FlexDepth is designed to address two key challenges in monocular depth estimation for driving scenarios:
	
	\begin{itemize}
		\item \textbf{Flexible model scaling}: enabling a unified decoder design that supports multiple efficiency–accuracy trade-offs.
		\item \textbf{Robust dynamic scene learning}: improving depth estimation in dynamic environments through a static-dynamic decoupled training strategy.
	\end{itemize}
	The content is organized as follows:
	\begin{itemize}
		
		\item \textbf{Sec.~\ref{sec:x_decoder_architecture}:} Decoder feature visualizations contrasting our method with conventional baselines.
		\item \textbf{Sec.~\ref{sec:x_dynamic_masking}:} Details of the Static-Dynamic Decoupled network, including the training-time $\mu$ prediction mechanism and visualization of generated masks.
		\item \textbf{Sec.~\ref{sec:x_decoder_effict}:} Comprehensive ablation study validating the effectiveness of the scalable decoder architecture across different backbone networks and model scales, along with detailed analysis of the static-dynamic decoupled training strategy.
		\item \textbf{Sec.~\ref{sec:x_model_scaling}:} Extended ablation studies covering encoder selection strategies and component validation across four model scales.
		\item \textbf{Sec.~\ref{sec:x_model_speeds}:} Comprehensive inference speed analysis across diverse hardware platforms (GPU and mobile).
		\item \textbf{Sec.~\ref{sec:x_comparison_with_foundation}:} Comparison with foundation models' zero-shot results.
		\item \textbf{Sec.~\ref{subsec:x_additional_generalization}:} Additional cross-dataset generalization results.
		
		\item \textbf{Sec.~\ref{sec:x_qualitative_results}:} Additional qualitative comparisons demonstrating performance in complex scenarios.
	\end{itemize}

	\subsection{Decoder Architecture Analyse}
	\label{sec:x_decoder_architecture}

	\begin{figure}[htbp]
	\centering
	\includegraphics[width=0.45\textwidth]{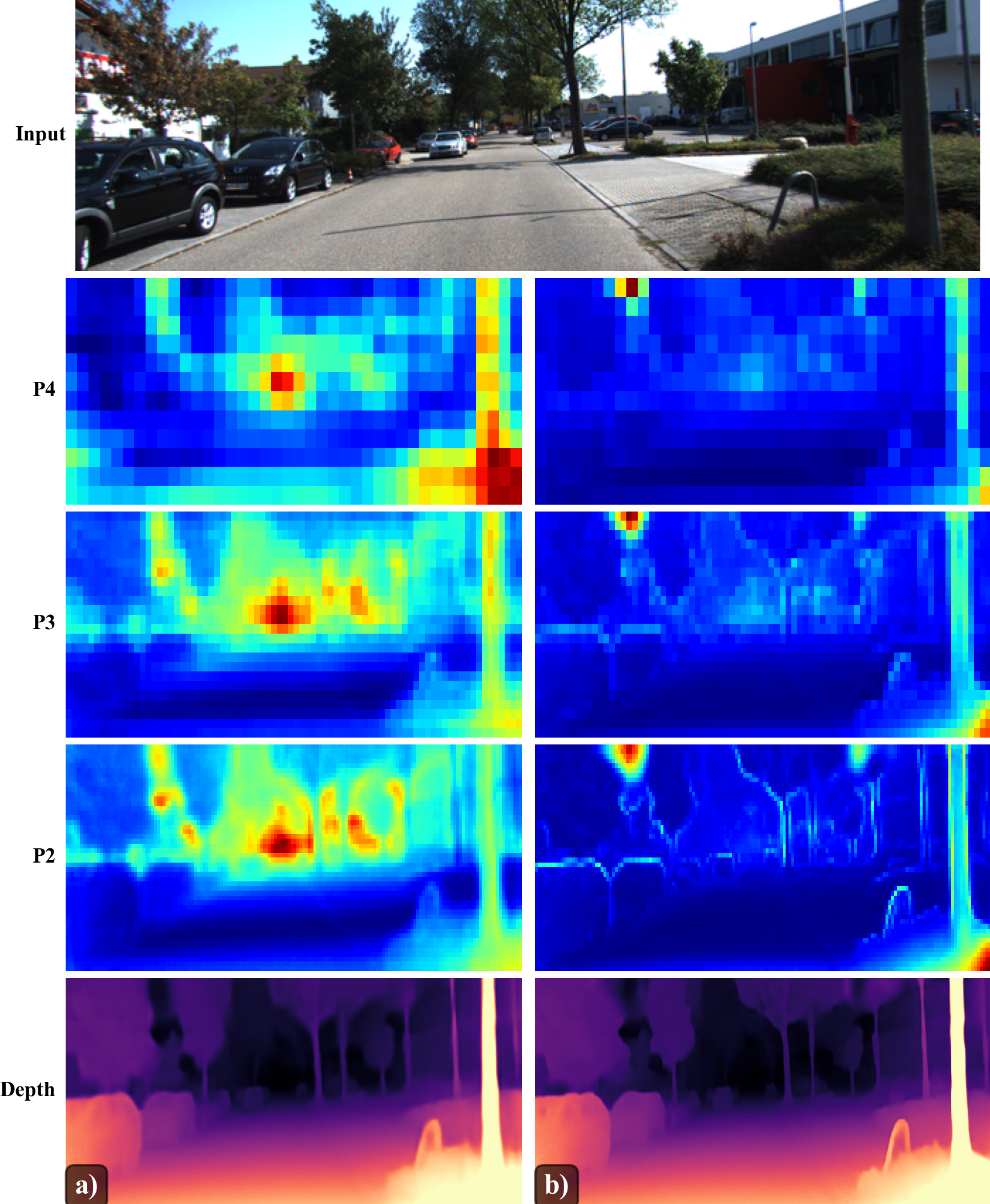}

\caption{
	\textbf{Decoder Feature Fusion  Comparison.} baseline (left) vs. our method (right) using same backbone. Our approach yields sharper feature boundaries during fusion and cleaner edges in final depth estimation.
}
	\label{fig:decoder_feature_map}
\end{figure}

	\textbf{Scale-Driven Decoder(SDD) vs. Conventional Approaches.}  contrasts our SDD with standard fusion paradigms~\cite{godard2019digging,zhang2023lite,woo2024prodepth}.  This design facilitates earlier feature interaction and minimizes information loss. As visualized in Figure~\ref{fig:decoder_feature_map}, our approach effectively preserves structural details—particularly at object boundaries and fine-grained textures—that are frequently smoothed out by traditional architectures.

\begin{figure}[htbp]
	\centering
	\includegraphics[width=0.35\textwidth]{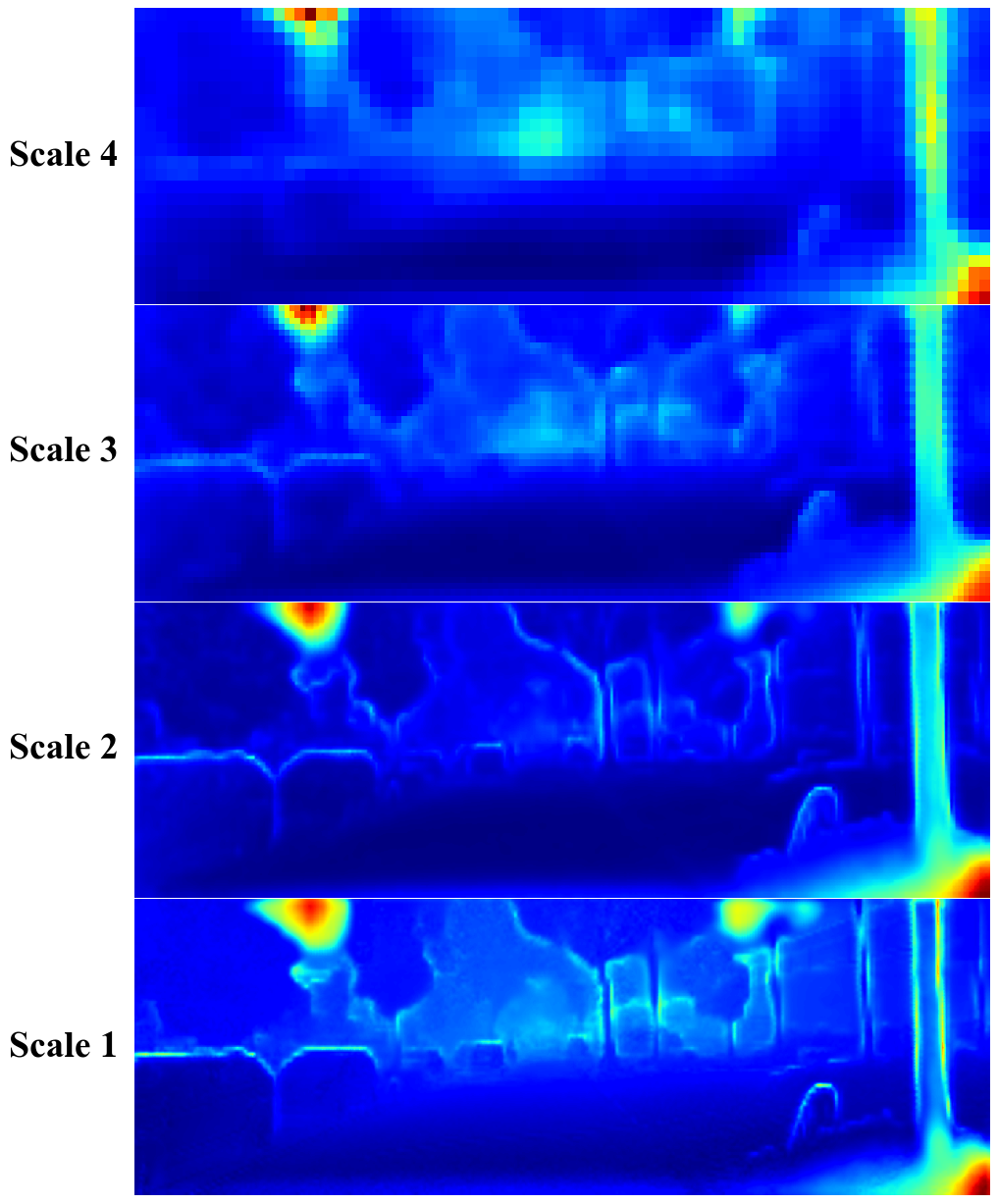}
\caption{\textbf{Feature Maps from the $\mathcal{P}$ across Multiple Scales.} This visualization demonstrates how our $\mathcal{P}$ Head maintains rich, multi-scale feature representations, showcasing its ability to produce consistent structural details before final depth prediction.}
\label{fig:decoder_dup_map}

\end{figure}

	\textbf{Inverted Upsampling Prediction Head.} Figure~\ref{fig:decoder_dup_map} demonstrates that this inversion yields sharper depth discontinuities and more coherent surfaces, a significant advantage for medium and large-scale models where precision is paramount.

	\subsection{Static-Dynamic Decoupled Mask:  Network and Predictions}
	\label{sec:x_dynamic_masking}
	\begin{figure*}[t]
    \centering
    \includegraphics[width=\textwidth]{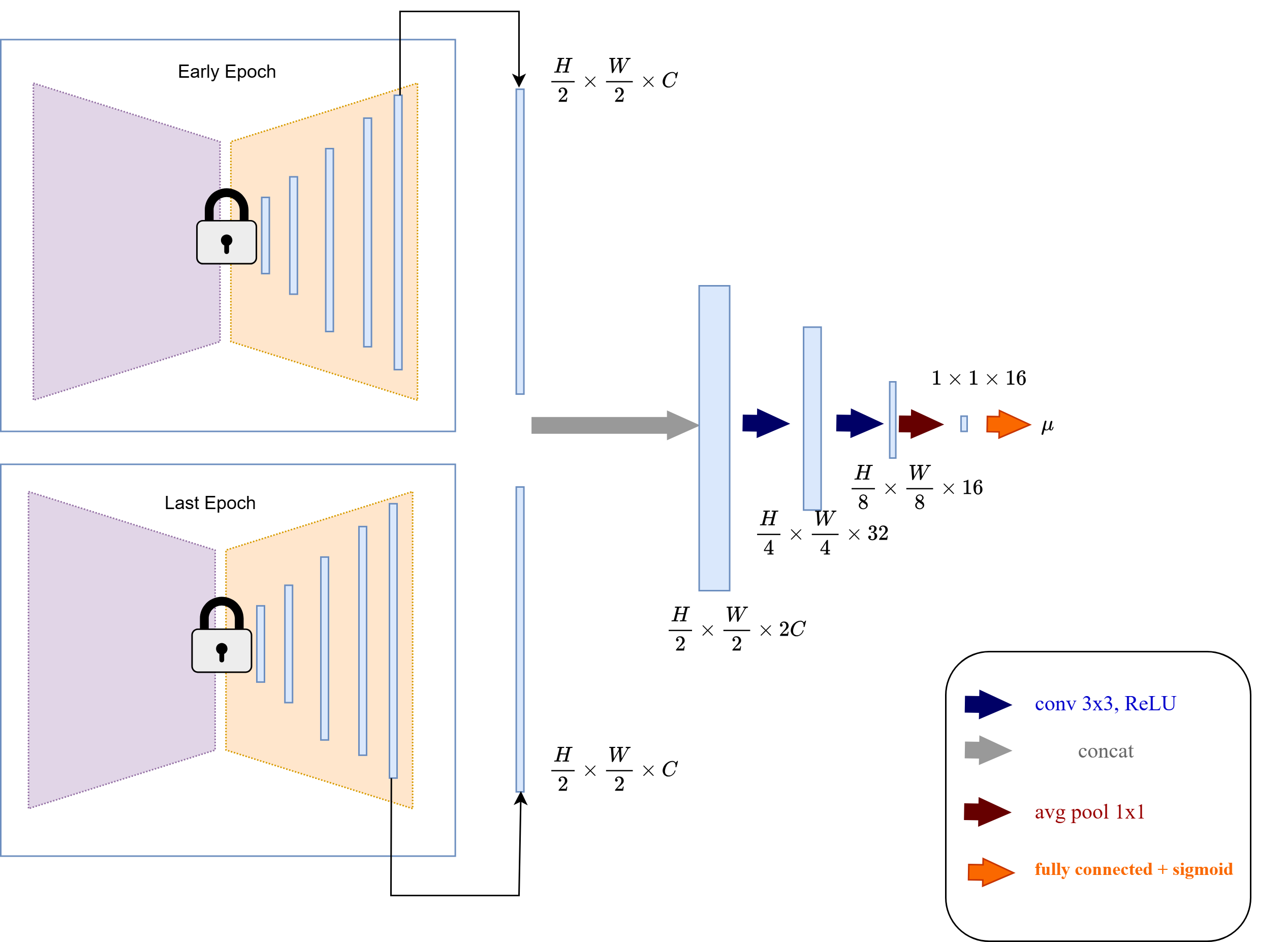}
    
    \caption{\textbf{Architecture for Adaptive Dynamic Masking.} This network is designed to generate a frame-specific $\mu$ parameter, which dynamically adjusts the masking threshold based on the scene content. This adaptive mechanism enhances robustness, particularly in diverse and highly dynamic environments.}
    \label{fig:x_u_predictor}
\end{figure*}
	
	Building upon the fixed-threshold masking approach of \cite{zhang2025depth}, our Static-Dynamic Decoupled Mask architecture (Figure~\ref{fig:x_u_predictor}) introduces an adaptive threshold generation mechanism that dynamically produces an image-specific threshold, $\mu_{dynamic}$, based on the model's training progress and input characteristics. This threshold adapts to both image content and the evolving model capacity, enabling more precise decoupling of static and dynamic regions. The masks are derived from the discrepancy between early-stage and late-stage training predictions, leveraging the inherent predictive instability observed in dynamic regions during training. Importantly, this adaptive masking is applied only during training, incurring minimal additional computation and having no impact on inference efficiency.
	
\begin{figure*}[t]
	\centering
	
	\begin{subfigure}[b]{\textwidth}
		\centering
		\includegraphics[width=0.22\textwidth]{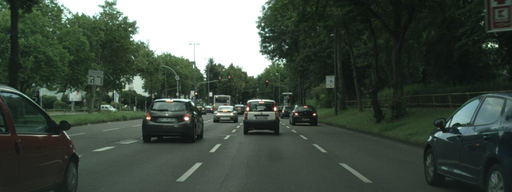}
		\includegraphics[width=0.22\textwidth]{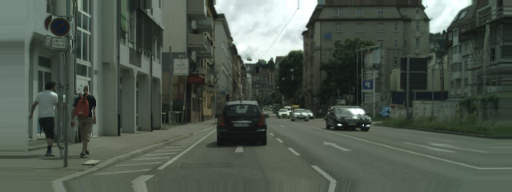}
		\includegraphics[width=0.22\textwidth]{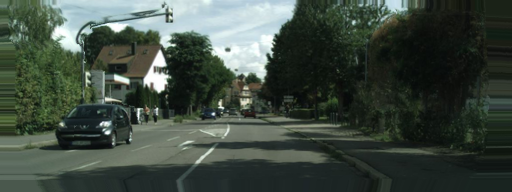}
		\includegraphics[width=0.22\textwidth]{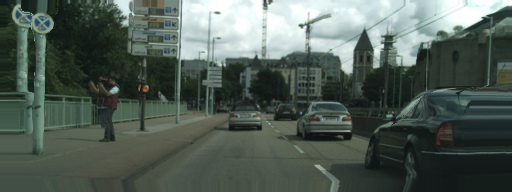}

	\end{subfigure}

	\begin{subfigure}[b]{\textwidth}
		\centering
		\includegraphics[width=0.22\textwidth]{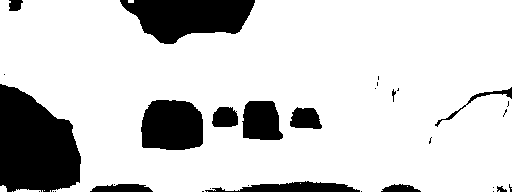}
		\includegraphics[width=0.22\textwidth]{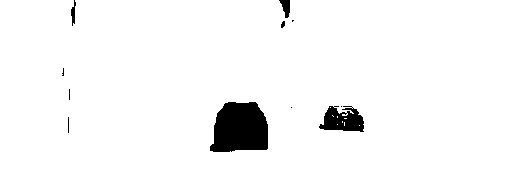}
		\includegraphics[width=0.22\textwidth]{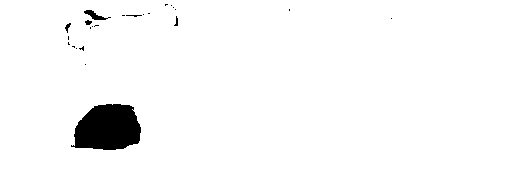}
		\includegraphics[width=0.22\textwidth]{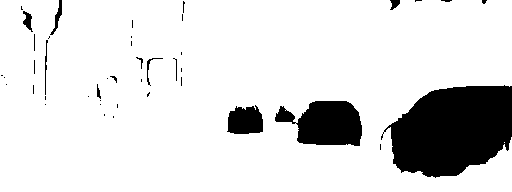}
		
	\end{subfigure}
	
	\caption{\textbf{static-dynamic decoupled mask  examples.} The top row shows input scenes. The bottom row presents the corresponding dynamically predicted masks, each with a scene-specific $\mu$ threshold. This illustrates how our method automatically generates a customized $\mu$ for precise dynamic region detection in varying environments.}
	\label{fig:dynamic_masking}
\end{figure*}
	
	Figure~\ref{fig:dynamic_masking} validates this adaptive approach. As illustrated, each input image is assigned a dynamically generated threshold $\mu_{\text{dynamic}}$ to binarize the exponential-difference mask :
	\[
	M = \big[\,|\text{disp}_i^{\beta} - \text{disp}_l^{\beta}| < \mu_{\text{dynamic}}\,\big],
	\]
	where $\text{disp}_i$ and $\text{disp}_l$ denote predictions from early and late training stages, respectively. By leveraging the higher predictive variance inherent to dynamic regions, this formulation effectively separates moving objects from static backgrounds without requiring manual threshold tuning. Following DSI~\cite{zhang2025depth}, we employ $\beta=0.7$ when constructing the discrepancy representation. Empirically, values around $0.7$ provide a good balance between mask sensitivity and stability. Smaller values tend to over-suppress valid pixels, whereas larger values may fail to sufficiently highlight dynamic regions.

	\paragraph{Adaptive Threshold Prediction.}
	
	The discrepancy function $\phi(\cdot)$ measures the feature inconsistency between
	the early-stage predictor $f_{s_i}$ and the late-stage predictor $f_{s_l}$.
	Since dynamic objects violate the static-world assumption, their predictions
	exhibit larger temporal instability than static regions.
	
	A lightweight predictor $g(\cdot)$  then estimates an image-specific threshold
	$\mu_{dynamic}$ from the discrepancy representation:
	
	\[
	\mu_{dynamic}=g(\phi(f_{s_i},f_{s_l}))
	\]
	
	which enables adaptive mask generation for different scenes and details the structure of $g(\phi(f_{s_i},f_{s_l}))$, as shown in Figure~\ref{fig:x_u_predictor}.

	\begin{table*}[htbp]
	
	\caption{Ablation experiments on our decoder and static-dynamic decoupled architectures. Legend: Dynamic: \checkmark=static-dynamic decoupled mask $M$, $\times$=w/o stage-2. F.=FLOPs, P.=Params.}
	\label{tab:x_decoder_effict}
	\centering
	\resizebox{\textwidth}{!}{
		
		\setlength{\tabcolsep}{2pt}
		\begin{tabular}{cccccccccccccccccc}
			\toprule
			\multirow{2}{*}{ID} &
			\multirow{2}{*}{Method} &
			
			\multirow{2}{*}{Dynamic} &
			\multirow{2}{*}{F.(G)} &
			\multirow{2}{*}{P.(M)} &
			\multicolumn{4}{c}{Depth Error $\downarrow$} &
			\multicolumn{3}{l}{Depth Accuracy $\uparrow$} \\ 
			&  &
		
			&
			&
			&
			\multicolumn{1}{c}{Abs Rel} &
			\multicolumn{1}{c}{Sq Rel} &
			\multicolumn{1}{c}{RMSE} &
			\multicolumn{1}{c}{RMSE log} &
			$\delta < 1.25 $ &
			$\delta < 1.25^2$ &
			$\delta < 1.25^3$
			\\ \midrule
			1 & MonoVit~\cite{zhao2022monovit}  & $\times$ & 59.7 & 33.6 & 	0.099 &
			0.708 &
			4.372 &
			0.175 &
			0.900 &
			0.967 &
			0.984 \\
			2 & MonoVit~\cite{zhao2022monovit} + Ours Nano Decoder & $\times$ & \textbf{26.7} & \textbf{23.4} & 	0.104 &	0.731 &	4.565&	0.179&	0.885&	0.963&	0.984 \\
			
			3 & MonoVit~\cite{zhao2022monovit} + Ours Small Decoder & $\times$ & 27.5 & 23.8 & 0.102&	0.708&	4.439&	0.176&	0.890&	0.965&	0.984 \\
			
			4 & MonoVit~\cite{zhao2022monovit} + Ours Medium/Large Decoder & $\times$ & 30.1 & 25.2 &0.101  & 0.757 &  4.433 &  0.177 &  0.899  & 0.966 &  0.984  \\
			5 & MonoVit~\cite{zhao2022monovit} + Ours X-Large Decoder & $\times$ & 33.2 & 25.8 &0.099  & 0.703 &  4.392 &  0.174 &  0.900  & 0.967 &  0.984  \\
			
			6 & MonoVit~\cite{zhao2022monovit} + Ours Nano Decoder & \checkmark & \textbf{26.7} & \textbf{23.4} & 	0.099 &	0.744&	4.437&	0.174&	0.898&	0.966&	0.984 \\
			
			7 & MonoVit~\cite{zhao2022monovit} + Ours Small Decoder & \checkmark & 27.5 & 23.8 & 0.098& 	0.703& 	4.359& 	0.173& 	0.900& 	0.967& 	0.984 \\
			
			8 & MonoVit~\cite{zhao2022monovit} + Ours Medium/Large Decoder & \checkmark & 30.1 & 25.2 &\textbf{0.095}&	0.690&	\textbf{4.290}&	\textbf{0.171}&	\textbf{0.907}&	\textbf{0.968}&	\textbf{0.984}  \\
			9 & MonoVit~\cite{zhao2022monovit} + Ours X-Large Decoder & \checkmark & 33.2 & 25.8 &\textbf{0.095}  & \textbf{0.685} &  4.309 &  0.172 &  \textbf{0.907}  & \textbf{0.968} &  \textbf{0.984}  \\
			\midrule
			
			10 & Monodepth2~\cite{godard2019digging}  & $\times$ & 8.0 & 14.3 & 	0.115 &
			0.903 &
			4.863 &
			0.193 &
			0.877 &
			0.959 &
			0.981 & \\
			11 & Monodepth2~\cite{godard2019digging} + Ours Nano Decoder & $\times$ & \textbf{4.8} & \textbf{11.4} & 	0.113& 	0.829& 	4.790& 	0.190& 	0.875& 	0.959& 	0.982 \\
			
			12 & Monodepth2~\cite{godard2019digging} + Ours Small Decoder & $\times$ & 5.4 & 11.7 &0.113&	0.860&	4.782&	0.189&	0.877&	0.960&	0.982 \\
			
			13 & Monodepth2~\cite{godard2019digging} + Ours Medium/Large Decoder & $\times$ & 8.1 & 13.3 &0.112  & 0.866 &  4.788 &  0.190 &  0.881  & 0.960 &  0.982  \\
			14 & Monodepth2~\cite{godard2019digging} + Ours X-Large Decoder & $\times$ & 11.1 & 13.9 &0.110  & 0.866 &  4.752 &  0.189 &  0.883  & 0.961 &  0.982  \\
			
			15 & Monodepth2~\cite{godard2019digging} + Ours Nano Decoder & \checkmark & \textbf{4.8} & \textbf{11.4} & 	0.113& 	0.827& 	4.781& 	0.190& 	0.875& 	0.959& 	0.982 \\
			
			16 & Monodepth2~\cite{godard2019digging} + Ours Small Decoder & \checkmark & 5.4 & 11.7 &0.113&	0.850&	4.762&	\textbf{0.183}&	0.877&	0.960&	0.982 \\
			
			17 & Monodepth2~\cite{godard2019digging} + Ours Medium/Large Decoder & \checkmark & 8.1 & 13.3 &\textbf{0.108}&	0.798&	4.661&	0.186&	\textbf{0.885}&	\textbf{0.962}&	0.982  \\
			18 & Monodepth2~\cite{godard2019digging} + Ours X-Large Decoder & \checkmark & 11.1 & 13.9 &0.109&	\textbf{0.795}&	\textbf{4.650}&	0.186&	0.884&	\textbf{0.962}&	\textbf{0.983}  \\
			\bottomrule
			
		\end{tabular}
	}

\end{table*}
	
	\subsection{Ablation Study on Scalable Decoder and Static-Dynamic Decoupled Training}
	\label{sec:x_decoder_effict}

	Table~\ref{tab:x_decoder_effict} presents ablation experiments validating our two key contributions: the scalable decoder architecture and the static-dynamic decoupled training strategy. Experiments are conducted on MonoViT~\cite{zhao2022monovit} and Monodepth2~\cite{godard2019digging} to verify generality across different backbones.

	\paragraph{Contribution I: Scalable Decoder Architecture.}
	We design five decoder variants: Nano, Small, Medium, Large, and X-Large.
	Their width multipliers are 0.25, 0.5, 1.0, 1.0, and 1.5, respectively.
	This design enables flexible trade-offs between accuracy and computational efficiency across diverse deployment scenarios.
	
	\textbf{Superior Efficiency-Performance Trade-off.}
	Comparing ID1 with ID5 reveals that our X-Large decoder achieves comparable performance to the original MonoViT decoder while being significantly more efficient: FLOPs reduce from 59.7G to 33.2G (44\% reduction) and parameters from 33.6M to 25.8M (23\% reduction), with identical Abs Rel of 0.099. This demonstrates that under the same encoder, our decoder design outperforms MonoViT's original decoder by achieving equivalent accuracy with substantially lower computational cost.
	
	On Monodepth2, comparing ID10 with ID12 shows even stronger advantages: our Small decoder outperforms the original decoder across all metrics despite using fewer resources—FLOPs reduce from 8.0G to 5.4G (33\% reduction), parameters from 14.3M to 11.7M (18\% reduction), while Abs Rel improves from 0.115 to 0.113 and $\delta<1.25^2$ increases from 0.959 to 0.960. This comprehensive improvement validates our decoder's superior design over Monodepth2's original architecture.
	
	\textbf{Scalability Analysis.}
	Larger decoder variants consistently improve performance. On MonoViT (ID2--ID5), the X-Large decoder (ID5) matches the baseline with only 55\% FLOPs. Similar trends on Monodepth2 (ID11--ID14) confirm that our scaling strategy generalizes across architectures.
	
	\paragraph{Contribution II: Static-Dynamic Decoupled Training.}
	Our training strategy separates static regions from dynamic objects through adaptive mask generation, addressing the photometric loss failure in dynamic scenes.
	
	\textbf{Performance Gains on MonoViT.}
	The two-stage training provides consistent improvements across decoder scales. For the X-Large decoder, ID5 vs.\ ID9 shows Abs Rel improving from 0.099 to 0.095 (4.0\% gain) and RMSE from 4.392 to 4.309. Notably, the progression ID1 $\rightarrow$ ID5 $\rightarrow$ ID9 demonstrates that our decoder first achieves efficiency parity with the baseline, and the two-stage training further elevates performance beyond the original MonoViT's capability, achieving the best Abs Rel of 0.095.
	
	\textbf{Performance Gains on Monodepth2.}
	Dynamic decoupling yields substantial improvements. For the Medium decoder (ID13 vs.\ ID17), Abs Rel drops from 0.112 to 0.108 and Sq Rel from 0.866 to 0.795. The X-Large decoder with dynamic decoupling (ID18) achieves the lowest Sq Rel of 0.795, demonstrating effective handling of dynamic regions.
	
	\paragraph{Design Rationale.}
	The Medium and Large decoders share identical channel configurations $[32, 64, 128, 256, 512]$ since both use width multiplier $w=1.0$ in our scaling rule:
	\[
	C_i = \min(\lfloor w \cdot B_i \rfloor, C^{\max}),
	\]
	where $B_i$ denotes base channels. This design balances capacity and efficiency while maintaining architectural consistency.
	
	In summary, our ablation demonstrates: (1) the scalable decoder achieves superior efficiency-performance trade-offs compared to original decoder designs, and (2) the static-dynamic decoupled training consistently improves depth accuracy across all configurations.
	
	Besides the experiments on dynamic regions in Cityscapes~\cite{cordts2016cityscapes} reported in the manuscript, to further verify the effectiveness of the proposed static-dynamic decoupling strategy, we separately evaluate depth estimation performance on static and dynamic regions in KITTI~\cite{6248074}. Dynamic regions are obtained from the semantic annotations ~\cite{jiang2021unsupervised} following the evaluation protocol used in DSI-MonoVit ~\cite{zhang2025depth}.
	\begin{table}[t]
		\centering
		\caption{\textbf{Dynamic-region analysis on KITTI~\cite{6248074} Eigen~\cite{eigen2015predicting} test split.}
			Static background and dynamic objects are segmented using semantic annotations from~\cite{jiang2021unsupervised}. FlexDepth consistently improves depth estimation accuracy on dynamic regions while also improving static-background performance.}
		\resizebox{\textwidth}{!}{
			\begin{tabular}{l c *{9}{c}}
				\toprule
				\multirow{2}{*}{Region} & \multirow{2}{*}{Pixels} & 
				\multicolumn{3}{c}{Abs Rel $\downarrow$} & 
				\multicolumn{3}{c}{Sq Rel $\downarrow$} & 
				\multicolumn{3}{c}{$\delta < 1.25$ $\uparrow$} \\
				\cmidrule(lr){3-5} \cmidrule(lr){6-8} \cmidrule(lr){9-11}
				& & DSI-MonoVit~\cite{zhang2025depth} & Ours & Gain (\%) & 
				DSI-MonoVit~\cite{zhang2025depth} & Ours & Gain (\%) & 
				DSI-MonoVit~\cite{zhang2025depth} & Ours & Gain (\%) \\
				\midrule
				Background & 11,379,145 & 0.092 & \textbf{0.089} & 3.3 & 
				0.656 & \textbf{0.565} & 13.9 & 
				0.910 & \textbf{0.914} & 0.4 \\
				Moving Objects & 438,130 & 0.177 & \textbf{0.159} & 10.2 & 
				1.918 & \textbf{1.484} & 22.6 & 
				0.794 & \textbf{0.819} & 3.2 \\
				Similar Moving Vehicles & 64,300 & 0.175 & \textbf{0.157} & 10.3 & 
				2.781 & \textbf{1.992} & 28.4 & 
				0.828 & \textbf{0.844} & 1.9 \\
				Dissimilar Moving Vehicles & 49,655 & 0.143 & \textbf{0.142} & 0.7 & 
				1.797 & \textbf{1.733} & 3.6 & 
				0.829 & \textbf{0.840} & 1.3 \\
				Cyclists/Bikers & 42,180 & 0.194 & \textbf{0.162} & 16.5 & 
				1.817 & \textbf{1.479} & 18.6 & 
				0.733 & \textbf{0.782} & 6.7 \\
				\bottomrule
			\end{tabular}
		}
		\label{tab:dynamic_analysis}
	\end{table}

	As shown in Table~\ref{tab:dynamic_analysis}, FlexDepth consistently improves depth estimation accuracy in both static and dynamic regions. More importantly, the largest gains are achieved on dynamic objects, demonstrating that the proposed decoupling strategy effectively mitigates violations of the static-scene assumption.
	
	\subsubsection{Stage-2 Control Experiment.}

	A potential concern is whether the improvements originate merely from additional training iterations. To isolate the contribution of the proposed dynamic masking strategy, we compare Stage-2 training against a standard fine-tuning baseline using an identical training budget.

	\begin{table}[t]
		\centering
		
		\begin{tabular}{l c c c c}
			\toprule
			\multirow{2}{*}{Method} & \multicolumn{2}{c}{KITTI~\cite{6248074}} & \multicolumn{2}{c}{Cityscapes~\cite{cordts2016cityscapes}} \\
			\cmidrule(lr){2-3} \cmidrule(lr){4-5}
			& AbsRel $\downarrow$ & $\delta_1$ $\uparrow$ & AbsRel $\downarrow$ & $\delta_1$ $\uparrow$ \\
			\midrule
			Stage 1 Only          & 0.100 & 0.900 & 0.116 & 0.882 \\
			Stage 1 + Fine-Tune   & 0.102 & 0.902 & 0.140 & 0.869 \\
			\textbf{Stage 1 + 2}  & \textbf{0.093} & \textbf{0.910} & \textbf{0.086} & \textbf{0.926} \\
			\bottomrule
		\end{tabular}%
		
		\caption{Effect of Stage 2 vs. fine-tuning: Comparison of depth estimation performance across training stages on KITTI~\cite{6248074} and Cityscapes~\cite{cordts2016cityscapes}. Best results are shown in bold}
		\label{tab:stage2gain}
	\end{table}
	Table~\ref{tab:stage2gain} demonstrates that simply extending the training schedule does not improve performance. In contrast, introducing the proposed dynamic masking strategy consistently improves results on both KITTI~\cite{6248074} and Cityscapes~\cite{cordts2016cityscapes}, confirming that the gains originate from the proposed static-dynamic decoupling mechanism rather than additional optimization.

	\subsection{Additional Ablation Studies}
	\label{sec:x_model_scaling}
	\begin{table*}
	\caption{\textbf{Ablation Study on Encoder Components}. Evaluation on the KITTI dataset~\cite{6248074} (Eigen split~\cite{eigen2015predicting}) at 640$\times$192 resolution. Decoder follows Monodepth2. ResNet-18 encoder results serve as baseline. \textbf{Bold} indicates best}
	\label{tab:ablation_encoder}
	\centering
	
	\setlength{\tabcolsep}{2pt}
	\resizebox{\textwidth}{!}{
	\begin{tabular}{@{}cccccccccccc@{}}
		\toprule
		\multirow{2}{*}{ID} &
		\multirow{2}{*}{Encoder} &
		\multirow{2}{*}{ Pretrained} &
		\multirow{2}{*}{FLOPs(G)} &
		\multirow{2}{*}{Params(M)} &
		\multicolumn{4}{c}{Depth Error↓} &
		\multicolumn{3}{c}{Depth Accuracy↑} \\ 
		&
		&
		&
		&
		&
		\multicolumn{1}{l}{Abs Rel} &
		\multicolumn{1}{l}{Sq Rel} &
		\multicolumn{1}{l}{RMSE} &
		\multicolumn{1}{l}{RMSE log} &
		\multicolumn{1}{l}{$\delta < 1.25$} &
		\multicolumn{1}{l}{$\delta < 1.25^2$} &
		\multicolumn{1}{l}{$\delta < 1.25^3$} \\ \midrule
		1 & ResNet-18 & ImageNet & 8.038 & 14.329 & 0.115    & 0.903 &  4.863     &  0.193   &  0.877         &  0.959 & \textbf{0.981} \\
		2 & 11n-cls & ImageNet & \textbf{3.383} & \textbf{3.455} &  0.113    & \textbf{0.863} &  4.770     &  0.191   & 0.876         & \textbf{0.96} & \textbf{0.981} \\
		3 & 11n-det & COCO     &  3.403    &  3.62     &  0.113    & 0.873          & 4.775          & 0.191   &  0.877   & 0.959   &  0.980     \\
		4 & 11n-seg & COCO-Seg &  3.403    &  3.62     & \textbf{0.111} &  0.864    & \textbf{4.748} & \textbf{0.19} & \textbf{0.880} & \textbf{0.960} & \textbf{0.981} \\ \bottomrule
	\end{tabular}
	}

\end{table*}
	\subsubsection{Encoder Selection.}
	\label{subsubsec:encoder_ablation}
	
	Table~\ref{tab:ablation_encoder} evaluates YOLO encoder backbones pre-trained under different regimes: ImageNet classification (11n-cls), COCO detection (11n-det), and COCO-Seg segmentation (11n-seg). The segmentation-pretrained variant (11n-seg) achieves the best performance (0.111 Abs Rel) while reducing FLOPs by 58\% and parameters by 75\% compared to the ResNet-18 baseline.
	
	We select YOLO11-Seg as the encoder because its feature representation aligns closely with the geometric requirements of depth estimation. Unlike classification (category-focused) or detection (box-focused), segmentation demands precise pixel-level boundary delineation. This sensitivity to fine edges and spatial structure directly translates to improved depth prediction quality.
	
	\subsubsection{Decoder Ablation Study.}
	\begin{table*}
\caption{\textbf{Ablation Study on Decoder Components across Model Scales.}  Evaluation on the KITTI dataset~\cite{6248074} (Eigen split~\cite{eigen2015predicting}) at 640$\times$192 resolution.  Legend: $\mathcal{C}$: \checkmark=HPB, $\circ$=HEB. $\mathcal{U}$: \checkmark=dynamic upsampling, $\circ$=bilinear. $\mathcal{P}$: \checkmark=inverted head, $\circ$=conventional head. Dynamic: \checkmark=static-dynamic decoupled mask $M$, $\circ$=fixed threshold, $\times$=w/o stage-2. \textbf{Bold} denotes the best result.}
\label{tab:ablation_decoder_efo}
\centering

\setlength{\tabcolsep}{2pt}
\resizebox{\textwidth}{!}{
\begin{tabular}{ccccccccccccc}
\toprule
\multirow{2}{*}{ID} &
\multirow{2}{*}{Encoder} &
  \multicolumn{3}{c}{Decoder} &
  \multirow{2}{*}{Dynamic} &
  \multirow{2}{*}{FLOPs(G)} &
  \multirow{2}{*}{Params(M)} &
  \multicolumn{4}{c}{Depth Error $\downarrow$} &
  \multicolumn{1}{l}{Depth Accuracy $\uparrow$} \\ 
 &  &
  $\mathcal{C}$ &
  $\mathcal{U}$ &
  $\mathcal{P}$ &
   &
   &
   &
   \multicolumn{1}{c}{Abs Rel} &
  \multicolumn{1}{c}{Sq Rel} &
  \multicolumn{1}{c}{RMSE} &
  \multicolumn{1}{c}{RMSE log} &
  $\delta \textless 1.25 $\\ \midrule

1 &    \multirow{3}{*}{11n-seg}  & $\circ$  & $\circ$ &$\circ$  & $\circ$ &\textbf{0.714} & \textbf{1.521} & \textbf{0.110} &	0.812 &	4.65 &	0.185 &	0.877
 \\

2&   & $\circ$ & \checkmark &$\circ$ &$\circ$ & 0.718 & 1.522 & \textbf{0.110} & 0.800 &	\textbf{4.678}&	0.186&	0.878
 
 \\
3 &   & $\circ$ & \checkmark &$\circ$& \checkmark &  0.718 &  1.522 & \textbf{0.110}	& \textbf{0.794}&	\textbf{4.678}&\textbf{0.184} &	\textbf{0.878} \\
 \midrule

 4 &  \multirow{3}{*}{11s-seg}     & $\circ$  & $\circ$ &$\circ$  & $\circ$ &\textbf{2.769} & \textbf{6.057} & \textbf{0.103} &	\textbf{0.697} &	4.507 &	0.179 &	0.886
 \\
 
 5&   & $\circ$ & \checkmark &$\circ$ &$\circ$ & 2.776 & 6.059 & 0.104 & 0.731 &	4.488&	\textbf{0.178}&	0.888
 
 \\
 6 &   & $\circ$ & \checkmark &$\circ$& \checkmark &   2.776 &  6.059 & 0.104	& 0.713&	\textbf{4.458}&	0.179 &	\textbf{0.890} \\
 \midrule

 7 &  \multirow{4}{*}{11m-seg}    & \checkmark  & $\circ$  &$\circ$  & $\circ$ &\textbf{9.537} & \textbf{12.536} & 0.099 &	0.660 &	4.321 &	0.174 &	0.899
 \\
 
 8&   & \checkmark & \checkmark &$\circ$ &$\circ$ & 9.551 & 12.539 & 0.099 & 0.662 &	4.295&	0.174&	0.900
 
 \\
 9 &   & \checkmark & \checkmark &\checkmark &$\circ$ &  9.994 &  12.722 & 0.098	& 0.675&	4.273&	0.174 &	\textbf{0.901} \\
 10 &   & \checkmark & \checkmark & \checkmark & \checkmark &  9.994 &  12.722 & \textbf{0.097}	& \textbf{0.649}&	\textbf{4.268}&	\textbf{0.173} &	\textbf{0.901} \\
 \midrule

 11 & \multirow{4}{*}{11l-seg}    & \checkmark  & $\circ$ &$\circ$  & $\circ$ &\textbf{11.062} & \textbf{15.017} & 0.097 &	0.654 &	4.272 &	0.172 &	0.902
 \\
 
 12&   & \checkmark & \checkmark &$\circ$ &$\circ$ & 11.076 & 15.020 & 0.097 & 0.653 &	4.241&	0.172&	0.903
 
 \\
 13 &   & \checkmark & \checkmark & \checkmark &$\circ$ &  11.519 &  15.203 & 0.096	& \textbf{0.634}&	4.225&	0.171 &	0.904 \\
 18 &   & \checkmark & \checkmark & \checkmark & \checkmark &  11.519 &  15.203 & \textbf{0.095}	& 0.642&	\textbf{4.199}&	\textbf{0.171} &	\textbf{0.906} \\

\bottomrule
\end{tabular}
}
\end{table*}
	
	\subsection{More Ablation Study on Decoder Components}
	
	Besides the ablation experiments on the  Flex-X-Large in scale mentioned in the main text, we also conducted ablation experiments on four other scales and added them to the following statement: To comprehensively evaluate the effectiveness of each proposed component, we conduct extensive ablation experiments across four encoder scales (11n-seg, 11s-seg, 11m-seg, and 11l-seg) on the KITTI~\cite{6248074} Eigen~\cite{eigen2015predicting} split at 640$\times$192 resolution. The results are summarized in Table~\ref{tab:ablation_decoder_efo}.
	
	\paragraph{Component-wise Analysis.}
	We systematically investigate four key components: (1) the Hierarchical Pyramid Block ($\mathcal{C}$: HPB vs. HEB), (2) the upsampling strategy ($\mathcal{U}$: dynamic upsampling vs. bilinear interpolation), (3) the prediction head design ($\mathcal{P}$: inverted head vs. conventional head), and (4) the static-dynamic decoupling mechanism (Dynamic: decoupled mask $M$ vs. fixed threshold vs. without stage-2). Each component contributes uniquely to the overall performance improvement.
	
	\paragraph{Dynamic Upsampling Effectiveness.}
	Comparing the baseline configuration (ID 1, 4, 7, 11) with dynamic upsampling variants (ID 2, 5, 8, 12), we observe consistent improvements in depth estimation accuracy across all encoder scales. Specifically, the dynamic upsampling mechanism reduces RMSE by 0.011m on 11l-seg (from 4.272 to 4.241) while introducing negligible computational overhead (only 0.014G FLOPs increase). This demonstrates that our content-adaptive upsampling strategy effectively preserves fine-grained spatial details during decoder feature propagation.
	
	\paragraph{Static-Dynamic Decoupling Impact.}
	The static-dynamic decoupling mechanism yields substantial performance gains across all model scales. On the 11n-seg encoder, enabling dynamic decoupling (ID 3 vs. ID 2) improves Sq Rel from 0.800 to 0.794 and boosts $\delta<1.25$ accuracy from 0.878 to a state-of-the-art 0.878 within the same parameter budget. More significantly, on the 11m-seg encoder (ID 10 vs. ID 9), the decoupling mechanism achieves the best performance within this scale: Abs Rel of 0.097, Sq Rel of 0.649, and $\delta<1.25$ accuracy of 0.901, representing relative improvements of 1.0\%, 3.9\%, and 0.6\% respectively over the non-decoupled variant.
	
	\paragraph{Encoder Scale Scalability.}
	Our decoder architecture demonstrates excellent scalability across different encoder capacities. For compact encoders (11n-seg and 11s-seg), the lightweight configuration ($\mathcal{C}=\circ$) with dynamic upsampling and decoupling achieves an optimal balance between efficiency and accuracy, requiring only 0.718G FLOPs and 1.522M parameters while maintaining competitive $\delta<1.25$ accuracy of 0.878. For larger encoders (11m-seg and 11l-seg), the full configuration with all components enabled consistently yields the best results. The largest model (11l-seg with all components) achieves state-of-the-art performance with Abs Rel of 0.095, RMSE of 4.199, and $\delta<1.25$ accuracy of 0.906.
	
	\paragraph{Efficiency-Accuracy Trade-off.}
	A notable observation is that our proposed components introduce minimal computational overhead while delivering significant accuracy improvements. The HPB module increases FLOPs by approximately 0.44G and parameters by 0.18M on 11m-seg and 11l-seg encoders, but contributes to consistent RMSE reductions. The dynamic upsampling module adds negligible overhead ($<$0.02G FLOPs) while effectively improving depth boundary sharpness. This favorable trade-off makes our decoder architecture suitable for both resource-constrained embedded systems and high-performance computing scenarios.
	
	\paragraph{Summary.}
	The ablation study validates the effectiveness of each proposed component. The static-dynamic decoupling mechanism consistently improves performance across all scales, while the combination of HPB, dynamic upsampling, and inverted head achieves optimal results on larger encoders. Notably, even the most lightweight configuration (11n-seg with dynamic upsampling and decoupling) achieves competitive performance with only 0.718G FLOPs, demonstrating the efficiency of our design philosophy.

	\subsection{Computational Efficiency: Cross-platform Analysis}
	\label{sec:x_model_speeds}
	
	Table~\ref{tab:speed_tab_full} extends our efficiency benchmarks to a wider range of hardware. On the desktop RTX 2080Ti, our Nano model achieves \textbf{547 FPS}, surpassing Monodepth2 by 29\%. On mobile platforms (Snapdragon 865), it reaches 13.8 FPS, nearly $3\times$ the baseline speed.
	
	The advantage persists across model scales. The Small variant balances performance and speed (487 FPS on 2080Ti; 8.1 FPS on SD865). Crucially, our largest models remain highly efficient: the XLarge variant runs at 98 FPS on a 2080Ti, outperforming MonoVit (95 FPS) despite significantly lower FLOPs. These results confirm the architecture's scalability and suitability for devices ranging from legacy GPUs to modern mobile processors.
	
	\begin{table*}[t]
		\centering
		\caption{\textbf{Speed tests at $640 \times 192$ resolution.} GPU benchmarks (RTX 4090, 2080Ti, V100) use batch size 16; mobile platforms (Snapdragon 865, 8 Elite) use batch size 1 in performance mode.}
		\setlength{\tabcolsep}{2pt}
		\label{tab:speed_tab_full}
		\resizebox{\textwidth}{!}{
			\begin{tabular}{ccccccccccccc}
				\toprule
				\multirow{2}{*}{Method} &
				\multicolumn{2}{c}{Full Model $\downarrow$} &
				\multicolumn{5}{c}{Latency (ms) $\downarrow$} &
				\multicolumn{5}{c}{Speed (FPS) $\uparrow$} \\
				\cmidrule(lr){2-3} \cmidrule(lr){4-8} \cmidrule(lr){9-13}
				& Params(M) & FLOPs(G) & 4090 & 2080Ti & V100 & SD 865 & SD 8Elite & 4090 & 2080Ti & V100 & SD 865 & SD 8Elite \\ 
				\midrule
				Monodepth2 & 14.329 & 8.038 & 0.7 & 2.4 & 1.9 & 207 & 154 & 1376 & 424 & 536 & 4.8 & 6.5 \\
				Lite-Mono-Tiny & 2.143 & 2.842 & 1.0 & 3.6 & 3.2 & 133 & 62 & 1048 & 271 & 312 & 7.5 & 15.9 \\
				Lite-Mono-Small & 2.472 & 4.746 & 1.2 & 4.2 & 3.3 & 178 & 93 & 816 & 238 & 298 & 5.6 & 10.7 \\
				Lite-Mono & 3.069 & 5.032 & 1.3 & 4.0 & 3.4 & 185 & 97 & 781 & 247 & 295 & 5.4 & 10.3 \\
				Lite-Mono-8M & 8.766 & 11.212 & 1.7 & 5.3 & 4.5 & 324 & 180 & 592 & 189 & 223 & 3.1 & 5.6 \\
				MonoVit & 33.631 & 59.679 & 3.8 & 10.5 & 7.9 & 767 & 532 & 265 & 95 & 126 & 1.3 & 1.9 \\
				\textbf{Flex-Nano (Ours)} & \textbf{1.522} & \textbf{0.718} & \textbf{0.6} & \textbf{1.8} & \textbf{1.3} & \textbf{72} & \textbf{26} & \textbf{1583} & \textbf{547} & \textbf{774} & \textbf{13.8} & \textbf{37.6} \\
				\textbf{Flex-Small (Ours)} & 6.059 & 2.776 & 0.8 & 2.0 & 1.6 & 123 & 53 & 1215 & 487 & 606 & 8.1 & 18.6 \\
				\textbf{Flex-Medium (Ours)} & 12.722 & 9.994 & 1.9 & 6.3 & 4.3 & 342 & 172 & 510 & 160 & 231 & 3.0 & 5.8 \\
				\textbf{Flex-Large (Ours)} & 15.203 & 11.519 & 2.1 & 6.5 & 4.6 & 373 & 194 & 467 & 153 & 216 & 2.7 & 5.2 \\
				\textbf{Flex-X-Large (Ours)} & 32.269 & 24.563 & 3.1 & 10.0 & 7.6 & 615 & 334 & 321 & 98 & 130 & 1.6 & 3.0 \\
				\bottomrule
			\end{tabular}
		}
	\end{table*}

	\subsection{Comparison with Foundation Depth Models}
	\label{sec:x_comparison_with_foundation}
	
	As introduced in the manuscript, foundation depth models (e.g., Depth Anything V1~\cite{yang2024depth}, DA2~\cite{yang2024depth_2}) differ from self-supervised monocular depth estimation methods~\cite{godard2019digging,zhang2023lite,watson2021temporal} in two key aspects of the KITTI~\cite{6248074} evaluation protocol: depth alignment and ground truth selection.
	
	\textbf{1) Depth Alignment.}
As detailed in the manuscript, self-supervised methods typically use median scaling to align predicted disparity to metric depth, whereas foundation models adopt least-squares (LS) alignment in the disparity space. This affine alignment provides more degrees of freedom than per-image median scaling, yielding inherently lower error metrics for the same predictions. 
	
\begin{figure}[t]
	\centering
	\includegraphics[width=0.48\textwidth]{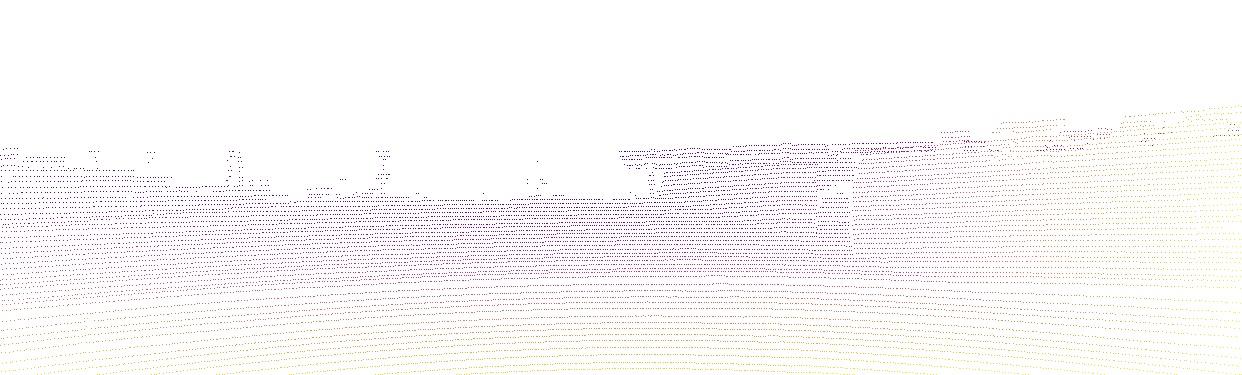}
	\hfill
	\includegraphics[width=0.48\textwidth]{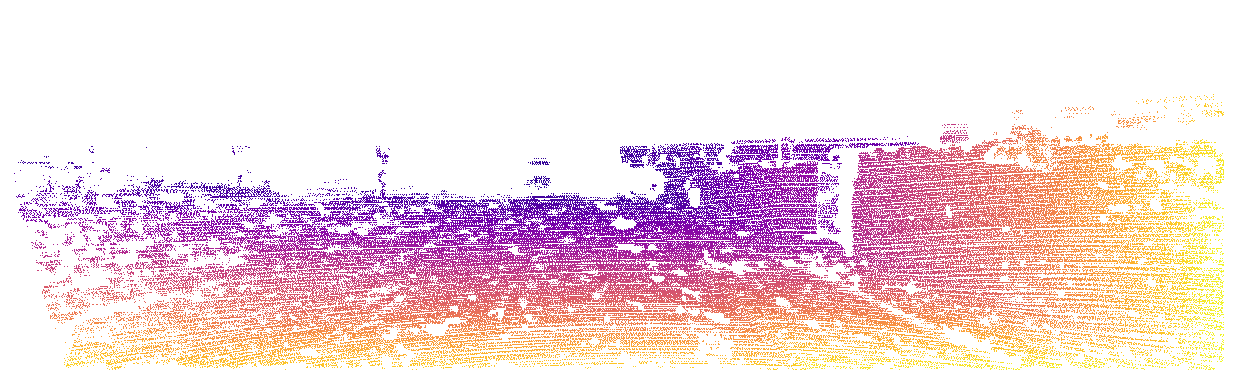}
	\caption{\textbf{Comparison of KITTI~\cite{6248074} ground truth density.} Left: self-supervised methods typically  use GT with sparse reprojected LiDAR points. Right: improved GT~\cite{uhrig2017sparsity} with denser depth from multi-frame stereo completion. }
	\label{fig:gt_compare}
\end{figure}
	
	\textbf{2) Ground Truth.}
	The self-supervised methods often use KITTI~\cite{6248074} Eigen split~\cite{eigen2015predicting} with reprojected LiDAR points as ground truth, which suffers from sparsity and does not handle moving objects.
	The improved Ground Truth~\cite{uhrig2017sparsity} uses 5 consecutive frames with stereo completion to handle dynamic objects, covering 652 of 697 Eigen~\cite{eigen2015predicting} test frames (93\%). As illustrated in Fig.~\ref{fig:gt_compare}, the improved GT provides a substantially denser reference and therefore a more comprehensive evaluation.
	
	Under this unified protocol, Table~\ref{tab:foundation_comparison} reports the quantitative comparison between DA2 (zero-shot, official inference pipeline) and Flex-X-Large (self-supervised on KITTI~\cite{6248074} only).
	For completeness, we also include DA2 results reported in~\cite{yang2024depth_2} (marked with $\ddagger$) and our reproduced results across multiple input resolutions and model scales.
	
	
	\begin{table}[t]
		\centering
		\caption{\textbf{Comparison with foundation depth models on KITTI~\cite{6248074} Eigen~\cite{eigen2015predicting} split.}
			$\ddagger$: reported in~\cite{yang2024depth_2}; DA2 results are reproduced using official official pipeline and relative-depth weights,
			Med: median scaling; LS: least-squares alignment;
			S-GT: Sparse LiDAR GT; Imp-GT: improved Ground Truth~\cite{uhrig2017sparsity}.}
		\label{tab:foundation_comparison}
		\small
		\setlength{\tabcolsep}{3.5pt}
		\resizebox{\textwidth}{!}{
			\begin{tabular}{lcccccccccccc}
				\toprule
				\multicolumn{1}{l}{\textit{Method}} & \multicolumn{2}{c}{\textit{Model}} & \multicolumn{2}{c}{\textit{Protocol}} & \multicolumn{1}{c}{\textit{GT}} & \multicolumn{7}{c}{\textit{Metrics}} \\
				\cmidrule(lr){2-3} \cmidrule(lr){4-5} \cmidrule(lr){6-6} \cmidrule(lr){7-13}
				& Params & GFLOPs & Res. & Align & Type & Abs Rel & Sq Rel & RMSE & RMSE$_{\log}$ & $\delta<1.25$ & $\delta<1.25^2$ & $\delta<1.25^3$ \\
				\midrule
				\multicolumn{13}{l}{\textit{Foundation models (zero-shot)}} \\
				\midrule
				DA2 (ViT-L)$^\ddagger$ & 335M & --- & --- & LS & Imp-GT & 0.074 & --- & --- & --- & 0.947 & --- & --- \\
				DA2 (ViT-S)$^\ddagger$ & 25M & --- & --- & LS & Imp-GT & 0.078 & --- & --- & --- & 0.939 & --- & --- \\
				DA2 (ViT-L) & 335M & 1028 & 1246$\times$378 & LS & Imp-GT & 0.070 & 0.376 & 3.178 & 0.104 & \textbf{0.956} & 0.992 & \textbf{0.998} \\
				DA2 (ViT-L) & 335M & 1947 & 1722$\times$518 & LS & Imp-GT & 0.070 & 0.347 & \textbf{3.083} & 0.104 & \textbf{0.956} & \textbf{0.993} & \textbf{0.998} \\
				DA2 (ViT-S) & 25M & 137 & 1722$\times$518 & LS & Imp-GT & 0.077 & 0.432 & 3.441 & 0.114 & 0.944 & 0.990 & 0.997 \\
				DA2 (ViT-S) & 25M & 73 & 1246$\times$378 & LS & Imp-GT & 0.077 & 0.461 & 3.502 & 0.114 & 0.944 & 0.989 & 0.997 \\
				DA2 (ViT-S) & 25M & 19 & 644$\times$196 & LS & Imp-GT & 0.110 & 0.814 & 4.941 & 0.160 & 0.881 & 0.972 & 0.993 \\
				DA2 (ViT-L) & 335M & 276 & 644$\times$196 & LS & Imp-GT & 0.092 & 0.578 & 4.114 & 0.136 & 0.915 & 0.983 & 0.996 \\
				\midrule
				\multicolumn{13}{l}{\textit{Self-supervised methods}} \\
				\midrule
				Flex-X-Large (Ours) & 32M & 25 & 640$\times$192 & Med & Imp-GT & 0.068 & 0.320 & 3.136 & 0.106 & 0.949 & 0.991 & 0.998 \\
				Flex-X-Large (Ours) & 32M & 25 & 640$\times$192 & LS & S-GT & 0.088 & 0.596 & 4.285 & 0.166 & 0.907 & 0.968 & 0.985 \\
				Flex-X-Large (Ours) & 32M & 25 & 640$\times$192 & LS & Imp-GT & \textbf{0.063} & \textbf{0.328} & \textbf{3.188} & \textbf{0.102} & 0.952 & 0.991 & \textbf{0.998} \\
				Flex-X-Large (Ours) & 32M & 25 & 640$\times$192 & Med & S-GT & 0.093 & 0.605 & 4.114 & 0.167 & 0.910 & 0.969 & 0.985 \\
				\bottomrule
			\end{tabular}
		}
	\end{table}
	
	Beyond the aggregated metrics, we further examine the qualitative behaviours of the two paradigms in Figs.~\ref{fig:depth_compare_1} and~\ref{fig:depth_compare_2}, where each column corresponds to the same scene and each row corresponds to a different model--resolution configuration.
	These visualisations are not intended as isolated evidence but as a complement to Table~\ref{tab:foundation_comparison}: they reveal failure modes that are averaged out by the metrics, and conversely they are anchored by the quantitative gap reported above.

	
	

	\usetikzlibrary{positioning,calc}
	\newlength{\blx}\newlength{\bly}
	\newlength{\trx}\newlength{\try}
	\newlength{\magw}\newlength{\magh}
	\newlength{\nodeW}\newlength{\nodeH}
	\newlength{\mimgW}\newlength{\mimgH}
	\newlength{\offX}\newlength{\offY}
	\newlength{\nswx}\newlength{\nswy}
	\newlength{\nnex}\newlength{\nney}
	
	\newcommand{\dw}{0.32}
	\newcommand{\magfactor}{4}
	
	\newcommand{\hboxdraw}[6]{%
		\path (#1.south west) -- (#1.south east) coordinate[pos=#2]   (#1-hbx);
		\path (#1.south west) -- (#1.south east) coordinate[pos={#2+#4}] (#1-hbx1);
		\path (#1.south west) -- (#1.north west) coordinate[pos=#3]   (#1-hby);
		\path (#1.south west) -- (#1.north west) coordinate[pos={#3+#5}] (#1-hby1);
		\coordinate (#1-bl) at (#1-hbx |- #1-hby);
		\coordinate (#1-tr) at (#1-hbx1 |- #1-hby1);
		\draw[red, thick] (#1-bl) rectangle (#1-tr);
		\pgfextractx{\nswx}{\pgfpointanchor{#1}{south west}}%
		\pgfextracty{\nswy}{\pgfpointanchor{#1}{south west}}%
		\pgfextractx{\nnex}{\pgfpointanchor{#1}{north east}}%
		\pgfextracty{\nney}{\pgfpointanchor{#1}{north east}}%
		\pgfmathsetlength{\nodeW}{\nnex-\nswx}%
		\pgfmathsetlength{\nodeH}{\nney-\nswy}%
		\pgfextractx{\blx}{\pgfpointanchor{#1-bl}{center}}%
		\pgfextracty{\bly}{\pgfpointanchor{#1-bl}{center}}%
		\pgfextractx{\trx}{\pgfpointanchor{#1-tr}{center}}%
		\pgfextracty{\try}{\pgfpointanchor{#1-tr}{center}}%
		\pgfmathsetlength{\magw}{(\trx-\blx)*\magfactor}%
		\pgfmathsetlength{\magh}{(\try-\bly)*\magfactor}%
		\coordinate (#1-mag-bl) at ([xshift=2pt, yshift={\nodeH-\magh}]#1.south east);
		\pgfmathsetlength{\mimgW}{\magfactor*\nodeW}%
		\pgfmathsetlength{\mimgH}{\magfactor*\nodeH}%
		\pgfmathsetlength{\offX}{-#2*\mimgW}%
		\pgfmathsetlength{\offY}{-#3*\mimgH}%
		\begin{scope}
			\clip (#1-mag-bl) rectangle ++(\magw,\magh);
			\node[inner sep=0, anchor=south west] at ([xshift=\offX, yshift=\offY]#1-mag-bl)
			{\includegraphics[width=\mimgW]{#6}};
		\end{scope}
		\draw[red, thick] (#1-mag-bl) rectangle ++(\magw,\magh);
		\draw[-stealth, red, thick] (#1-tr) -- ([yshift=\magh]#1-mag-bl);
	}

	\newcommand{\dwI}{0.32}
	\newcommand{\magfactorI}{3}
	\newcommand{\framexAI}{0.65}\newcommand{\frameyAI}{0.30}
	\newcommand{\framewAI}{0.05}\newcommand{\framehAI}{0.30}
	\newcommand{\framexBI}{0.63}\newcommand{\frameyBI}{0.50}
	\newcommand{\framewBI}{0.10}\newcommand{\framehBI}{0.30}
	
	\begin{figure*}[t]
		\centering
		\renewcommand{\dw}{\dwI}
		\renewcommand{\magfactor}{\magfactorI}
		
		\begin{tikzpicture}
			\node[inner sep=0, anchor=west] (lbl0) at (-0.02\textwidth,0) {\scriptsize\makebox[0.02\textwidth][r]{Input}};
			\node[inner sep=0, right=2pt of lbl0] (r0c0) {\includegraphics[width=\dw\textwidth]{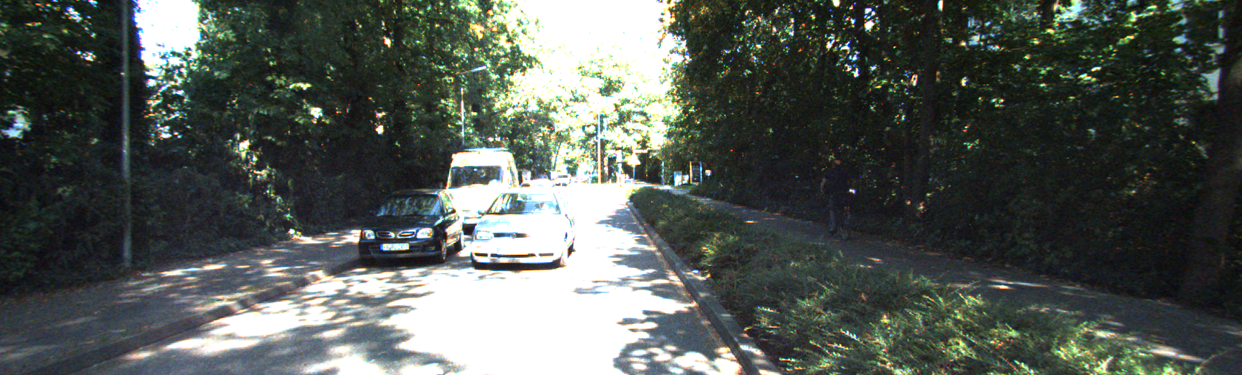}};
			\node[inner sep=0, right=20pt of r0c0] (r0c1) {\includegraphics[width=\dw\textwidth]{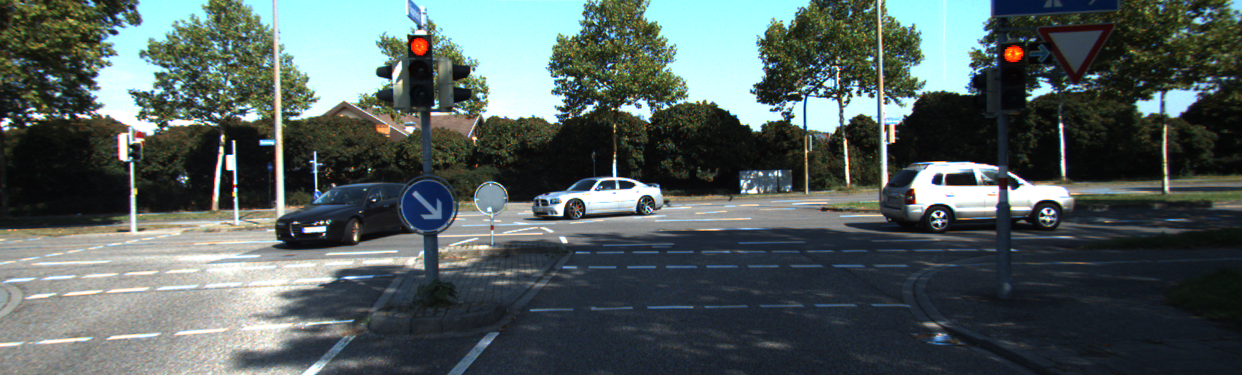}};
			\hboxdraw{r0c0}{\framexAI}{\frameyAI}{\framewAI}{\framehAI}{sec_suppl/imgs/depth_compare/000077_original}
			\hboxdraw{r0c1}{\framexBI}{\frameyBI}{\framewBI}{\framehBI}{sec_suppl/imgs/depth_compare/000044_original}
		\end{tikzpicture}\\[1pt]
		\begin{tikzpicture}
			\node[inner sep=0, anchor=west] (lbl1) at (-0.02\textwidth,0) {\scriptsize\makebox[0.02\textwidth][r]{\shortstack[r]{DA2-L\\$1722{\times}518$}}};
			\node[inner sep=0, right=2pt of lbl1] (r1c0) {\includegraphics[width=\dw\textwidth]{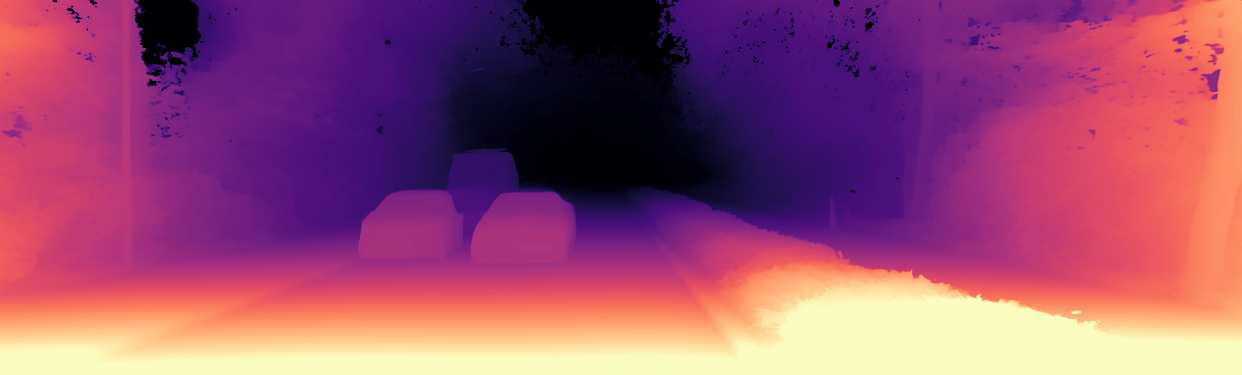}};
			\node[inner sep=0, right=20pt of r1c0] (r1c1) {\includegraphics[width=\dw\textwidth]{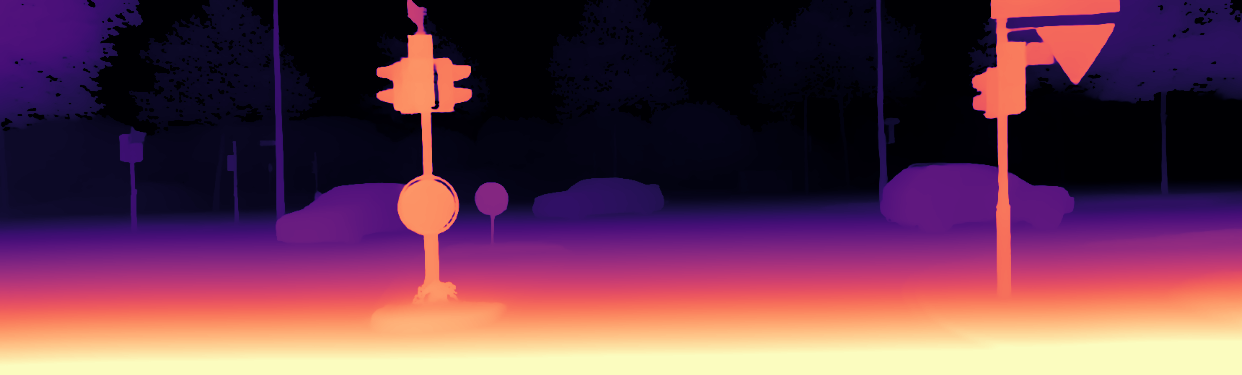}};
			\hboxdraw{r1c0}{\framexAI}{\frameyAI}{\framewAI}{\framehAI}{sec_suppl/imgs/depth_compare/000077_vitl_518}
			\hboxdraw{r1c1}{\framexBI}{\frameyBI}{\framewBI}{\framehBI}{sec_suppl/imgs/depth_compare/000044_vitl_518}
		\end{tikzpicture}\\[1pt]
		\begin{tikzpicture}
			\node[inner sep=0, anchor=west] (lbl2) at (-0.02\textwidth,0) {\scriptsize\makebox[0.02\textwidth][r]{\shortstack[r]{DA2-L\\$1246{\times}378$}}};
			\node[inner sep=0, right=2pt of lbl2] (r2c0) {\includegraphics[width=\dw\textwidth]{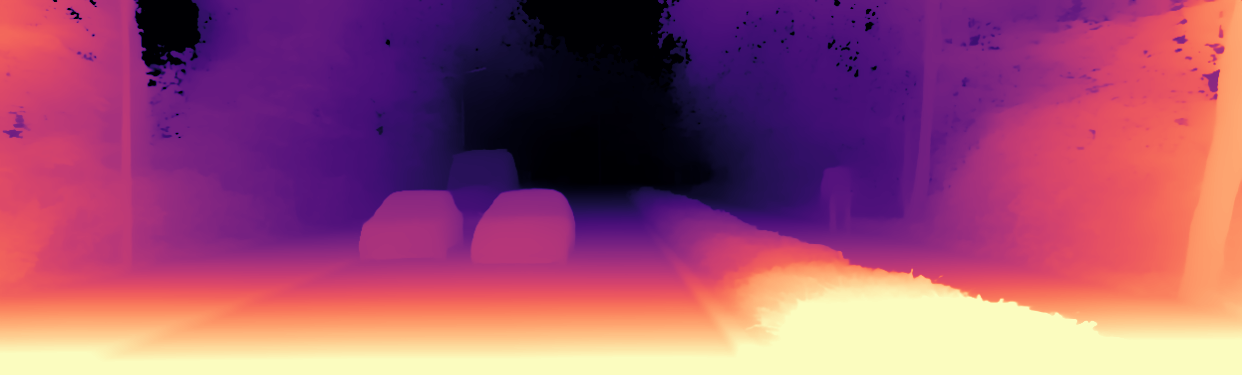}};
			\node[inner sep=0, right=20pt of r2c0] (r2c1) {\includegraphics[width=\dw\textwidth]{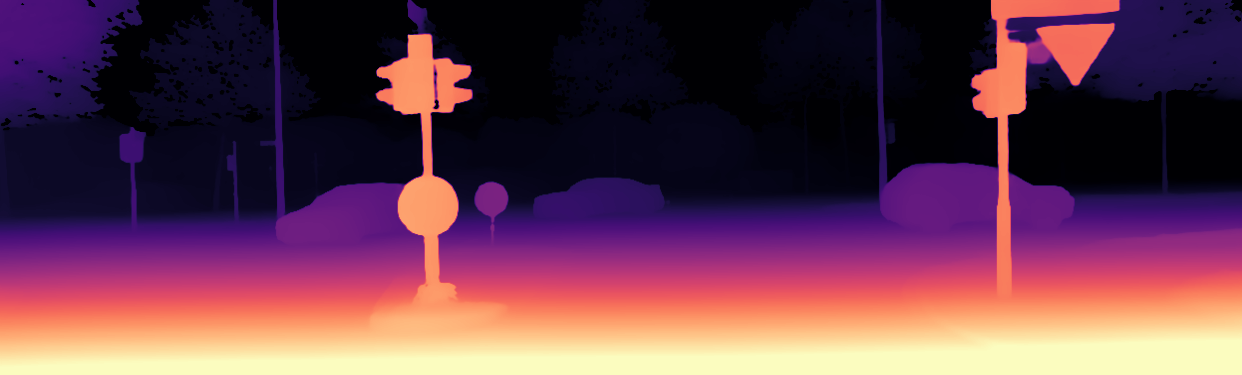}};
			\hboxdraw{r2c0}{\framexAI}{\frameyAI}{\framewAI}{\framehAI}{sec_suppl/imgs/depth_compare/000077_vitl_1218}
			\hboxdraw{r2c1}{\framexBI}{\frameyBI}{\framewBI}{\framehBI}{sec_suppl/imgs/depth_compare/000044_vitl_1218}
		\end{tikzpicture}\\[1pt]
		\begin{tikzpicture}
			\node[inner sep=0, anchor=west] (lbl3) at (-0.02\textwidth,0) {\scriptsize\makebox[0.02\textwidth][r]{\shortstack[r]{DA2-L\\$644{\times}196$}}};
			\node[inner sep=0, right=2pt of lbl3] (r3c0) {\includegraphics[width=\dw\textwidth]{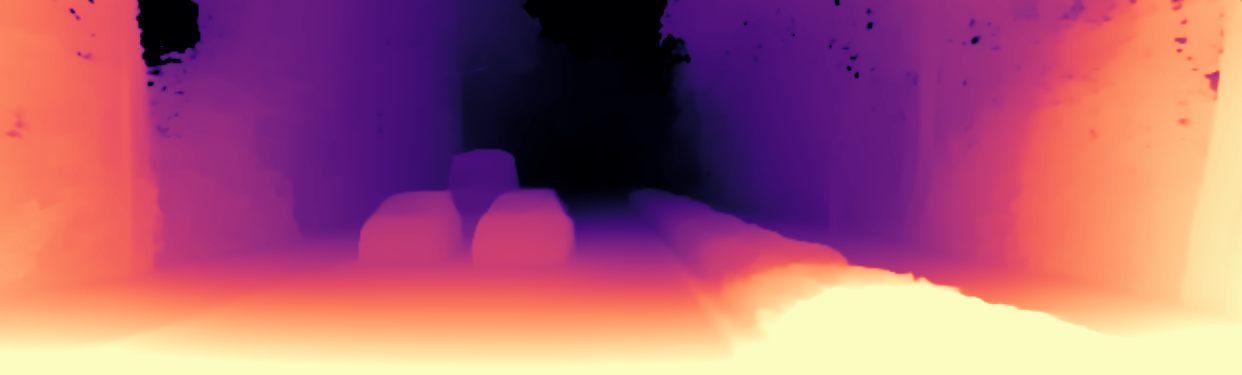}};
			\node[inner sep=0, right=20pt of r3c0] (r3c1) {\includegraphics[width=\dw\textwidth]{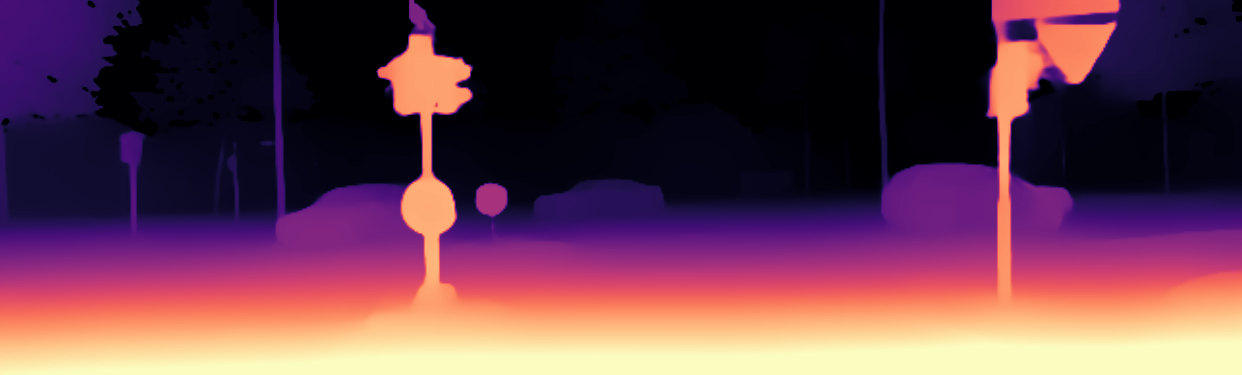}};
			\hboxdraw{r3c0}{\framexAI}{\frameyAI}{\framewAI}{\framehAI}{sec_suppl/imgs/depth_compare/000077_vitl_644}
			\hboxdraw{r3c1}{\framexBI}{\frameyBI}{\framewBI}{\framehBI}{sec_suppl/imgs/depth_compare/000044_vitl_644}
		\end{tikzpicture}\\[1pt]
		\begin{tikzpicture}
			\node[inner sep=0, anchor=west] (lbl4) at (-0.02\textwidth,0) {\scriptsize\makebox[0.02\textwidth][r]{\shortstack[r]{DA2-S\\$1722{\times}518$}}};
			\node[inner sep=0, right=2pt of lbl4] (r4c0) {\includegraphics[width=\dw\textwidth]{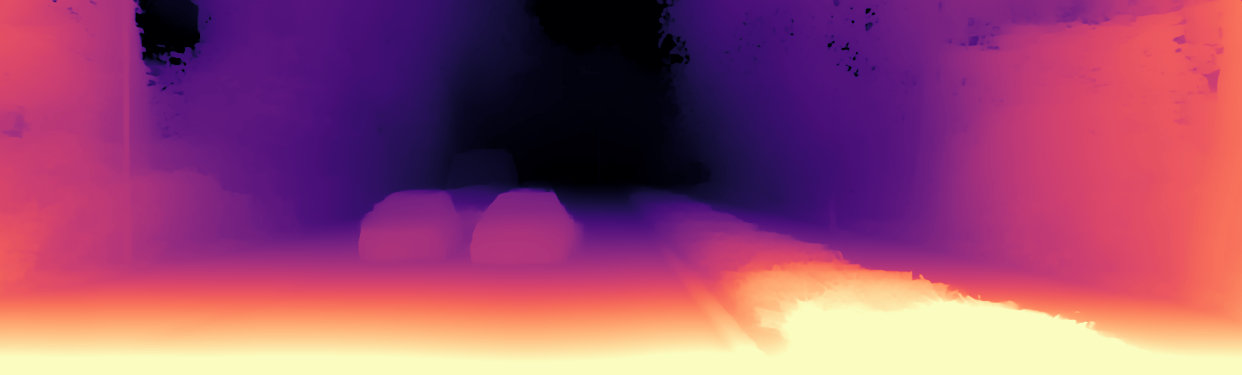}};
			\node[inner sep=0, right=20pt of r4c0] (r4c1) {\includegraphics[width=\dw\textwidth]{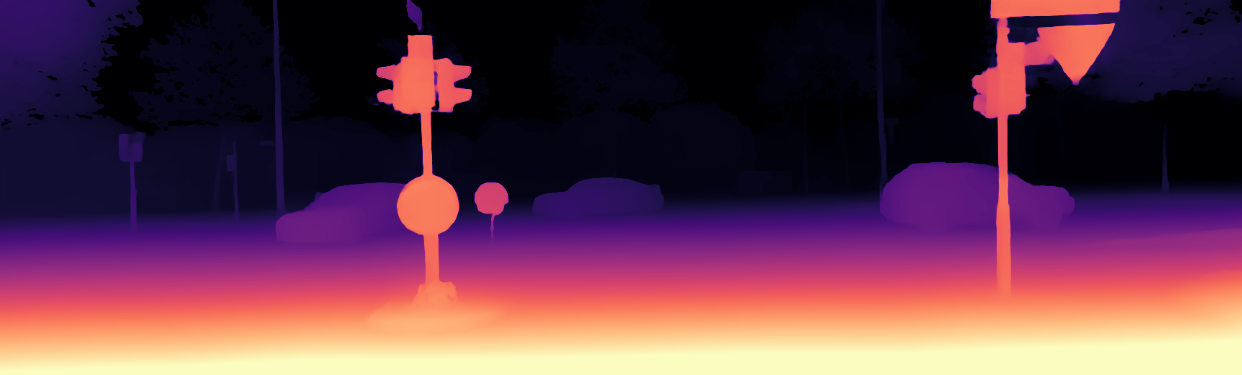}};
			\hboxdraw{r4c0}{\framexAI}{\frameyAI}{\framewAI}{\framehAI}{sec_suppl/imgs/depth_compare/000077_vits_518}
			\hboxdraw{r4c1}{\framexBI}{\frameyBI}{\framewBI}{\framehBI}{sec_suppl/imgs/depth_compare/000044_vits_518}
		\end{tikzpicture}\\[1pt]
		\begin{tikzpicture}
			\node[inner sep=0, anchor=west] (lbl5) at (-0.02\textwidth,0) {\scriptsize\makebox[0.02\textwidth][r]{\shortstack[r]{DA2-S\\$1246{\times}378$}}};
			\node[inner sep=0, right=2pt of lbl5] (r5c0) {\includegraphics[width=\dw\textwidth]{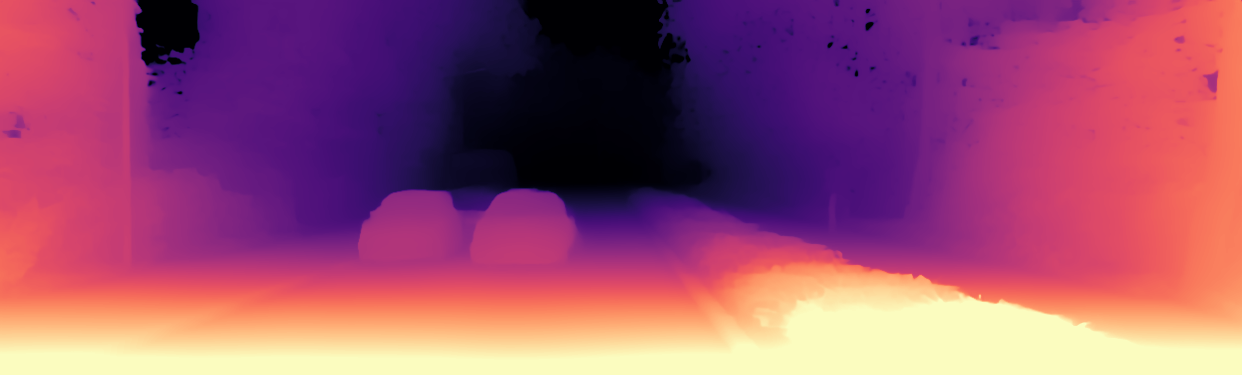}};
			\node[inner sep=0, right=20pt of r5c0] (r5c1) {\includegraphics[width=\dw\textwidth]{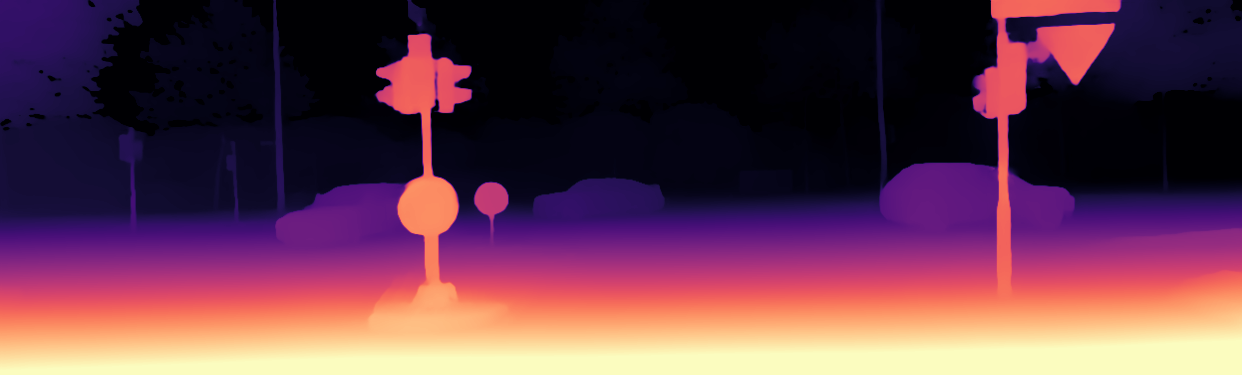}};
			\hboxdraw{r5c0}{\framexAI}{\frameyAI}{\framewAI}{\framehAI}{sec_suppl/imgs/depth_compare/000077_vits_1218}
			\hboxdraw{r5c1}{\framexBI}{\frameyBI}{\framewBI}{\framehBI}{sec_suppl/imgs/depth_compare/000044_vits_1218}
		\end{tikzpicture}\\[1pt]
		\begin{tikzpicture}
			\node[inner sep=0, anchor=west] (lbl6) at (-0.02\textwidth,0) {\scriptsize\makebox[0.02\textwidth][r]{\shortstack[r]{DA2-S\\$644{\times}196$}}};
			\node[inner sep=0, right=2pt of lbl6] (r6c0) {\includegraphics[width=\dw\textwidth]{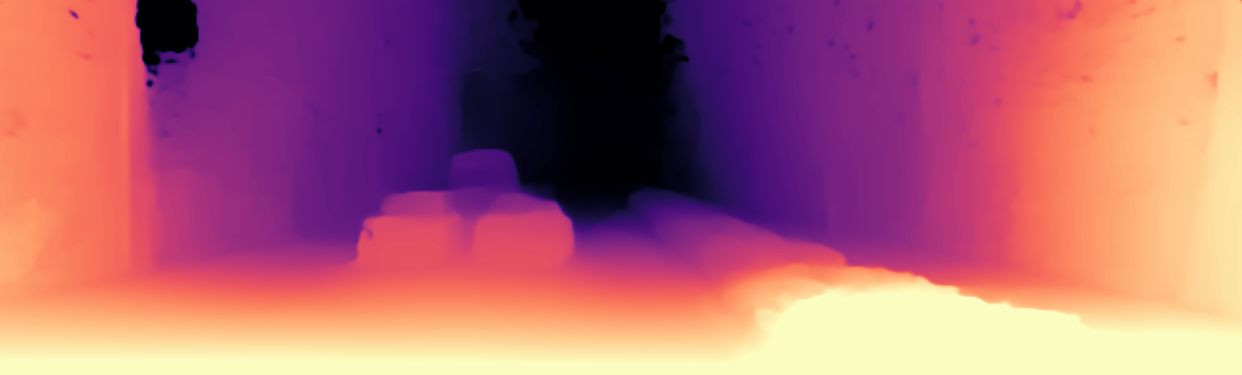}};
			\node[inner sep=0, right=20pt of r6c0] (r6c1) {\includegraphics[width=\dw\textwidth]{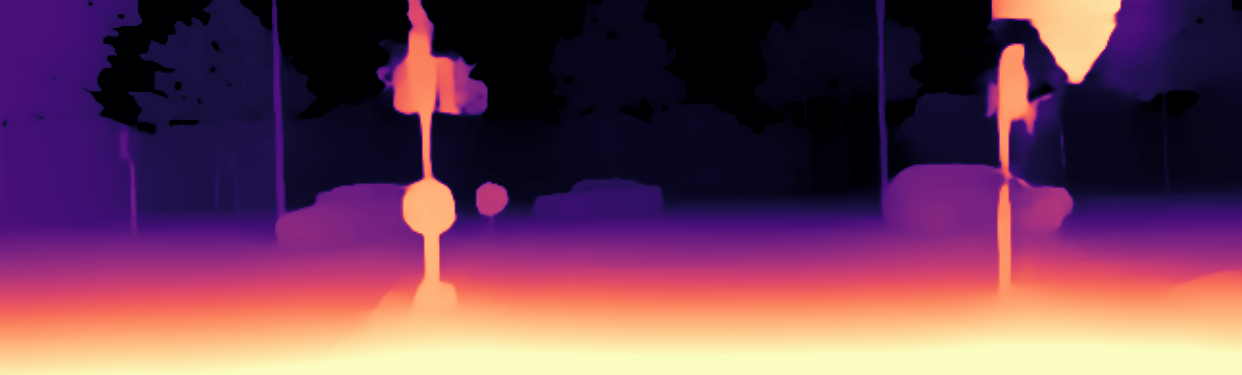}};
			\hboxdraw{r6c0}{\framexAI}{\frameyAI}{\framewAI}{\framehAI}{sec_suppl/imgs/depth_compare/000077_vits_644}
			\hboxdraw{r6c1}{\framexBI}{\frameyBI}{\framewBI}{\framehBI}{sec_suppl/imgs/depth_compare/000044_vits_644}
		\end{tikzpicture}\\[1pt]
		\begin{tikzpicture}
			\node[inner sep=0, anchor=west] (lbl7) at (-0.02\textwidth,0) {\scriptsize\makebox[0.02\textwidth][r]{\shortstack[r]{Flex-X-Large\\$640{\times}192$}}};
			\node[inner sep=0, right=2pt of lbl7] (r7c0) {\includegraphics[width=\dw\textwidth]{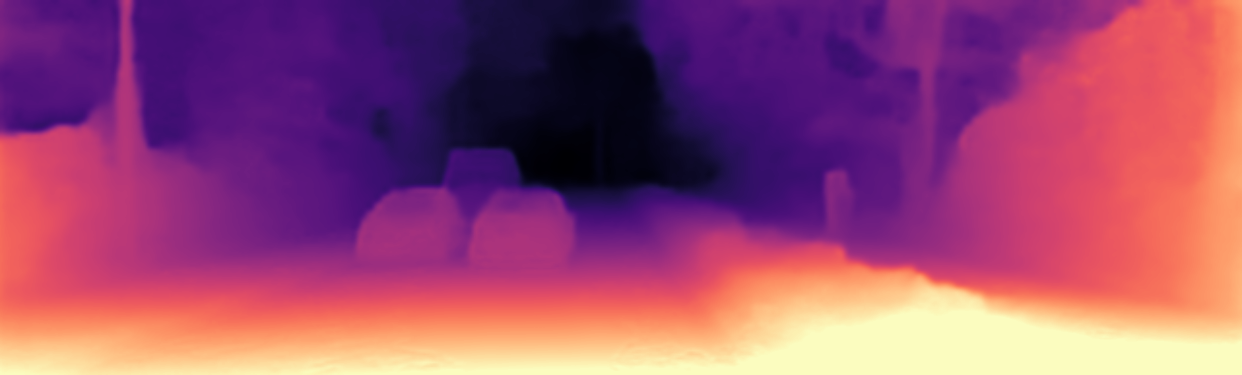}};
			\node[inner sep=0, right=20pt of r7c0] (r7c1) {\includegraphics[width=\dw\textwidth]{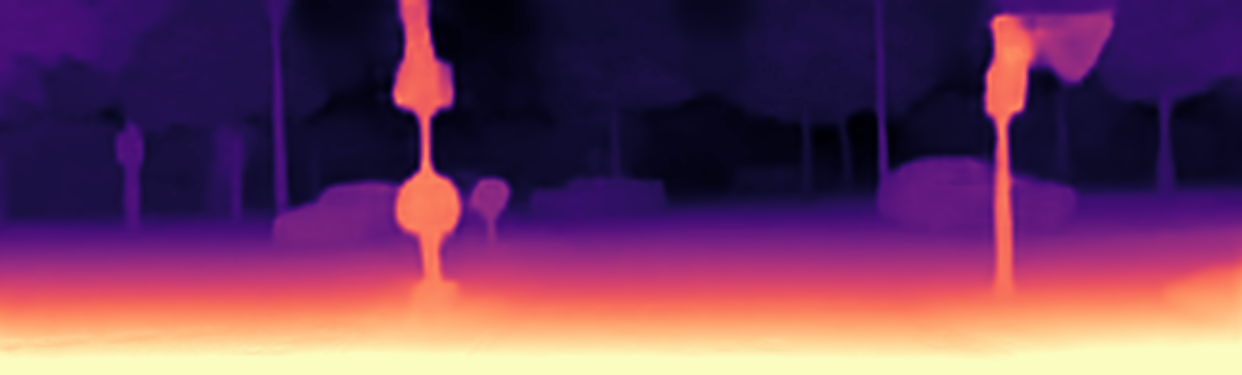}};
			\hboxdraw{r7c0}{\framexAI}{\frameyAI}{\framewAI}{\framehAI}{sec_suppl/imgs/depth_compare/000077_oursx}
			\hboxdraw{r7c1}{\framexBI}{\frameyBI}{\framewBI}{\framehBI}{sec_suppl/imgs/depth_compare/000044_oursx}
		\end{tikzpicture}\\[4pt]
		\begin{tikzpicture}
			\node[inner sep=0, anchor=west] (lblh) at (-0.02\textwidth,0) {\scriptsize\makebox[0.02\textwidth][r]{}};
			\node[inner sep=0, right=2pt of lblh] (hc0) {\scriptsize\makebox[\dw\textwidth][c]{(a)}};
			\node[inner sep=0, right=20pt of hc0] (hc1) {\scriptsize\makebox[\dw\textwidth][c]{(b)}};
		\end{tikzpicture}
		
		\caption{\textbf{Qualitative depth comparison on KITTI~\cite{6248074}.}
			Each column shows the same scene; each row shows a different model and resolution.
			\textit{Resolutions:} $1722\times518$ (short-edge resized to 518, aspect ratio preserved,following DA2~\cite{yang2024depth_2} pipeline);
			$1246\times378$ (near KITTI~\cite{6248074} native $1242\times375$, aspect preserved,dimensions divisible by DA2's patch size 14);
			$644\times196$ (close to self-supervised $640\times192$, aspect preserved, dimensions divisible by DA2's patch size 14).
			Red rectangles highlight the same region within each column; magnified insets (right) show the corresponding details at $3\times$.
			\textbf{(a)} The highlighted pedestrian is consistently missed by DA2 at all evaluated resolutions and model scales (ViT-L and ViT-S), whereas Flex-X-Large recovers a coherent depth profile for this target.
			This indicates that foundation models, despite their large capacity, can drop semantically salient objects under zero-shot generalization to KITTI~\cite{6248074}, while our self-supervised model preserves them.
			\textbf{(b)} Three distant tree trunks are present in the highlighted region.
			Flex-X-Large predicts depth for all three trunks, while DA2 (both ViT-L and ViT-S, across all resolutions) only renders the rightmost trunk and loses the other two, revealing a systematic under-segmentation of thin far-field structures.
			These observations suggest that self-supervised training on in-domain data yields more globally complete depth at far ranges despite using far fewer parameters and no external training data, complementing the strengths of large-scale pre-trained foundation models.}
		\label{fig:depth_compare_1}
	\end{figure*}
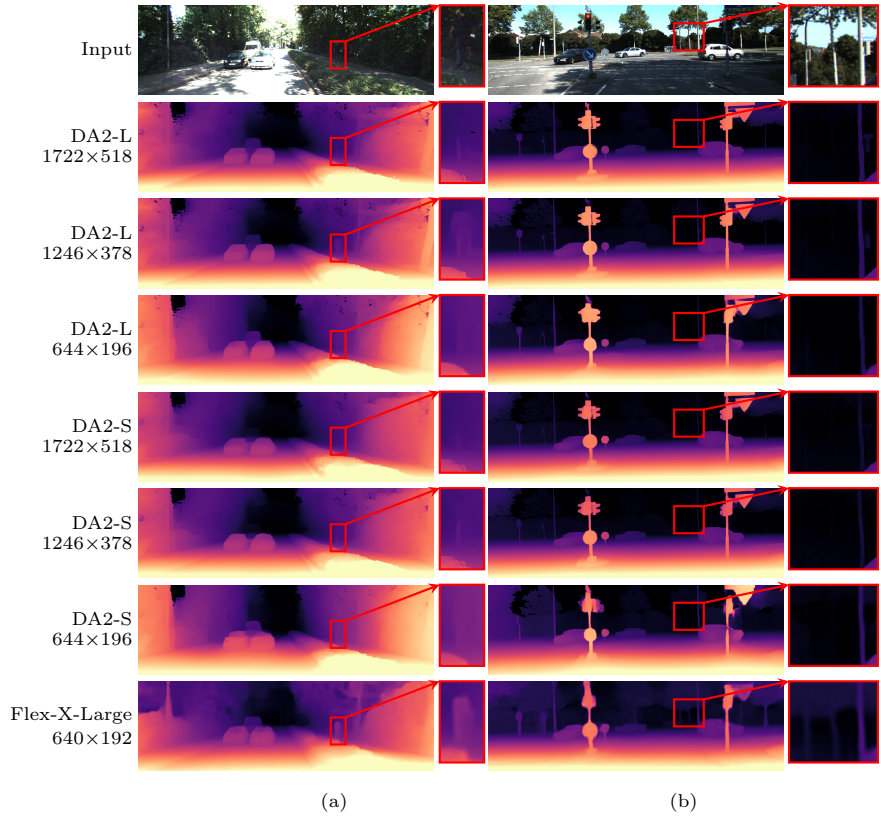

	\newcommand{\dwII}{0.32}
	\newcommand{\magfactorII}{6}
	\newcommand{\framexCI}{0.41}\newcommand{\frameyCI}{0.45}
	\newcommand{\framewCI}{0.032}\newcommand{\framehCI}{0.16}
	\newcommand{\framexDI}{0.30}\newcommand{\frameyDI}{0.44}
	\newcommand{\framewDI}{0.05}\newcommand{\framehDI}{0.16}
	
	\begin{figure*}[t]
		\centering
		\renewcommand{\dw}{\dwII}
		\renewcommand{\magfactor}{\magfactorII}
		
		\begin{tikzpicture}
			\node[inner sep=0, anchor=west] (lbl0) at (-0.02\textwidth,0) {\scriptsize\makebox[0.02\textwidth][r]{Input}};
			\node[inner sep=0, right=2pt of lbl0] (r0c0) {\includegraphics[width=\dw\textwidth]{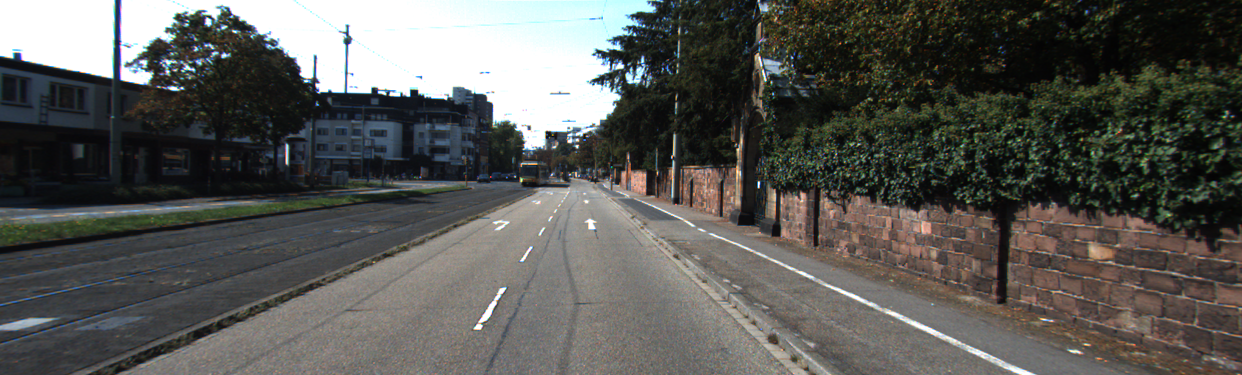}};
			\node[inner sep=0, right=25pt of r0c0] (r0c1) {\includegraphics[width=\dw\textwidth]{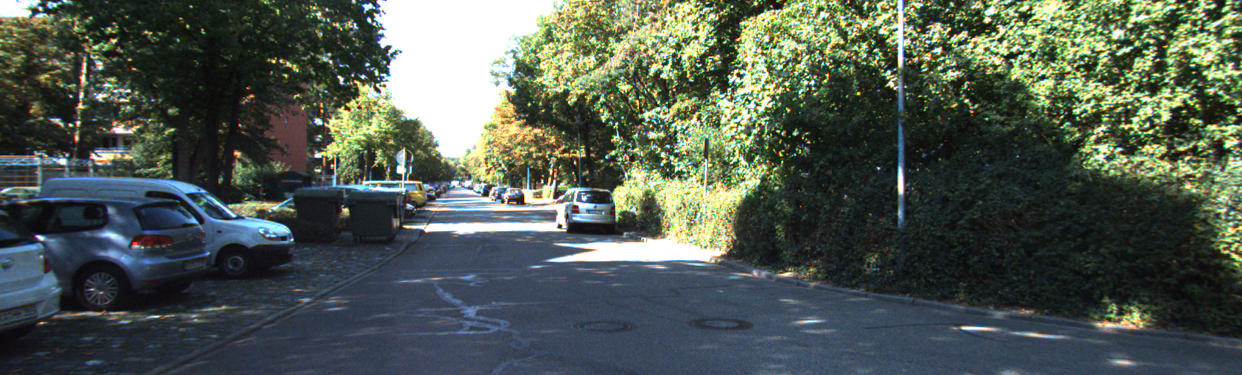}};
			\hboxdraw{r0c0}{\framexCI}{\frameyCI}{\framewCI}{\framehCI}{sec_suppl/imgs/depth_compare/000010_original}
			\hboxdraw{r0c1}{\framexDI}{\frameyDI}{\framewDI}{\framehDI}{sec_suppl/imgs/depth_compare/000099_original}
		\end{tikzpicture}\\[1pt]
		\begin{tikzpicture}
			\node[inner sep=0, anchor=west] (lbl1) at (-0.02\textwidth,0) {\scriptsize\makebox[0.02\textwidth][r]{\shortstack[r]{DA2-L\\$1722{\times}518$}}};
			\node[inner sep=0, right=2pt of lbl1] (r1c0) {\includegraphics[width=\dw\textwidth]{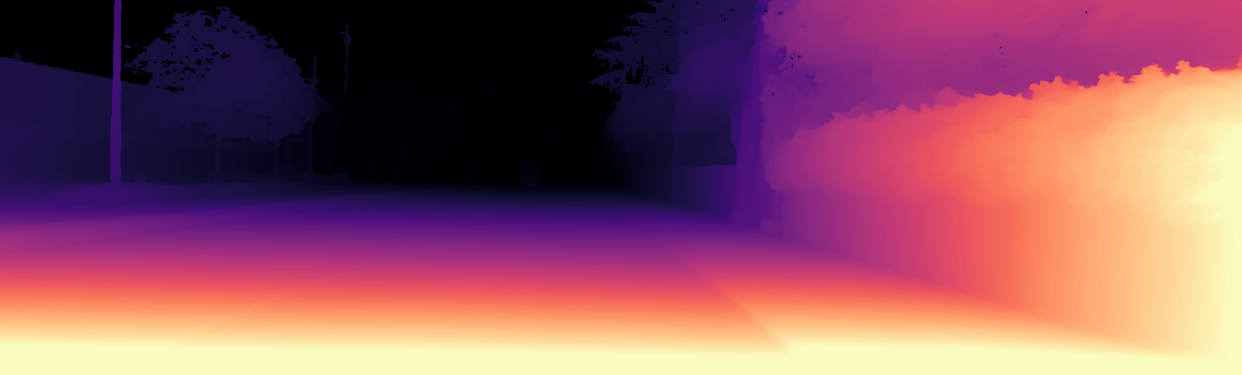}};
			\node[inner sep=0, right=25pt of r1c0] (r1c1) {\includegraphics[width=\dw\textwidth]{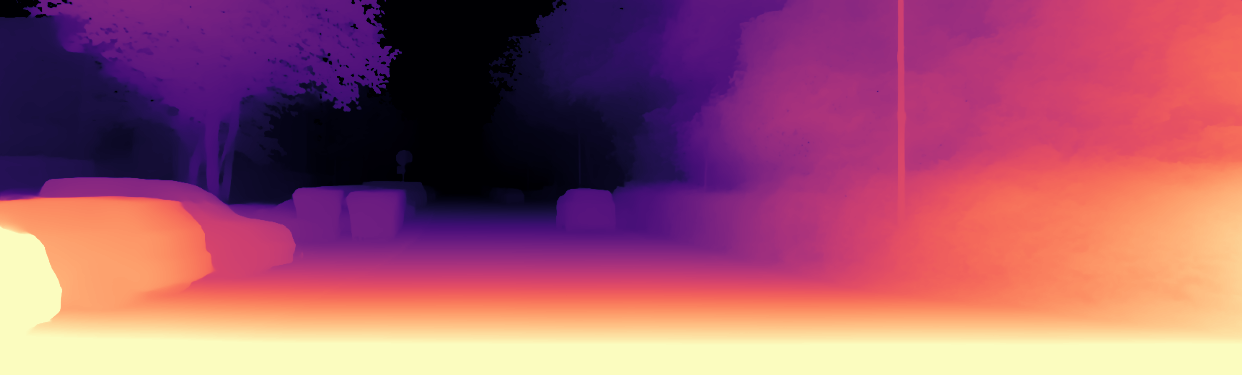}};
			\hboxdraw{r1c0}{\framexCI}{\frameyCI}{\framewCI}{\framehCI}{sec_suppl/imgs/depth_compare/000010_vitl_518}
			\hboxdraw{r1c1}{\framexDI}{\frameyDI}{\framewDI}{\framehDI}{sec_suppl/imgs/depth_compare/000099_vitl_518}
		\end{tikzpicture}\\[1pt]
		\begin{tikzpicture}
			\node[inner sep=0, anchor=west] (lbl2) at (-0.02\textwidth,0) {\scriptsize\makebox[0.02\textwidth][r]{\shortstack[r]{DA2-L\\$1246{\times}378$}}};
			\node[inner sep=0, right=2pt of lbl2] (r2c0) {\includegraphics[width=\dw\textwidth]{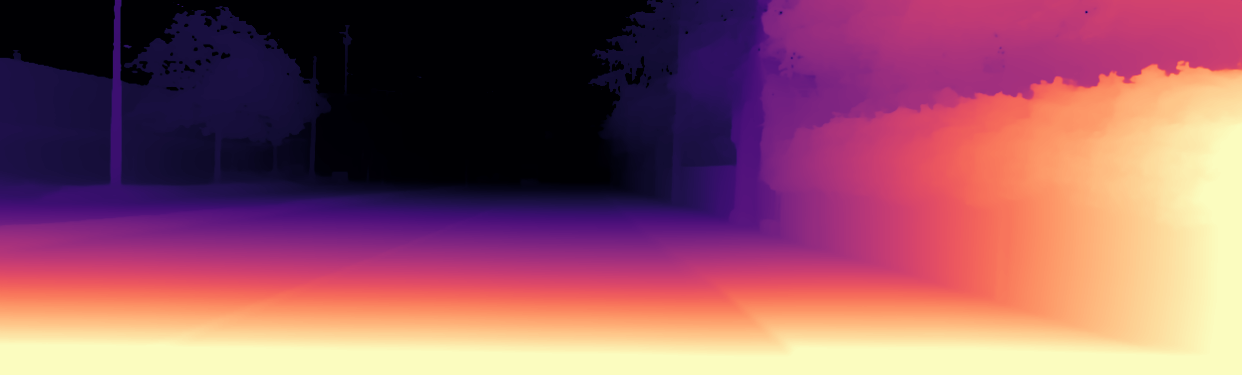}};
			\node[inner sep=0, right=25pt of r2c0] (r2c1) {\includegraphics[width=\dw\textwidth]{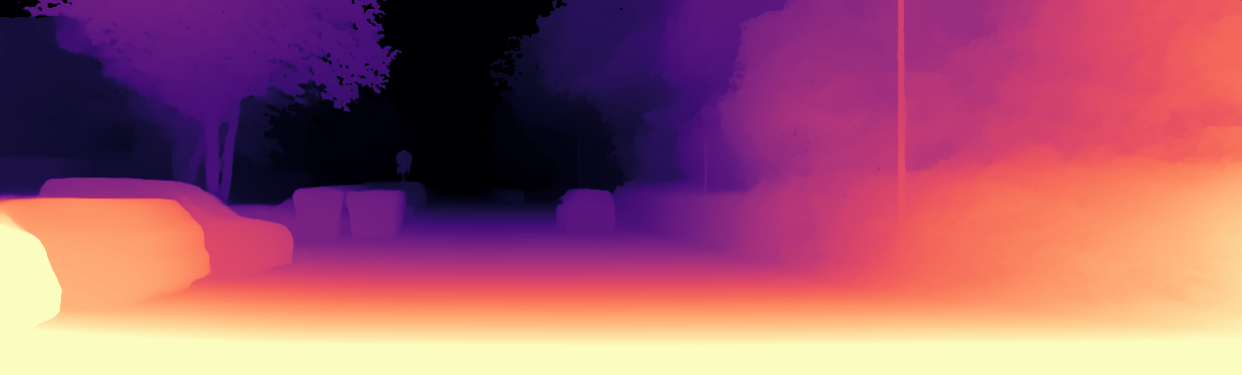}};
			\hboxdraw{r2c0}{\framexCI}{\frameyCI}{\framewCI}{\framehCI}{sec_suppl/imgs/depth_compare/000010_vitl_1218}
			\hboxdraw{r2c1}{\framexDI}{\frameyDI}{\framewDI}{\framehDI}{sec_suppl/imgs/depth_compare/000099_vitl_1218}
		\end{tikzpicture}\\[1pt]
		\begin{tikzpicture}
			\node[inner sep=0, anchor=west] (lbl3) at (-0.02\textwidth,0) {\scriptsize\makebox[0.02\textwidth][r]{\shortstack[r]{DA2-L\\$644{\times}196$}}};
			\node[inner sep=0, right=2pt of lbl3] (r3c0) {\includegraphics[width=\dw\textwidth]{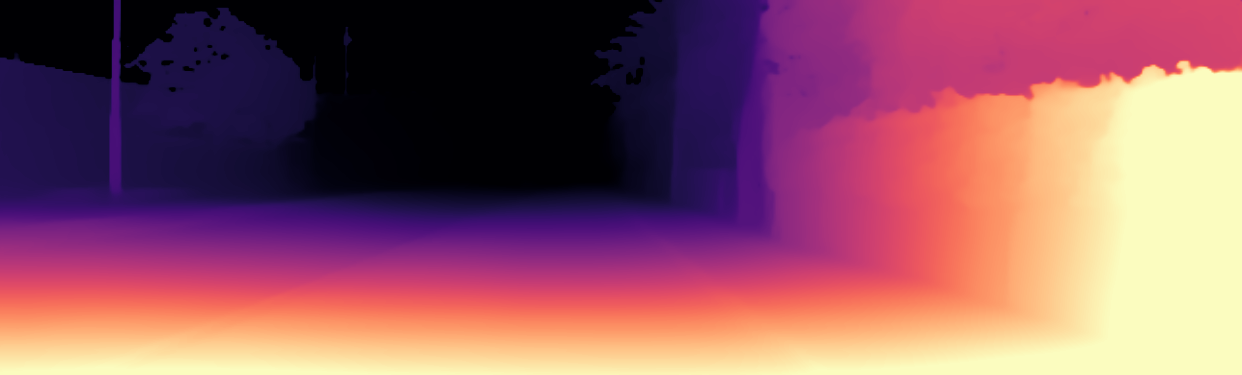}};
			\node[inner sep=0, right=25pt of r3c0] (r3c1) {\includegraphics[width=\dw\textwidth]{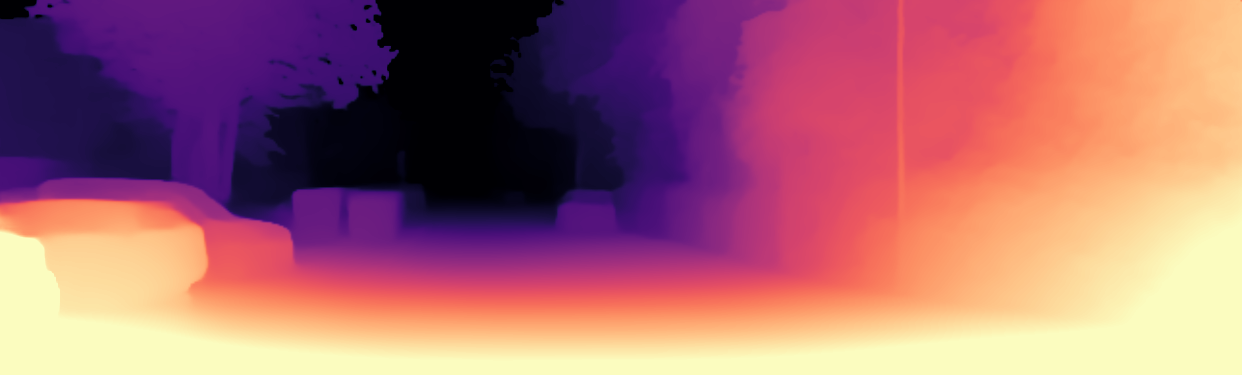}};
			\hboxdraw{r3c0}{\framexCI}{\frameyCI}{\framewCI}{\framehCI}{sec_suppl/imgs/depth_compare/000010_vitl_644}
			\hboxdraw{r3c1}{\framexDI}{\frameyDI}{\framewDI}{\framehDI}{sec_suppl/imgs/depth_compare/000099_vitl_644}
		\end{tikzpicture}\\[1pt]
		\begin{tikzpicture}
			\node[inner sep=0, anchor=west] (lbl4) at (-0.02\textwidth,0) {\scriptsize\makebox[0.02\textwidth][r]{\shortstack[r]{DA2-S\\$1722{\times}518$}}};
			\node[inner sep=0, right=2pt of lbl4] (r4c0) {\includegraphics[width=\dw\textwidth]{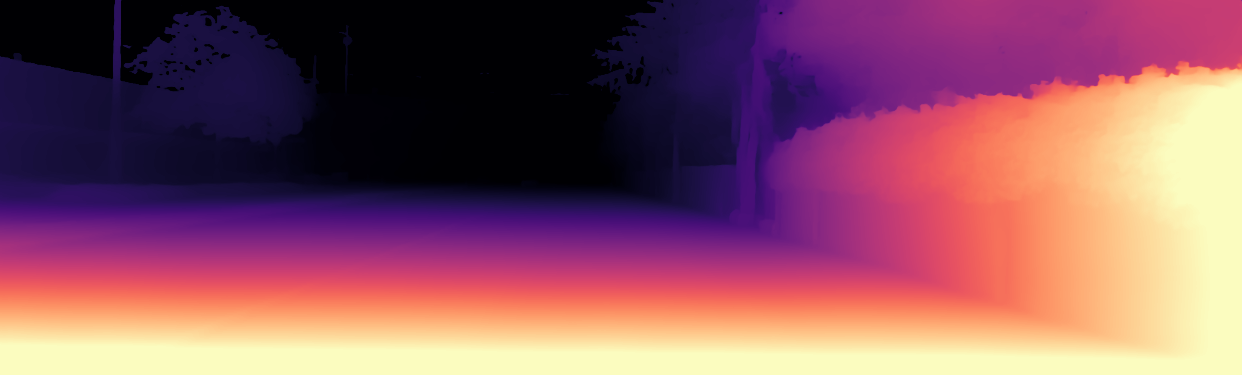}};
			\node[inner sep=0, right=25pt of r4c0] (r4c1) {\includegraphics[width=\dw\textwidth]{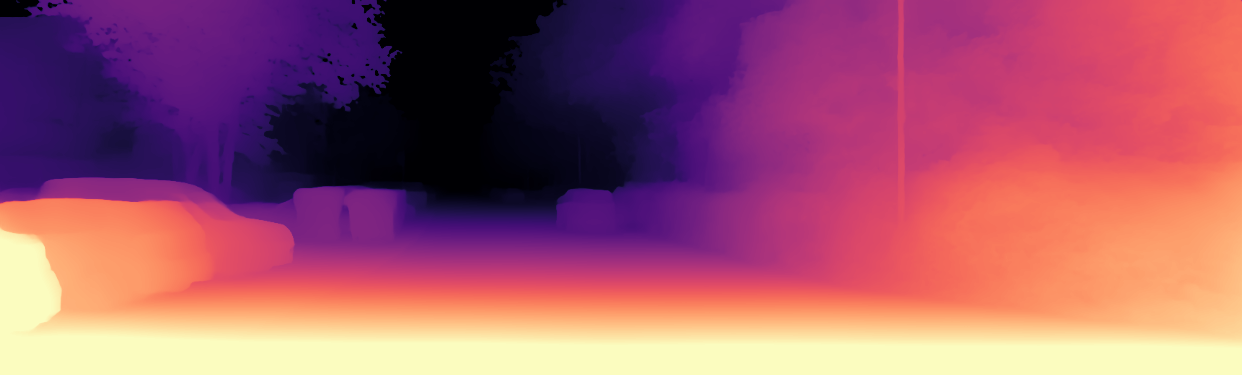}};
			\hboxdraw{r4c0}{\framexCI}{\frameyCI}{\framewCI}{\framehCI}{sec_suppl/imgs/depth_compare/000010_vits_518}
			\hboxdraw{r4c1}{\framexDI}{\frameyDI}{\framewDI}{\framehDI}{sec_suppl/imgs/depth_compare/000099_vits_518}
		\end{tikzpicture}\\[1pt]
		\begin{tikzpicture}
			\node[inner sep=0, anchor=west] (lbl5) at (-0.02\textwidth,0) {\scriptsize\makebox[0.02\textwidth][r]{\shortstack[r]{DA2-S\\$1246{\times}378$}}};
			\node[inner sep=0, right=2pt of lbl5] (r5c0) {\includegraphics[width=\dw\textwidth]{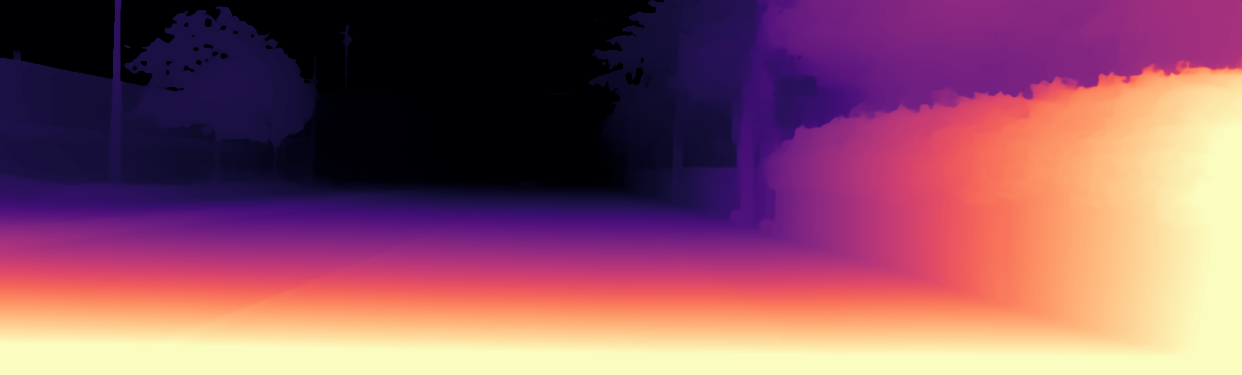}};
			\node[inner sep=0, right=25pt of r5c0] (r5c1) {\includegraphics[width=\dw\textwidth]{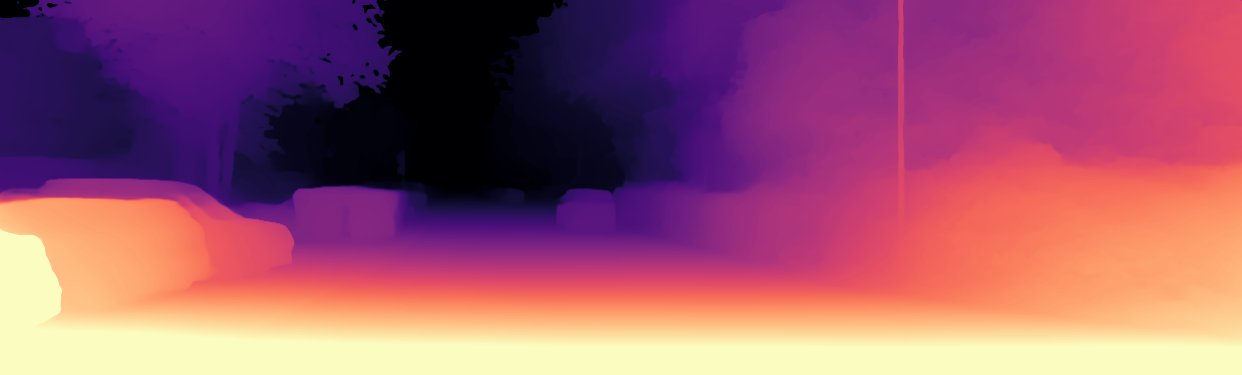}};
			\hboxdraw{r5c0}{\framexCI}{\frameyCI}{\framewCI}{\framehCI}{sec_suppl/imgs/depth_compare/000010_vits_1218}
			\hboxdraw{r5c1}{\framexDI}{\frameyDI}{\framewDI}{\framehDI}{sec_suppl/imgs/depth_compare/000099_vits_1218}
		\end{tikzpicture}\\[1pt]
		\begin{tikzpicture}
			\node[inner sep=0, anchor=west] (lbl6) at (-0.02\textwidth,0) {\scriptsize\makebox[0.02\textwidth][r]{\shortstack[r]{DA2-S\\$644{\times}196$}}};
			\node[inner sep=0, right=2pt of lbl6] (r6c0) {\includegraphics[width=\dw\textwidth]{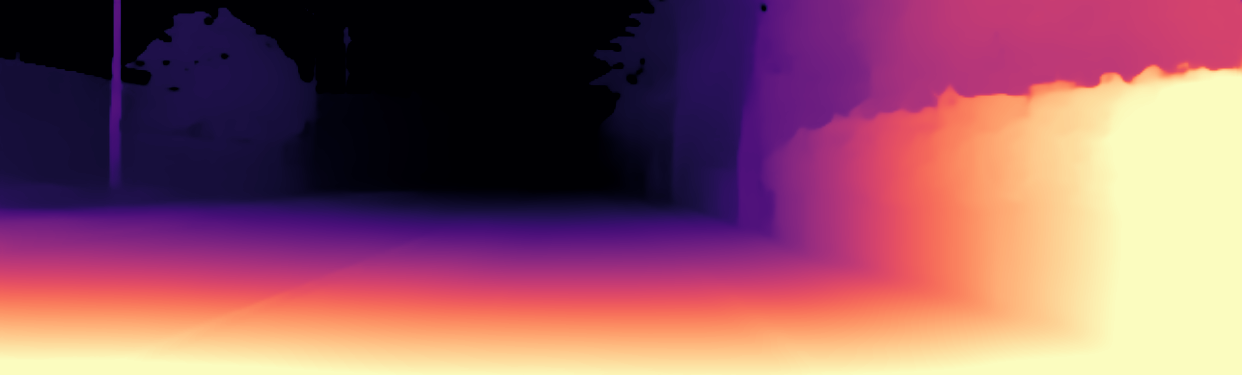}};
			\node[inner sep=0, right=25pt of r6c0] (r6c1) {\includegraphics[width=\dw\textwidth]{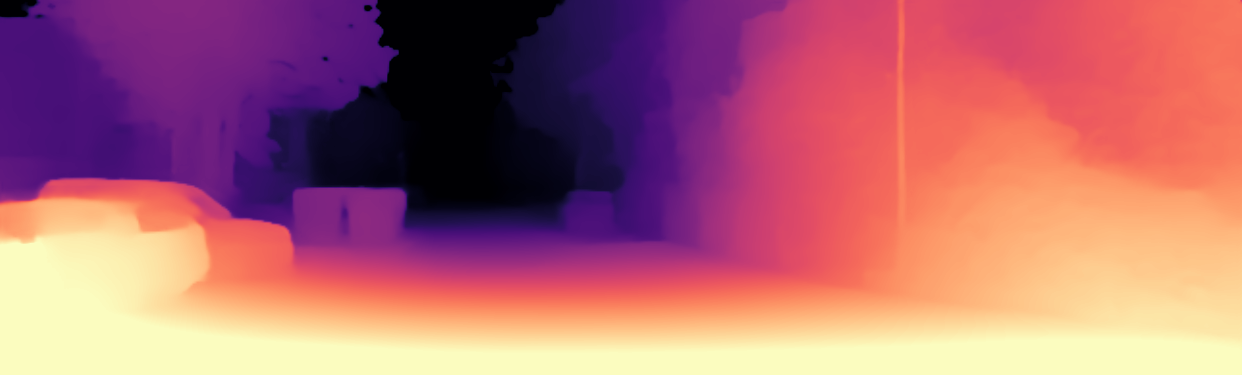}};
			\hboxdraw{r6c0}{\framexCI}{\frameyCI}{\framewCI}{\framehCI}{sec_suppl/imgs/depth_compare/000010_vits_644}
			\hboxdraw{r6c1}{\framexDI}{\frameyDI}{\framewDI}{\framehDI}{sec_suppl/imgs/depth_compare/000099_vits_644}
		\end{tikzpicture}\\[1pt]
		\begin{tikzpicture}
			\node[inner sep=0, anchor=west] (lbl7) at (-0.02\textwidth,0) {\scriptsize\makebox[0.02\textwidth][r]{\shortstack[r]{Flex-X-Large\\$640{\times}192$}}};
			\node[inner sep=0, right=2pt of lbl7] (r7c0) {\includegraphics[width=\dw\textwidth]{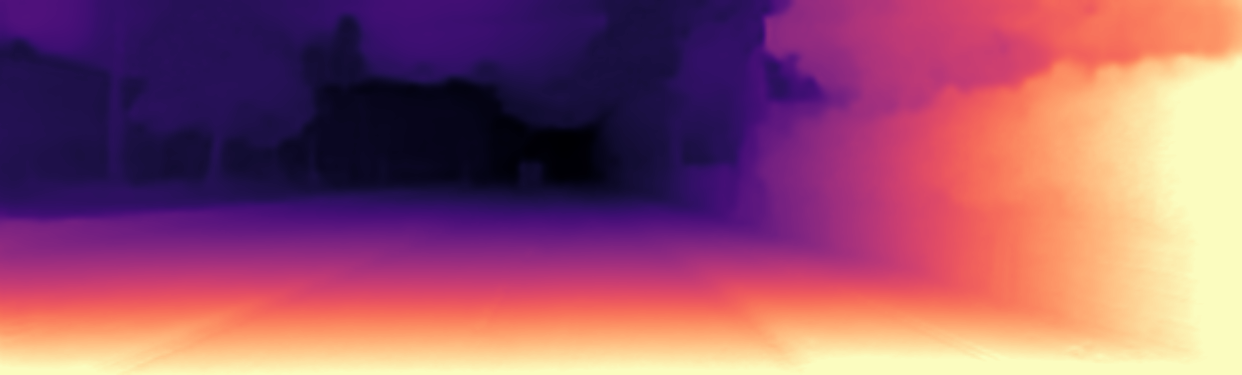}};
			\node[inner sep=0, right=25pt of r7c0] (r7c1) {\includegraphics[width=\dw\textwidth]{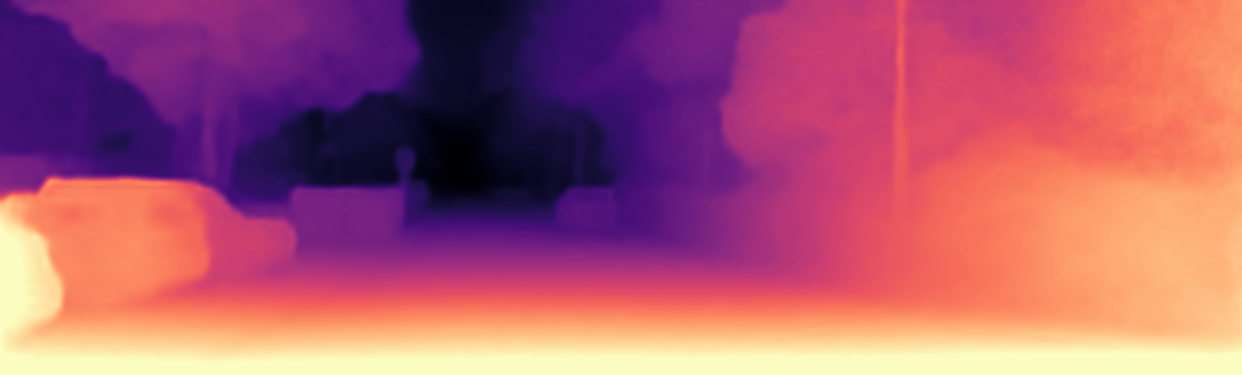}};
			\hboxdraw{r7c0}{\framexCI}{\frameyCI}{\framewCI}{\framehCI}{sec_suppl/imgs/depth_compare/000010_oursx}
			\hboxdraw{r7c1}{\framexDI}{\frameyDI}{\framewDI}{\framehDI}{sec_suppl/imgs/depth_compare/000099_oursx}
		\end{tikzpicture}\\[4pt]
		\begin{tikzpicture}
			\node[inner sep=0, anchor=west] (lblh) at (-0.02\textwidth,0) {\scriptsize\makebox[0.02\textwidth][r]{}};
			\node[inner sep=0, right=2pt of lblh] (hc0) {\scriptsize\makebox[\dw\textwidth][c]{(a)}};
			\node[inner sep=0, right=25pt of hc0] (hc1) {\scriptsize\makebox[\dw\textwidth][c]{(b)}};
		\end{tikzpicture}
		
		\caption{\textbf{Qualitative depth comparison on KITTI~\cite{6248074} (2/2).}
			Continuation of Fig.~\ref{fig:depth_compare_1} with two additional scenes.
			Red rectangles highlight the same region within each column; magnified insets (right) show the corresponding details at $6\times$.
			\textbf{(a)} A distant vehicle in the highlighted region is detected and assigned a coherent depth value only by Flex-X-Large; all DA2 variants (ViT-L and ViT-S, across all evaluated resolutions) fail to recover this object, producing flat or fragmented predictions in the corresponding area.
			This suggests that zero-shot foundation models can silently miss small, far-field traffic participants, a critical limitation for autonomous driving applications.
			\textbf{(b)} A roadside traffic sign is correctly captured only by Flex-X-Large and by DA2 ViT-L at the highest input resolution ($1722\times518$); all other DA2 configurations---including ViT-L at lower resolutions and ViT-S at every resolution---fail to assign a meaningful depth to this thin, distant structure.
			This further confirms that foundation models are sensitive to input resolution and model scale when handling thin far-field objects, whereas our self-supervised model delivers consistent predictions without such dependencies.}
		\label{fig:depth_compare_2}
	\end{figure*}

	
	\textbf{Discussion.}
	The quantitative results in Table~\ref{tab:foundation_comparison} and the qualitative comparisons in Figs.~\ref{fig:depth_compare_1}--\ref{fig:depth_compare_2} jointly provide several observations, which we summarise below.
	
	\textbf{1) Fair Comparison.}
	First, it should be emphasised that the comparison is inherently between two different paradigms.
	DA2~\cite{yang2024depth_2} is evaluated in a purely zero-shot setting without seeing any KITTI~\cite{6248074} training images, whereas Flex-X-Large is trained using only unlabeled KITTI~\cite{6248074} images without any ground-truth depth.
	The two settings reflect complementary assumptions about data availability; therefore, the gap reported below should be interpreted as a comparison between two distinct operating regimes rather than a head-to-head superiority of one paradigm over the other.
	
	\textbf{2) Quantitative.}
	Under the unified LS + improved-GT protocol, Flex-X-Large achieves an Abs Rel of 0.063, surpassing DA2 ViT-L (0.070) while requiring only 32M parameters and 25 GFLOPs, compared with 335M parameters and over 1000 GFLOPs---roughly an order-of-magnitude reduction on both axes.
	Our reproduction of DA2 using the official inference pipeline yields slightly better results than the values reported in~\cite{yang2024depth_2} (e.g., Abs Rel 0.070 vs.\ 0.074 for ViT-L at $1722\times518$),  which may suggest that the official implementation provides stronger performance than the published benchmark, or alternatively that the original paper used a more conservative evaluation protocol (or both).
	At a comparable input resolution ($644\times196$), Flex-X-Large (Abs Rel 0.063) also compares favourably to DA2 ViT-S (0.110) with a similar parameter budget (32M vs.\ 25M), suggesting that in-domain self-supervision can compensate for the absence of large-scale pre-training at this operating point.
	
	\textbf{3) Qualitative.}
	The qualitative comparisons in Figs.~\ref{fig:depth_compare_1}--\ref{fig:depth_compare_2} further reveal complementary behaviours between the two paradigms.
	Foundation models preserve impressive local structures and fine details, benefiting from large-scale pre-training, especially in the near field.
	However, under zero-shot transfer they occasionally fail to recover distant or thin objects, including pedestrians, vehicles, traffic signs, and tree trunks, as illustrated in the highlighted regions.
	In contrast, Flex-X-Large produces more complete far-range predictions after adapting to the target domain through self-supervision, recovering targets that are silently dropped by all DA2 variants regardless of model scale or input resolution.
	
	\textbf{4) Take-home Message.}
	Overall, these results suggest that foundation models and self-supervised methods are complementary rather than mutually exclusive.
	While foundation models provide strong zero-shot generalisation and rich local details, self-supervised learning remains an attractive solution for domain-specific deployment, offering competitive accuracy with substantially lower computational cost after adaptation using only unlabeled target-domain data.

	\subsection{Additional Cross-Dataset Generalization Results}
	\label{subsec:x_additional_generalization}
	To further demonstrate the robustness and zero-shot generalization capability of our proposed method, we conduct extensive cross-dataset evaluations. Specifically, we train our model (\textbf{Flex-X-Large}) exclusively on the KITTI~\cite{6248074} dataset and directly evaluate its performance on three distinct and challenging outdoor driving and scene datasets---Cityscapes~\cite{cordts2016cityscapes}, Make3D~\cite{saxena2008make3d}, and DDAD~\cite{guizilini20203d}---without any fine-tuning or domain adaptation. 
	
	Table~\ref{tab:zeroshot} provides a detailed quantitative comparison with the state-of-the-art method, DSI-MonoVit~\cite{zhang2025depth}. The results clearly manifest that our method exhibits superior cross-domain transferability and structural consistency across diverse environments:
		\begin{itemize}
		\item \textbf{Cityscapes (C.)~\cite{cordts2016cityscapes}:} Flex-X-Large outperforms DSI-MonoVit~\cite{zhang2025depth} by a significant margin on urban driving scenes. It reduces the Absolute Relative error (AbsRel) from 0.138 to 0.129 and boosts the $\delta_1$ accuracy from 0.815 to 0.833, indicating its high proficiency in handling complex multi-object traffic layouts.
		\item \textbf{Make3D (M.)~\cite{saxena2008make3d}:} On the Make3D dataset, which features drastically different camera viewpoints, resolutions, and structural characteristics compared to KITTI~\cite{6248074}, Flex-X-Large still achieves a lower AbsRel (0.279 vs. 0.288) and a higher $\delta_1$ score (0.611 vs. 0.595). This validates that our model captures intrinsic geometric cues rather than merely overfitting to dataset-specific biases.
		\item \textbf{DDAD (D.)~\cite{guizilini20203d}:} On the long-range and challenging DDAD~\cite{guizilini20203d} dataset, although DSI-MonoVit~\cite{zhang2025depth} yields a marginally better performance, Flex-X-Large delivers highly competitive results (an AbsRel of 0.158 vs. 0.156), maintaining stable and robust depth predictions.
	\end{itemize}
	In summary, these zero-shot evaluation outcomes strongly demonstrate that our framework possesses excellent generalization proficiency and real-world applicability, effectively mitigating the common domain-gap issue in self-supervised monocular depth estimation.
	
	\begin{table}[h]
		\centering
		\caption{KITTI~\cite{6248074}  Zero-Shot Generalization on Cityscapes (C.)~\cite{cordts2016cityscapes}, Make3D (M.)~\cite{saxena2008make3d}, and DDAD (D.)~\cite{guizilini20203d}. $\downarrow$ indicates lower is better, and $\uparrow$ indicates higher is better. The best results are highlighted in bold.}
		\label{tab:zeroshot}
		\resizebox{0.97\columnwidth}{!}{
			\begin{tabular}{lcccccc}
				\toprule
				Method & (C.) AbsRel $\downarrow$ & (C.) $\delta_1 \uparrow$ & (M.) AbsRel $\downarrow$ & (M.) $\delta_1 \uparrow$ & (D.) AbsRel $\downarrow$ & (D.) $\delta_1 \uparrow$ \\
				\midrule
				DSI-MonoVit~\cite{zhang2025depth} & 0.138 & 0.815 & 0.288 & 0.595 & \textbf{0.156} & \textbf{0.779} \\
				\textbf{Flex-X-Large} & \textbf{0.129} & \textbf{0.833} & \textbf{0.279} & \textbf{0.611} & 0.158 & 0.773 \\
				\bottomrule
			\end{tabular}
		}
	\end{table}

	\subsection{Qualitative Results}
	\label{sec:x_qualitative_results}
	\begin{figure*}[t]
	\centering
	
	\includegraphics[width=\textwidth]{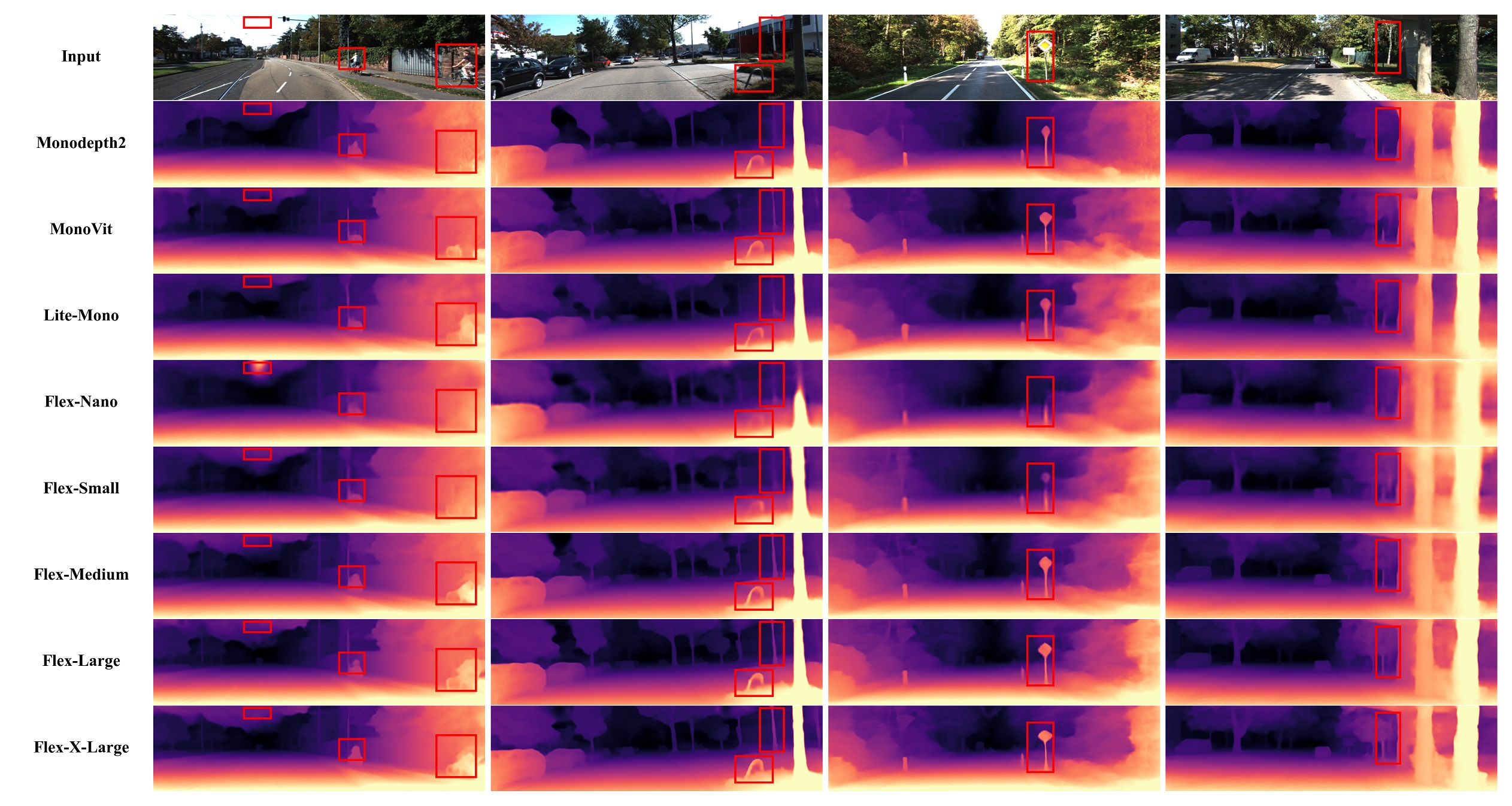}

     \caption{\textbf{Qualitative Results on the KITTI Dataset~\cite{6248074} (I).} Representative examples demonstrating the efficacy of our method in various outdoor driving scenarios. Note the accurate depth details and sharp boundaries.}
     \label{fig:kitti_x_comparison}
\end{figure*}

\begin{figure*}[t]
	\centering
		\includegraphics[width=\textwidth]{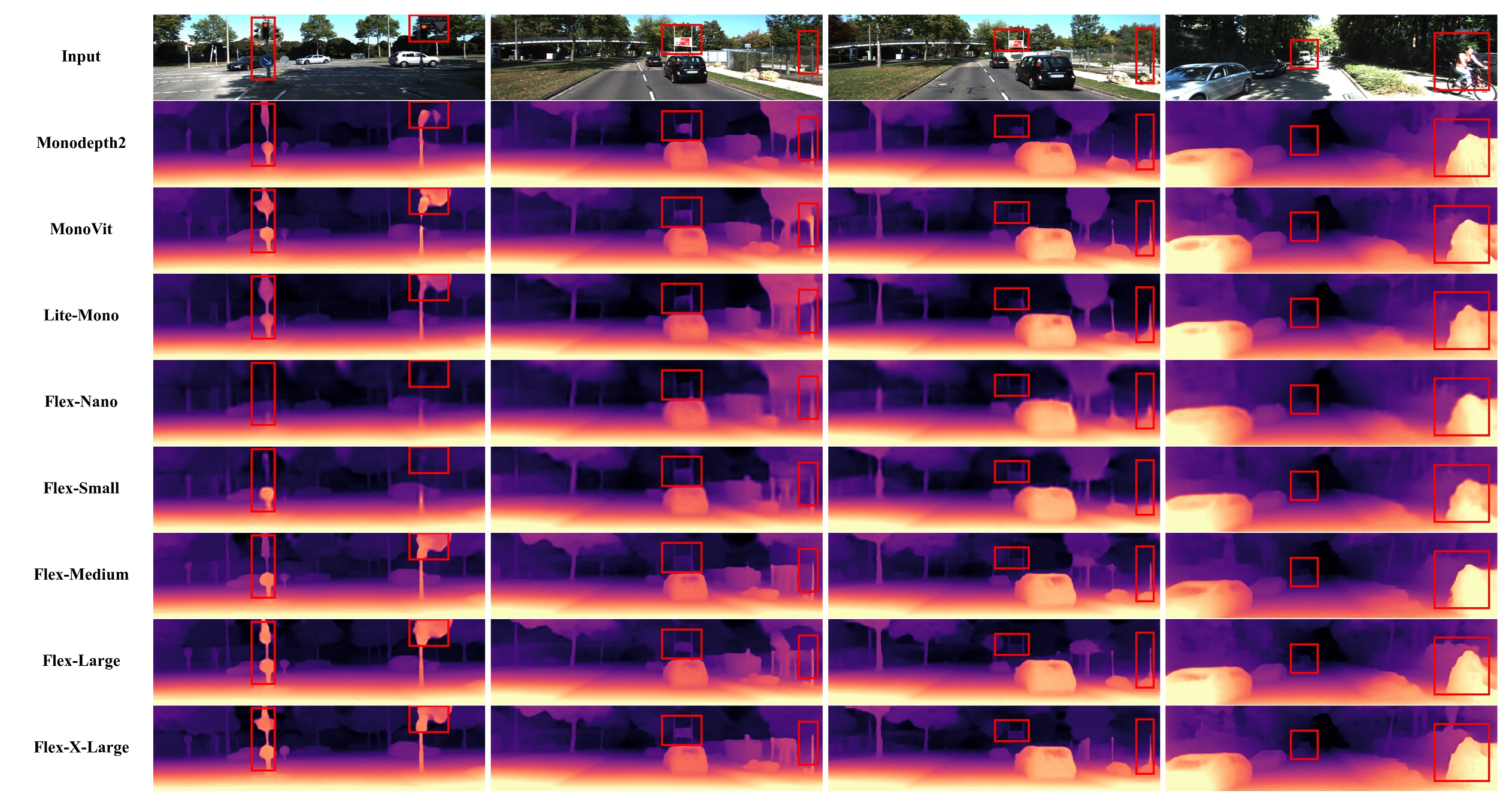}
	\caption{\textbf{Qualitative Results on the KITTI Dataset~\cite{6248074} (II).} Additional qualitative examples from the KITTI dataset, highlighting performance in challenging lighting conditions and diverse scene compositions.}
	\label{fig:kitti_x_comparison_1}
\end{figure*}

	We first present qualitative evaluations on the KITTI~\cite{6248074} in Figures~\ref{fig:kitti_x_comparison} and \ref{fig:kitti_x_comparison_1}. These visualizations highlight the model's capability to preserve high-frequency structural details, exhibiting sharp edge delineation and consistent depth gradients across scenes of varying complexity.
	
\begin{figure*}[t]
	\centering
	\includegraphics[width=\textwidth]{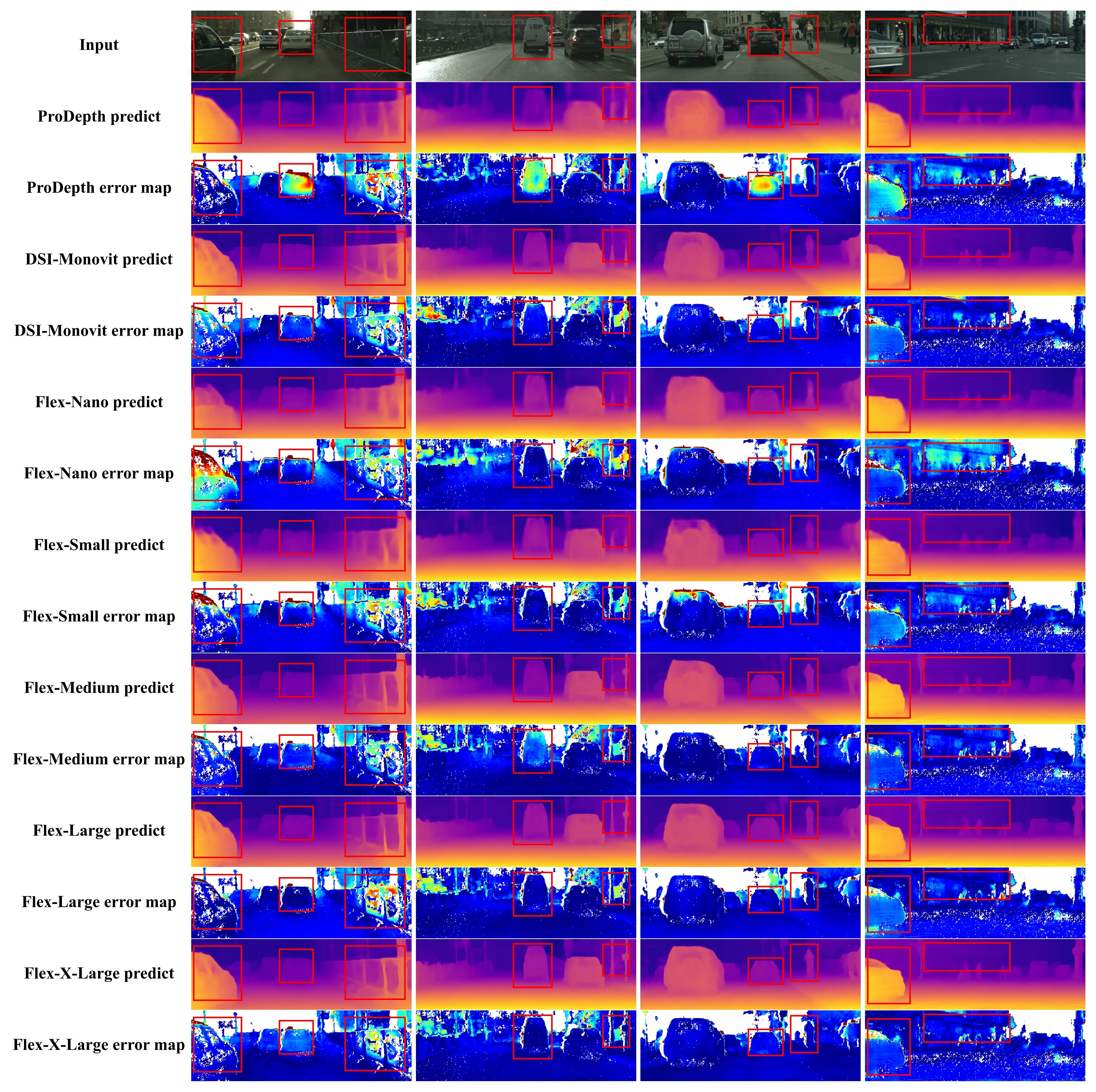}
	\caption{\textbf{Qualitative results on medium-to-long range vehicle scenes.} Distant moving vehicles appear small in the frame, posing challenges for depth estimation due to limited pixel coverage. Our single-frame method accurately delineates these vehicles without temporal context, demonstrating the effectiveness of our dynamic masking strategy in scenarios where motion cues are minimal.}
	\label{fig:cityscapes_x_comparison}
\end{figure*}

\begin{figure*}[t]
	\centering
	\includegraphics[width=\textwidth]{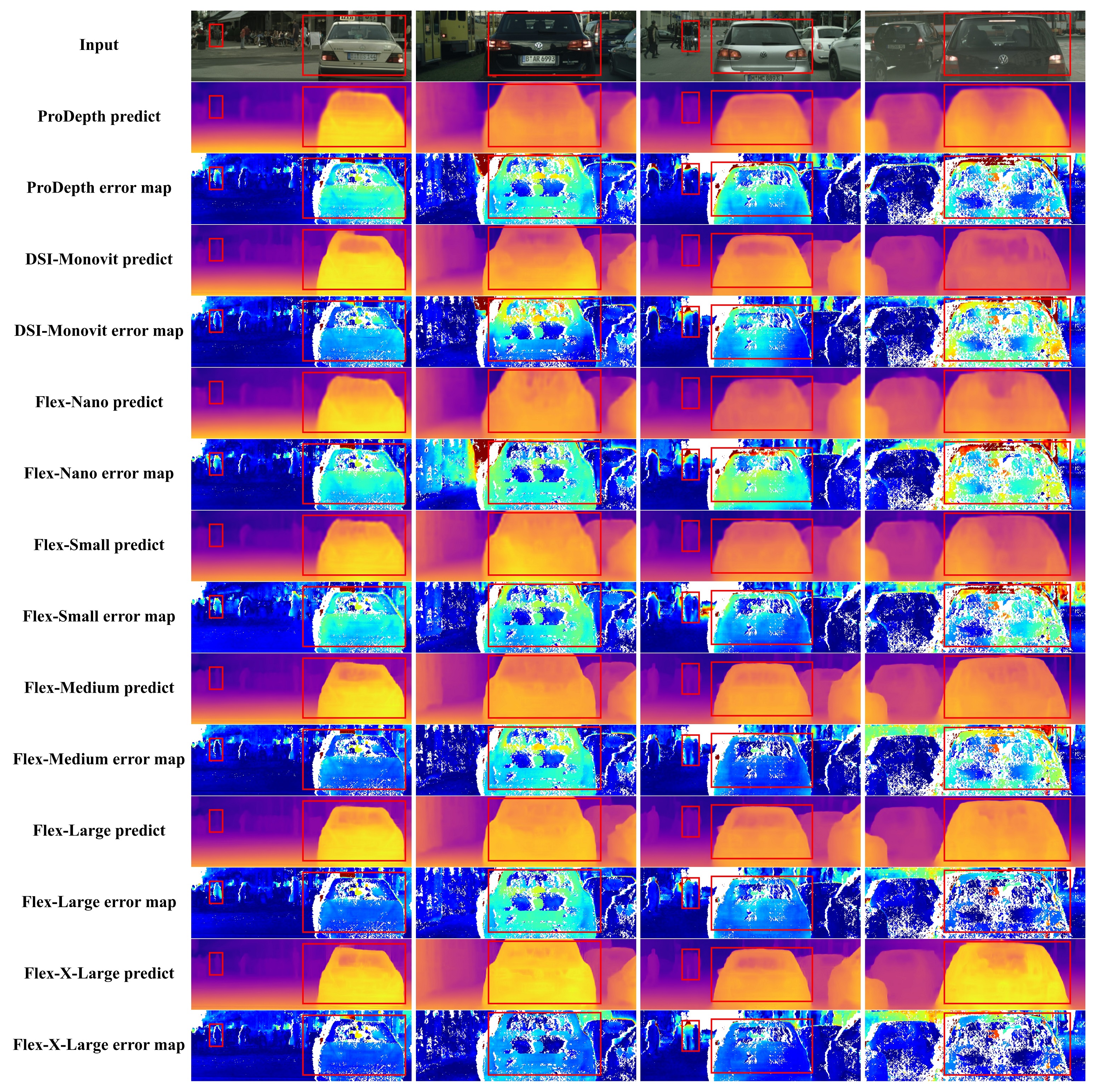}
	\caption{\textbf{Qualitative results on close-range car-following scenes.} In tailgating scenarios, dynamic objects occupy large portions of the frame, creating severe occlusion and large motion regions that challenge conventional methods. Our approach achieves artifact-free depth predictions by adaptively filtering these dominant dynamic regions during training.}
	\label{fig:cityscapes_x_comparison_1}
\end{figure*}

\begin{figure*}[t]
	\centering
	\includegraphics[width=\textwidth]{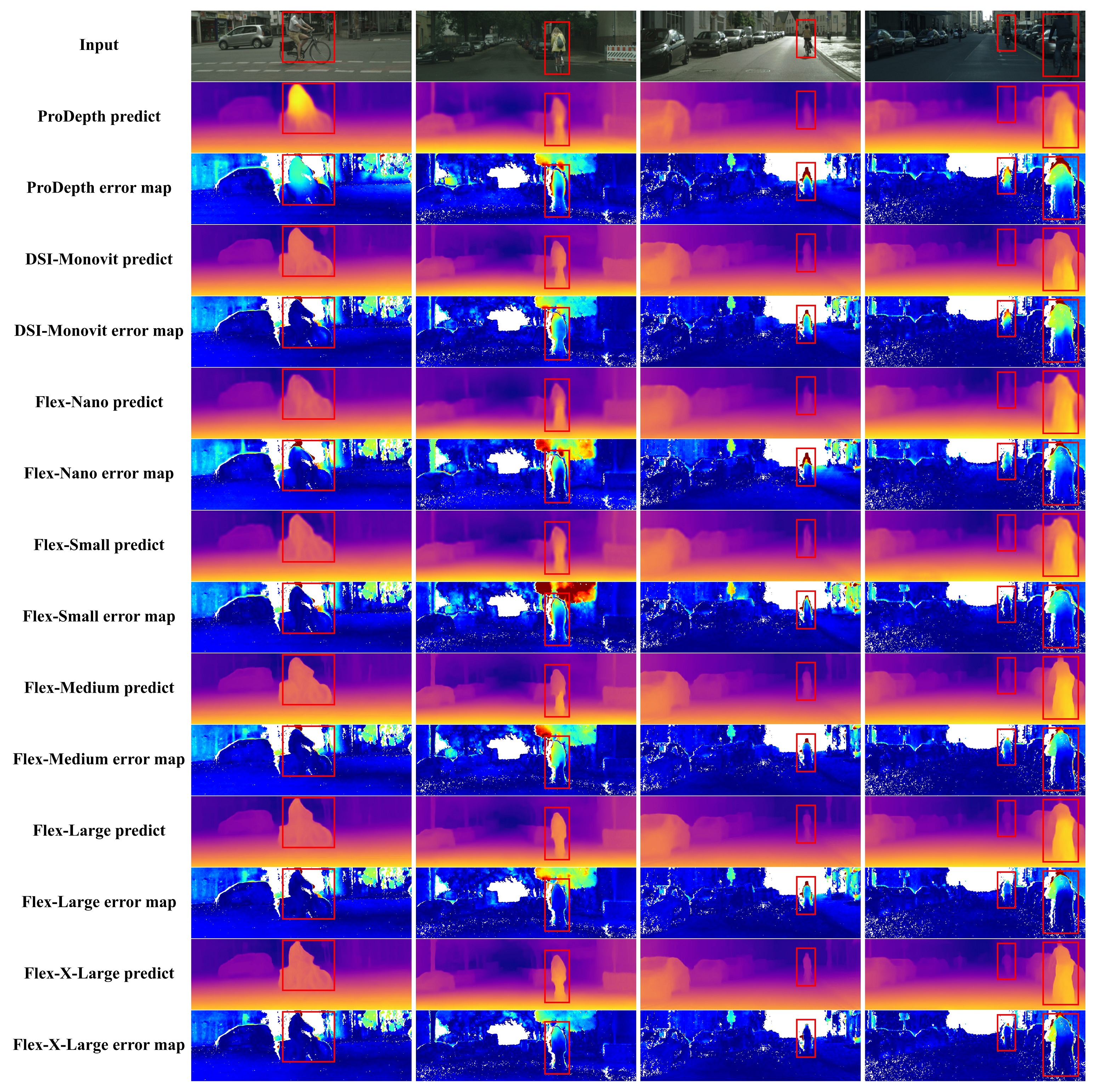}
	\caption{\textbf{Qualitative results on pedestrian and cyclist scenes.} Comparison against multi-frame ProDepth~\cite{woo2024prodepth} and single-frame DSI-Monovit~\cite{zhang2025depth}. Top: depth maps; Bottom: error maps (blue=low, red=high). Non-rigid motion from pedestrians and cyclists poses unique challenges. Our method yields lower errors on these moving subjects, outperforming both temporal and single-frame dynamic-handling approaches.}
	\label{fig:cityscapes_x_comparison_2}
\end{figure*}

	The core of our qualitative analysis focuses on dynamic urban environments in the Cityscapes dataset~\cite{cordts2016cityscapes}, shown in Figures~\ref{fig:cityscapes_x_comparison}, \ref{fig:cityscapes_x_comparison_1}, and~\ref{fig:cityscapes_x_comparison_2}. We evaluate three distinct challenging scenarios: medium-to-long range vehicle detection where dynamic objects occupy limited pixels, close-range car-following where moving vehicles dominate the frame, and pedestrian-cyclist scenes with non-rigid motion patterns.
	
	As illustrated in the comparative visualizations, our \textit{single-frame} method demonstrates superior handling of moving agents compared to both \textit{multi-frame} baselines (ManyDepth~\cite{watson2021temporal}, ProDepth~\cite{woo2024prodepth}) and \textit{single-frame dynamic-aware} approaches (DSI-Monovit~\cite{zhang2025depth}). Temporal methods often struggle with motion artifacts due to occlusion or dynamic violations of the static world assumption, particularly in close-range scenarios where large displacements occur. Meanwhile, existing single-frame dynamic methods may not fully capture the diversity of motion patterns. Our approach—driven by the proposed static-dynamic decoupling strategy—produces significantly cleaner boundaries on moving vehicles and pedestrians across all three scenarios. The error maps in Figure~\ref{fig:cityscapes_x_comparison_2} explicitly corroborate this, showing reduced error magnitude in dynamic regions even when compared against DSI-Monovit, a method specifically designed for dynamic scene handling in the single-frame setting.

	\clearpage
	\par\vfill\par

	%
	%

\end{document}